\documentclass[utf8]{FrontiersinHarvard} 

\usepackage{url,hyperref,lineno,microtype,subcaption}
\usepackage[onehalfspacing]{setspace}
\usepackage{multirow}
\usepackage{multicol}
\usepackage{cellspace}
\usepackage[section]{placeins}
\def\keyFont{\fontsize{8}{11}\helveticabold }
\def\firstAuthorLast{Dunn} 
\def\Authors{Jonathan Dunn}
\def\Address{New Zealand Institute for Language, Brain and Behaviour \\Department of Linguistics, University of Canterbury \\Christchurch, New Zealand}
\def\corrAuthor{Jonathan Dunn}

\def\corrEmail{jonathan.dunn@canterbury.ac.nz}

\begin{document}
\onecolumn
\firstpage{1}

\title {Syntactic Variation Across the Grammar: \\Modelling a Complex Adaptive System} 

\author[\firstAuthorLast ]{\Authors} 
\address{} 
\correspondance{} 

\extraAuth{}

\maketitle

\begin{abstract}

\section{}
While language is a complex adaptive system, most work on syntactic variation observes a few individual constructions in isolation from the rest of the grammar. This means that the grammar, a network which connects thousands of structures at different levels of abstraction, is reduced to a few disconnected variables. This paper quantifies the impact of such reductions by systematically modelling dialectal variation across 49 local populations of English speakers in 16 countries. We perform dialect classification with both an entire grammar as well as with isolated nodes within the grammar in order to characterize the syntactic differences between these dialects. The results show, first, that many individual nodes within the grammar are subject to variation but, in isolation, none perform as well as the grammar as a whole. This indicates that an important part of syntactic variation consists of interactions between different parts of the grammar. Second, the results show that the similarity between dialects depends heavily on the sub-set of the grammar being observed: for example, New Zealand English could be more similar to Australian English in phrasal verbs but at the same time more similar to UK English in dative phrases.

\tiny
 \keyFont{ \section{Keywords:} complex adaptive system, computational syntax, computational sociolinguistics, construction grammar, syntactic variation, dialect classification} 
\end{abstract}

\section{Introduction}

Within linguistics and cognitive science, language is increasingly viewed as a complex adaptive system \citep{Bybee2007, Beckner2009}. For example, usage-based theories of syntax like Construction Grammar \citep{g06a, l08} view the grammar as a network which contains structures at different levels of abstraction. The network structure of the grammar is made up of inheritance relations (mother-child) and similarity relations (sibling-sibling) between constructions. A \textit{construction} in this context is a symbolic mapping between form and meaning, where an individual construction is unique either syntactically or semantically. For example, there is an inheritance relationship between the schematic ditransitive construction, with examples like (1a), and idiomatic constructions, with examples like (1b) and (1c). While some of the properties of the ditransitive are inherited by these idiomatic children, they also retain unique and non-compositional meanings. It is this network structure which makes the grammar a complex system. As with any complex system, there are emergent properties of the grammar which cannot be described by looking at individual constructions in isolation.

(1a) \textit{write the store a check} \\
\indent(1b) \textit{give me a hand} \\
\indent(1c) \textit{give me a break}

The challenge is that most work on syntactic variation does exactly this: observing a few constructions that have been removed from the context of the larger grammar and modelled as discrete and independent variables. The contribution of this paper is to systematically evaluate whether the picture we get of syntactic variation changes depending on which sub-sets of the grammar we inspect. In other words, to what degree does our view of syntactic variation (for example, the similarity between New Zealand English and Australian English) depend on the sub-set of the grammar which we are observing?

This paper makes two significant contributions to models of dialectal variation: First, we examine dialect areas at three levels of spatial granularity. This includes \textit{regional dialects} (like North American English), \textit{national dialects} (like Canadian English), and \textit{local dialects} (like Ontario English). This is the first computational study to systematically experiment with different levels of granularity when modelling dialectal variation. Second, we examine different nodes or clusters of constructions within the grammar. This includes macro-clusters which include hundreds of constructions and smaller micro-clusters which contain dozens of constructions. This is the first computational study to systematically experiment with the distribution of spatial variation across an entire grammar.

In the first case, in order to understand syntactic variation we must view the population of speakers itself as a complex network. While most work on syntactic variation considers only a few segments of the population, this paper uses observations from 49 local populations distributed across 16 countries. Speakers of English  within one regional dialect are in contact with speakers of other dialects through immigration, short-term travel, media, and digital communication. Thus, the first challenge is to conduct a dialect survey across all representative populations of English speakers in order to understand syntactic variation across the entire population network.

In the second case, in order to understand syntactic variation we must view the grammar itself as a complex network so that we can observe variation in its entirety rather than in isolated and disconnected portions of the grammar. In this paper we use Computational Construction Grammar \citep{d17, Dunn2022b} to provide an unsupervised network of constructions. For these experiments, this grammar network is learned using independent data from the same register as the geographic corpora used to represent dialects (tweets).

Our theoretical question is whether syntactic variation, as captured by systematic grammatical comparisons between dozens of regional populations, is influenced by the sub-set of the grammar network which is used to model variation. We operationalize a model of dialect as a classifier which learns to predict the dialect membership of held-out samples. A dialect classifier is able to handle high-dimensional spaces, important for viewing variation across an entire grammar. And, importantly, the quality of a dialect classifier can be measured using its prediction accuracy on held-out samples. Our goal here is not simply to find sub-sets of the grammar which are in variation but rather to determine how accurate and robust these sub-sets of the grammar are for characterizing dialectal variation as a whole: for example, how accurately does a portion of the grammar characterize the difference between New Zealand and Australian English? To answer this question, we use prediction accuracy, error analysis, and feature pruning methods to determine the quality of dialect models that rely on different nodes within the grammar.

\section{Related Work}

Current knowledge of large-scale linguistic variation (i.e., across many countries) consists of (i) non-syntactic studies of lexical variation, (ii) quantitative corpus-based studies of syntactic variation, and (iii) computational studies of syntactic variation. In the first case, lexical variation is the most approachable type of linguistic variation because it does not require any learning of grammars or representations. Thus, early large-scale corpus-based studies of variation focused on the usage of lexical items \citep{eosx10, mbpgzv13, eosx14, donoso-sanchez-2017-dialectometric, Rahimi2017a}. The challenge for lexical variation is to define the envelope of variation (i.e., discover the set of alternations to avoid topic-specific models). Two main approaches to this are, first, to rely on existing dialect surveys to provide hand-crafted alternations \citep{10.3389/frai.2019.00011} and, second, to use contextual embeddings to develop clusters of senses of words \citep{Lucy2021}. In either case, lexical variation is a simpler phenomenon than syntactic variation because the number of potential alternations for each lexical item is limited. Recent work on semantic variation \citep{Dunn2023a} has expanded this scope by looking at the conceptual rather than the lexical level and using participant-based measures like abstractness ratings and age-of-acquisition to determine what causes a concept to be subject to dialectal variation.

Most corpus-based approaches to syntactic variation choose a single construction to examine and then model variation within that construction alone \citep{Buchstaller2008, Grieve2012, ss16, cr17, gs18, Deshors2020, Schneider2020, Xu2022, Rautionaho2022, Larsson2023, Li2023}. While this line of work can reveal small-scale \textit{syntactic} variation and change, it can never account for \textit{grammatical} variation. The difference between these two terms is important in this context: a single syntactic feature, like aspectual marking or noun pluralization, may be in variation but we cannot understand the variation without contextualizing it within the entire grammar. In other words, if the grammar is in fact a complex adaptive system, then measuring variation in a single construction is like assuming that the weather in Miami, Florida is independent of both the weather in Orlando and the current conditions of the Atlantic. By analogy, previous work has shown differences of behaviour in small-scale population networks vs large-scale networks \citep{Laitinen2020}. This paper examines the impact of the granularity or size of the network for both the underlying population (e.g., regional vs local dialects) and the grammar itself (e.g., different clusters of constructions within the grammar).

The first computational work which viewed syntactic variation from the perspective of a complex adaptive system used 135 grammatical alternations in English, chosen manually to include features which can be extracted using regular expressions \citep{g16}. The alternations include examples like \textit{anyone} vs \textit{anybody} and \textit{hear of} vs \textit{hear about}. This study used a corpus of letters to the editor from 240 US cities, similar in spatial granularity to the local dialects in this paper. While this early work assumed a starting set of simple alternations, it was followed by work which focused on discovering the set of variants while instead assuming the spatial boundaries of dialect areas \citep{d18b}. The advantage of this approach is that it both expands the scope of the study (by including more complex constructional features) while also scaling across languages \citep{10.3389/frai.2019.00015}. 

The other difference in these two approaches is that Grieve's early work relies on factor analysis to group together grammatical alternations according to their patterns of variation. In order to provide a measure of predictive accuracy on a held-out test set, by which a better model makes better predictions, more recent computational work has instead taken a classification approach \citep{dunn-2019-modeling, dunn-wong-2022-stability}. As discussed further in the section on Computational Construction Grammar, constructions are organized into a network structure using similarity measures directly within the grammar. This means that the nodes within which variation occurs are derived independently of the model of dialectal variation.

There are two questions which remain given this previous work: First, how much of the predictive accuracy of a dialect model is contained in different nodes within the grammar? One issue with surface-level alternations like \textit{anyone} vs \textit{anybody} is that, while in variation, they do not capture more schematic differences between dialects and, overall, do not hold much predictive power given their relative scarcity. Second, previous work has always focused on a given size of spatial granularity, usually at the country or city level. This paper uses three levels of granularity to help understand complexity in the underlying population network as well.

\section{Data}

The data used for these experiments is drawn from geo-referenced social media posts (tweets), a source with a long history as an observation of dialectal production (c.f., \citealt{eosx14, 10.3389/frai.2019.00015, 10.3389/frai.2019.00011}). The corpus is drawn from 16 English-speaking countries, as shown in Table \ref{tab:1}. Countries are grouped into larger regional dialects (such as North American vs South Asian English). And each country is divided into potentially many sub-areas using spatial clustering (for example, American English is divided into nine local dialect groups). Language identification is undertaken using two existing models to make the corpus comparable with existing work on mapping digital language use \citep{Dunn2020, Dunn2022}. This corpus thus provides three levels of granularity: 7 regions, 16 countries, and 49 local areas. 

\begin{table*}[h]
\centering
\begin{tabular}{|lllc|Sr|r|}
\hline
\textbf{Region} & \multicolumn{2}{c}{\textbf{Country}} & \textbf{Areas} & \textbf{Corpus Size} & \textbf{N. Samples} \\
\hline
\multirow{2}{*}{Africa, Southern} & South Africa & ZA & 2 & 9,205,166 words & 2,299 \\
~ & Zimbabwe & ZW & 1 & 3,252,685 words & 767 \\
\hline
\multirow{2}{*}{Africa, Sub-Saharan} & Kenya & KE & 3 & 5,537,772 words & 1,310 \\
~ & Nigeria & NG & 3 & 6,139,948 words & 1,492 \\
\hline
\multirow{2}{*}{North America} & Canada & CA & 4 & 16,801,386 words & 4,261 \\
~  & United States & US & 9 & 26,050,840 words & 5,802 \\
\hline
\multirow{3}{*}{Asia, South} & Bangladesh & BD & 2 & 6,287,670 words & 1,649 \\
~ & India & IN & 7 & 28,935,606 words & 7,045 \\
~ & Pakistan & PK & 2 & 12,765,491 words & 3,271 \\
\hline
\multirow{3}{*}{Asia, Southeast} & Indonesia & ID & 1 & 2,392,074 words & 546 \\
~ & Malaysia & MY & 1 & 8,580,789 words & 2,052 \\
~ & Philippines & PH & 2 & 9,907,209 words & 2,402 \\
\hline
\multirow{2}{*}{Europe} & Ireland & IE & 1 & 13,287,397 words & 3,360 \\
~ & United Kingdom & UK & 5 & 20,307,094 words & 4,890 \\
\hline
\multirow{2}{*}{Oceania} & Australia & AU & 4 & 23,163,447 words & 5,914 \\
~ & New Zealand & NZ & 2 & 8,113,382 words & 2,047 \\
\hline
\textbf{Total} & \multicolumn{2}{c}{\textbf{16 Countries}} & \textbf{49 Areas} & \textbf{200,727,956 words} & \textbf{49,107} \\
\hline
  \end{tabular}
  \caption{Distribution of Sub-Corpora by Region. Each sample is a unique sub-corpus with the same distribution of keywords, each approximately 3,910 words in length.}
  \label{tab:1}
\end{table*}

The main challenge is to control for other sources of variation like topic or register that would lead to a successful classification model but would not be directly connected with the local population being observed. In other words, we need to constrain the production of the local populations to a specific set of topics: if New Zealand tweets are focused on economics and Australian tweets on rugby, the impact of register would be a potential confound. For this reason, we develop a set of 250 common lexical items (c.f., Appendix 1) which are neither purely functional (like \textit{the} is) nor purely topical (like \textit{Biden} is). For each location we create sub-corpora which are composed of one unique tweet for each of these keywords. Thus, each location is represented by a number of sub-corpora which each have the same fixed distribution of key lexical items. This allows us to control for wide variations in topic or content or register, factors that would otherwise potentially contribute non-dialectal sources of variation.

Each sub-corpus thus contains 250 tweets, one for each keyword. This creates sub-corpora with an average of approximately 3,910 words. The distribution of sub-corpora (called \textit{samples}) is show in Table \ref{tab:1}. For example, the US is represented by a corpus of 26 million words divided into 5,802 individual samples. Because these samples have the same distribution of lexical items, the prediction accuracy of the dialect classifier should not be influenced by topic-specific patterns in each region. The use of lexically-balanced samples, while important for forcing a focus on dialectal variation, reduces the overall size of the corpus that is available because tweets without the required keywords are discarded.

\begin{figure*}[t]
\centering
\fbox{\includegraphics[width = 500pt]{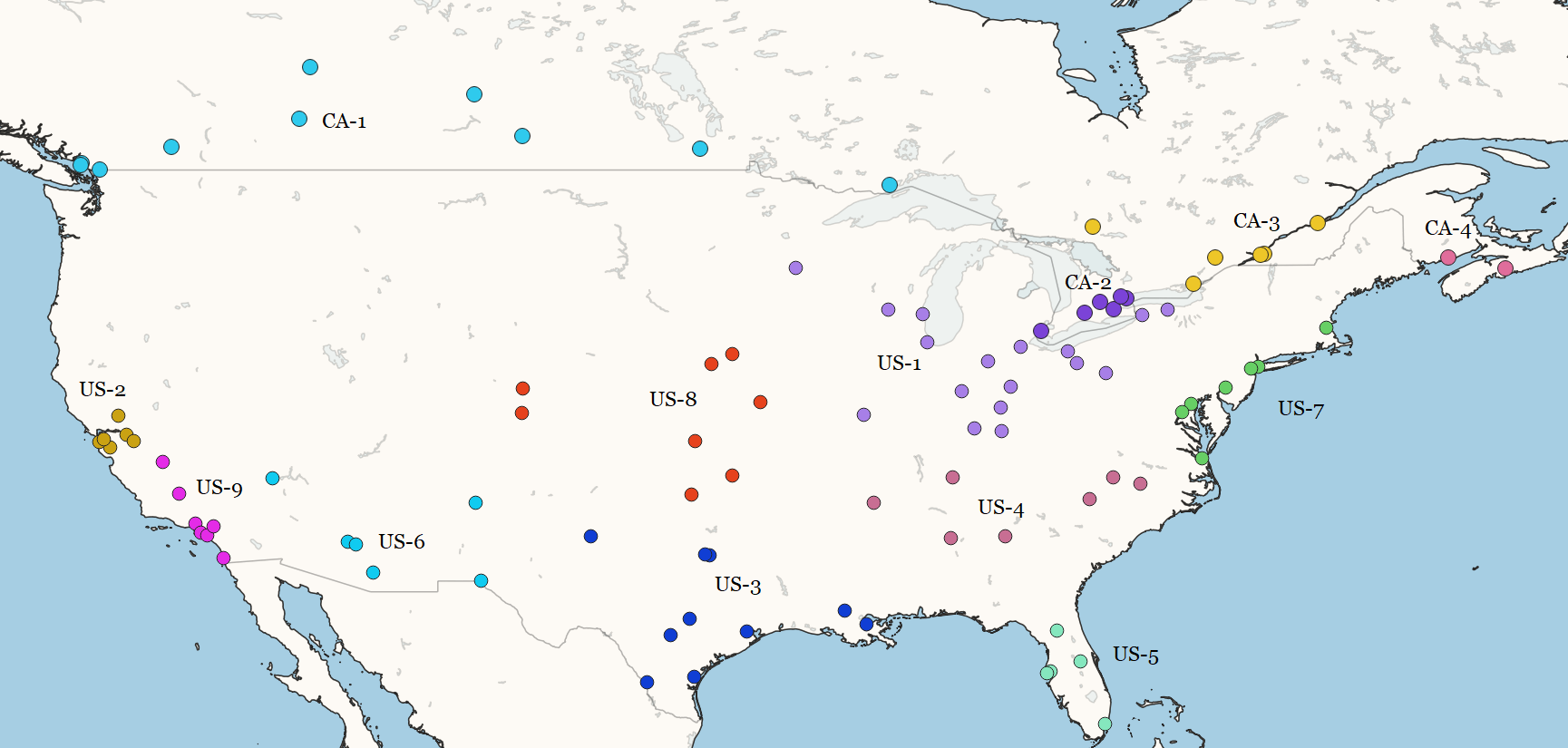}}
\caption{Distribution of Local Dialects in North America}
\label{fig:north_america}
\end{figure*}

To form the local areas, we organize the data around the nearest airport (within a threshold of a 25km radius) as a proxy for urban areas. We then use the density-based H-DBSCAN algorithm to cluster airports into groups that represent contiguous local areas \citep{Campello2013a, Campello2015, McInnes2017}. The result is a set of local areas within a country, each of which is composed of multiple adjacent urban areas. For example, the nine areas within the United States are shown in Figure \ref{fig:north_america}, where each color represents a different group. Manual adjustments of unclustered or borderline points is then undertaken to produce the final clusters. The complete set of local areas are documented in the supplementary materials. It is important to note that these local areas are entirely spatial in the sense that no linguistic information has been included in their formation. These areas represent local geographic groups, but not a linguistically-defined speech community.

Our experiments in this paper operate at three levels of spatial granularity: first, distinguishing between regional dialects, like North American English vs South Asian English; second, distinguishing between country-level dialects, like American English vs Canadian English; third, distinguishing between local dialects within regions, like Midwestern American English vs Central Canadian English. These different levels of granularity allow us to test how well different portions of the grammar are able to model increasingly fine distinctions between dialects.

\section*{Methods}

The basic approach in this paper is to use an unsupervised grammar derived from the Construction Grammar paradigm (CxG) as a feature space for dialect classification. A dialect classifier is trained to predict the regional dialect of held-out samples using the frequency of constructions within the sample. This section describes in more detail both the grammar model and the dialect models.

\subsection*{Computational Construction Grammar}

A construction grammar is a network of form-meaning mappings at various levels of schematicity. As discussed above, the grammar is a network with inheritance relationships and similarity relationships between pairs of constructions. CxG is a usage-based approach to syntax which, in practical terms, means that more item-specific constructions are learned first and then generalized into more schematic constructions \citep{Nevens2022, Doumen2023}. The grammar learning algorithm used in this paper is taken from previous work \citep{d17, d18, Dunn2019freq, Dunn2021a, Dunn2022b, Dunn2021b}, with the grammar trained from the same register as the dialectal data (tweets). Rather than describe the computational details of this line of work, this section instead analyzes constructions within the grammar as examples of the kinds of features used to model syntactic variation. The complete grammar together with examples is available in the supplementary material and the codebase for computational CxG is available as a Python package.\footnote{\href{https://github.com/jonathandunn/c2xg/tree/v2.0}{https://github.com/jonathandunn/c2xg/tree/v2.0}}

\begin{figure*}[!h]
\centering
\includegraphics[width = 500pt]{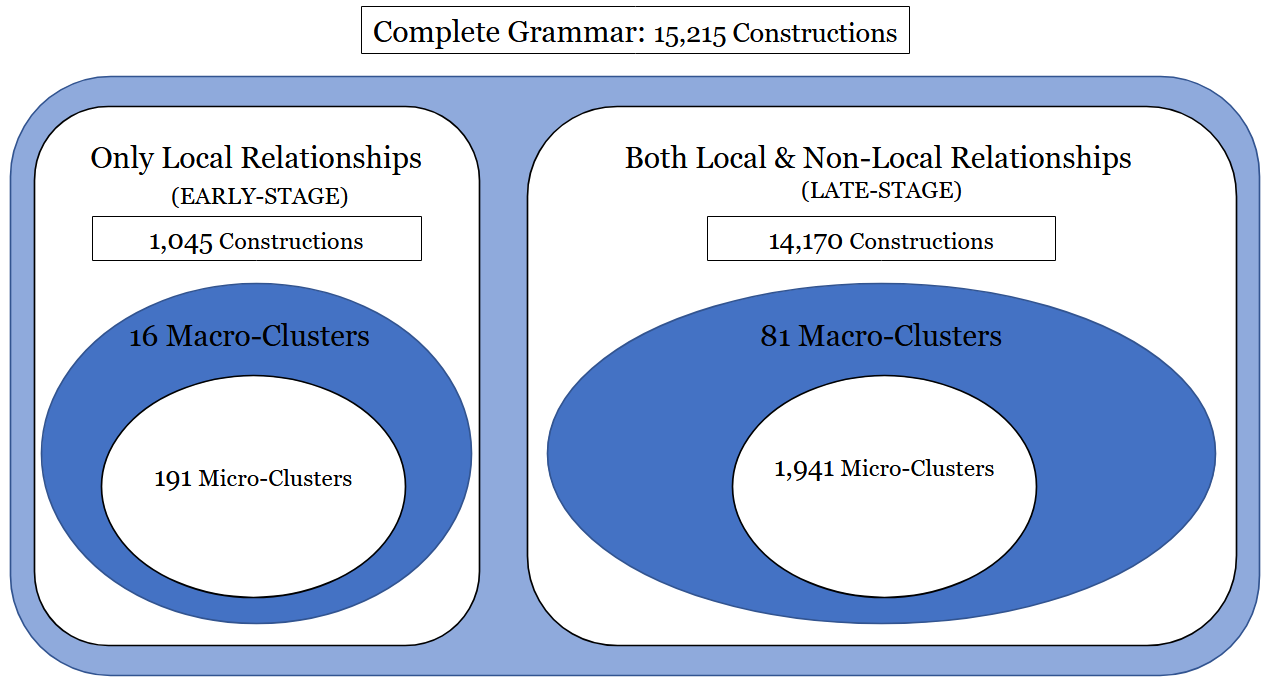}
\caption{Break-Down of the Grammar Used in the Experiments by Construction Type}
\label{fig:grammar}
\end{figure*}

A break-down of the grammar used in the experiments is shown in Figure \ref{fig:grammar}, containing a total of 15,215 individual constructions. Constructions are represented as a series of slot-constraints and the first distinction between constructions involves the types of constraints used. Computational CxG uses three types of slot-fillers: lexical (\textsc{lex}, for item-specific constraints), syntactic (\textsc{syn}, for form-based or local co-occurrence constraints), and semantic (\textsc{sem}, for meaning-based or long-distance co-occurrence constraints). As shown in (2), slots are separated by dashes in the notation used here. Thus, \textsc{syn} in (2) describes the type of constraint and \textit{determined-permitted} provides its value using two central exemplars of that constraint. Examples or tokens of the construction from a test corpus of tweets are shown in (2a) through (2d).

\noindent(2) [ \textsc{syn}: \textit{determined-permitted} -- \textsc{syn}: \textit{to} -- \textsc{syn}: \textit{pushover-backtrack} ] \\
\indent(2a) refused to play \\
\indent(2b) tried to watch \\
\indent(2c) trying to run \\
\indent(2d) continue to drive

Thus, the construction in (2) contains three slots, each defined using a syntactic constraint. These constraints are categories learned at the same time that the grammar itself is learned, formulated within an embedding space. An embedding that captures local co-occurrence information is used for formulating syntactic constraints (a continuous bag-of-words fastText model with a window size of 1) while an embedding which instead captures long-distance co-occurrence information is used for formulating semantic constraints (a skip-gram fastText model with a window size of 5). Constraints are then formulated as centroids within that embedding space. Thus, the tokens for the construction in (2) are shown in (2a) through (2d). For the first slot-constraint, the name (\textit{determined-permitted}) is derived from the lexical items closest to the centroid of the constraint. The proto-type structure of categories is modeled using cosine distance as a measure of how well a particular slot-filler satisfies the constraint. Here the lexical items “reluctant", “ready", “refusal", and “willingness" appear as fillers sufficiently close to the centroid to satisfy the slot-constraint. The contruction itself is a complex verb phrase in which the main verb encodes the agent's attempts to carry out the event encoded in the infinitive verb. This can be contrasted semantically with the construction in (3), which has the same form but instead encodes the agent's preparation for carrying out the social action encoded in the infinitive verb.

\noindent(3) [ \textsc{syn}: \textit{determined-permitted} -- \textsc{syn}: \textit{to} -- \textsc{syn}: \textit{demonstrate-reiterate} ] \\
\indent(3a) reluctant to speak \\
\indent(3b) ready to exercise \\
\indent(3c) refusal to recognize \\
\indent(3d) willingness to govern

An important idea in CxG is that structure is learned gradually, starting with item-specific surface forms and moving to increasingly schematic and productive constructions. This is called \textit{scaffolded learning} because the grammar has access to its own previous analysis for the purpose of building more complex constructions \citep{Dunn2022b}. In computational CxG this is modelled by learning over iterations with different sets of constraints available. For example, the constructions in (2) and (3) are learned with only access to the syntactic constraints, while the constructions in (4) and (5) have access to lexical and semantic constraints as well. This allows grammars to become more complex while not assuming basic structures or categorizations until they have been learned. In the dialect experiments below we distinguish between \textsc{early-stage} grammars (which only contain syntactic constraints) and \textsc{late-stage} grammars (which contain lexical, syntactic, and semantic constraints).

\noindent(4) [ \textsc{lex}: “the" -- \textsc{sem}: \textit{way} -- \textsc{lex}: “to" ] \\
\indent(4a) the chance to \\
\indent(4b) the way to \\
\indent(4c) the path to \\
\indent(4d) the steps to

Constructions have different levels of abstractness or schematicity. For example, the construction in (4) functions as a modifier, as in the X position in the sentence “Tell me [X] bake yeast bread." This construction is not purely item-specific because it has multiple types or examples. But it is less productive than the location-based noun phrase construction in (5) which will have many more types in a corpus of the same size. CxG is a form of lexico-grammar in the sense that there is a continuum between item-specific and schematic constructions, exemplified here by (4) and (5), respectively. The existence of constructions at different levels of abstraction makes it especially important to view the grammar as a network with similar constructions arranged in local nodes within the grammar.

\noindent(5) [ \textsc{lex}: “the" -- \textsc{sem}: \textit{streets} ] \\
\indent(5a) the street \\
\indent(5b) the sidewalk \\
\indent(5c) the pavement \\
\indent(5d) the avenues

A grammar or constructicon is not simply a set of constructions but rather a network with both taxonomic and similarity relationships between constructions. In computational CxG this is modelled by using pairwise similarity relationships between constructions at two levels: (i) representational similarity (how similar are the slot-constraints which define the construction) and (ii) token-based similarity (how similar are the examples or tokens of two constructions given a test corpus). Matrices of these two pairwise similarity measures are used to cluster constructions into smaller and then larger groups. For example, the phrasal verbs in (6) through (8) are members of a single cluster of phrasal verbs. Each individual construction has a specific meaning: in (6), focusing on the social attributes of a communication event; in (7), focusing on a horizontally-situated motion event; in (8), focusing on a motion event interpreted as a social state. These constructions each have a unique meaning but a shared form. The point here is that at a higher-order of structure, there are a number of phrasal verb constructions which share the same schema. These constructions have sibling relationships with other phrasal verbs and a taxonomic relationship with the more schematic phrasal verb construction. These phrasal verbs are an example of a \textit{micro-cluster} referenced in the dialect experiments below (c.f., \citealt{DunnInRevision}).

\noindent(6) [ \textsc{sem}: \textit{screaming-yelling} -- \textsc{syn}: \textit{through} ] \\
\indent(6a) stomping around \\
\indent(6b) cackling on \\
\indent(6c) shouting out \\
\indent(6d) drooling over

\noindent(7) [ \textsc{sem}: \textit{rolled-turned} -- \textsc{syn}: \textit{through} ] \\
\indent(7a) rolling out \\
\indent(7b) slid around \\
\indent(7c) wiped out \\
\indent(7d) swept through

\noindent(8) [ \textsc{sem}: \textit{sticking-hanging} -- \textsc{syn}: \textit{through} ] \\
\indent(8a) poking around \\
\indent(8b) hanging out \\
\indent(8c) stick around \\
\indent(8d) hanging around

An even larger structure within the grammar is based on groups of these micro-clusters, structures which we will call \textit{macro-clusters}. A macro-cluster is much larger because it contains many sub-clusters which themselves contain individual constructions. An example of a macro-cluster is given with five constructions in (9) through (13) which all belong to same neighborhood of the grammar. The partial noun phrase in (9) points to a particular sub-set of some entity (as in, “parts of the recording"). The partial adpositional phrase in (10) points specifically to the end of some temporal entity (as in, “towards the end of the show"). In contrast, the partial noun phrase in (11) points a particular sub-set of a spatial location (as in, “the edge of the sofa"). A more specific noun phrase in (12) points to a sub-set of a spatial location with a fixed level of granularity (i.e., at the level of a city or state). And, finally, in (13) an adpositional phrase points to a location within a spatial object. The basic idea here is to use these micro-clusters and macro-clusters as features for dialect classification in order to determine how variation is distributed across the grammar.

\noindent(9) [ \textsc{sem}: \textit{part} -- \textsc{lex}: “of" -- \textsc{syn}: \textit{the} ] \\
\indent(9a) parts of the \\
\indent(9b) portion of the \\
\indent(9c) class of the \\
\indent(9d) division of the

\noindent(10) [ \textsc{syn}: \textit{through} -- \textsc{sem}: \textit{which-whereas} -- \textsc{lex}: “end" -- \textsc{lex}: “of" -- \textsc{syn}: \textit{the} ] \\
\indent(10a) at the end of the \\
\indent(10b) before the end of the \\
\indent(10c) towards the end of the

\noindent(11) [ \textsc{sem}: \textit{which-whereas} -- \textsc{sem}: \textit{way} -- \textsc{lex}: “of" ] \\
\indent(11a) the edge of \\
\indent(11b) the side of \\
\indent(11c) the corner of \\
\indent(11d) the stretch of
	
\noindent(12) [ \textsc{sem}: \textit{which-whereas} -- \textsc{syn}: \textit{southside-northside} -- \textsc{syn}: \textit{chicagoland} ] \\
\indent(12a) in north texas \\
\indent(12b) of southern california \\
\indent(12c) in downtown dallas \\
\indent(12d) the southside chicago

\noindent(13) [ \textsc{lex}: “of" -- \textsc{syn}: \textit{the} -- \textsc{syn}: \textit{courtyard-balcony} ] \\
\indent(13a) of the gorge \\
\indent(13b) of the closet \\
\indent(13c) of the room \\
\indent(13d) of the palace

The examples in this section have illustrated some of the fundamental properties of CxG and also provide a discussion of some of the features which are used in the dialect classification study. A more detailed linguistic examination of the contents of a grammar like this is available elsewhere \citep{Dunn2023}. A break-down of the contents of the grammar is shown in Figure \ref{fig:grammar}. The 15,215 total constructions are first divided into different scaffolds (early-stage vs late-stage), with a smaller number of local-only constructions which tend to be more schematic (1,045 vs. 14,170 constructions in the late-stage grammar). This grammar has a network structure and contains 2,132 micro-clusters (e.g., the phrasal verbs discussed above). At an even higher level of structure, there are 97 macro-clusters or neighborhoods within the grammar (e.g., the sub-set referencing constructions discussed above). We can thus look at variation across the entire grammar, across different levels of scaffolded structure, and across different levels of abstraction. The main reason for doing this is to determine whether all nodes within the grammar vary across dialects in the same way.

\subsection*{Dialect Classification}

A dialect classifier is a supervised discriminative approach to modelling dialects: given labelled training data, the model learns to distinguish between dialects like American and Canadian English using syntactic features from computational CxG. There are two advantages to taking a classification approach: First, classifiers work well in high-dimensional spaces while more traditional methods from quantitative sociolinguistics do not scale across tens of thousands of potentially redundant structural features. Second, dialect classifiers provide a ground-truth measure of quality in the form of prediction accuracy: we know how robust a classifier model is given how well it is able to distinguish between different dialects. The classification of dialects or varieties of a language is a robust area, although most work views this as an engineering challenge rather than as a way to learn about dialects themselves \citep{Belinkov2016, gpaa16, b18, Kroon2018, zampieri_nakov_scherrer_2020}.

Following previous work on using dialect classification to model linguistic variation \citep{d18b, 10.3389/frai.2019.00015, dunn-2019-modeling, dunn-wong-2022-stability}, we use a Linear Support Vector Machine for classification. The advantage of an SVM over neural classifiers is that we can inspect the features which are useful for dialect classification; and the advantage over Logistic Regression or Naive Bayes is a better handling of redundant or correlated features.

The data is divided into a training set (80\%) and a testing set (20\%) with the same split used for each classification experiment. This means that the models for each level of spatial granularity (region, country, local area) are directly comparable across feature types. These dialect classifiers become our means of observing a grammar's ability to capture spatial variation: a better grammar models dialectal variation with a higher prediction accuracy. This means that it is better able to describe the differences between dialects. For instance, the usage of phrasal verbs in (6) to (8) might differ significantly between Canada and New Zealand English, while at the same time accounting for only a minimal percentage of the overall syntactic difference between these dialects. A predictive model like a classifier, however, is evaluated precisely on how well it characterizes the total syntactic difference between each dialect.\footnote{This holds true unless multiple models reach the same ceiling of accuracy, a situation which does not occur in this study.}

\begin{figure*}[t]
\centering
\includegraphics[width = 500pt]{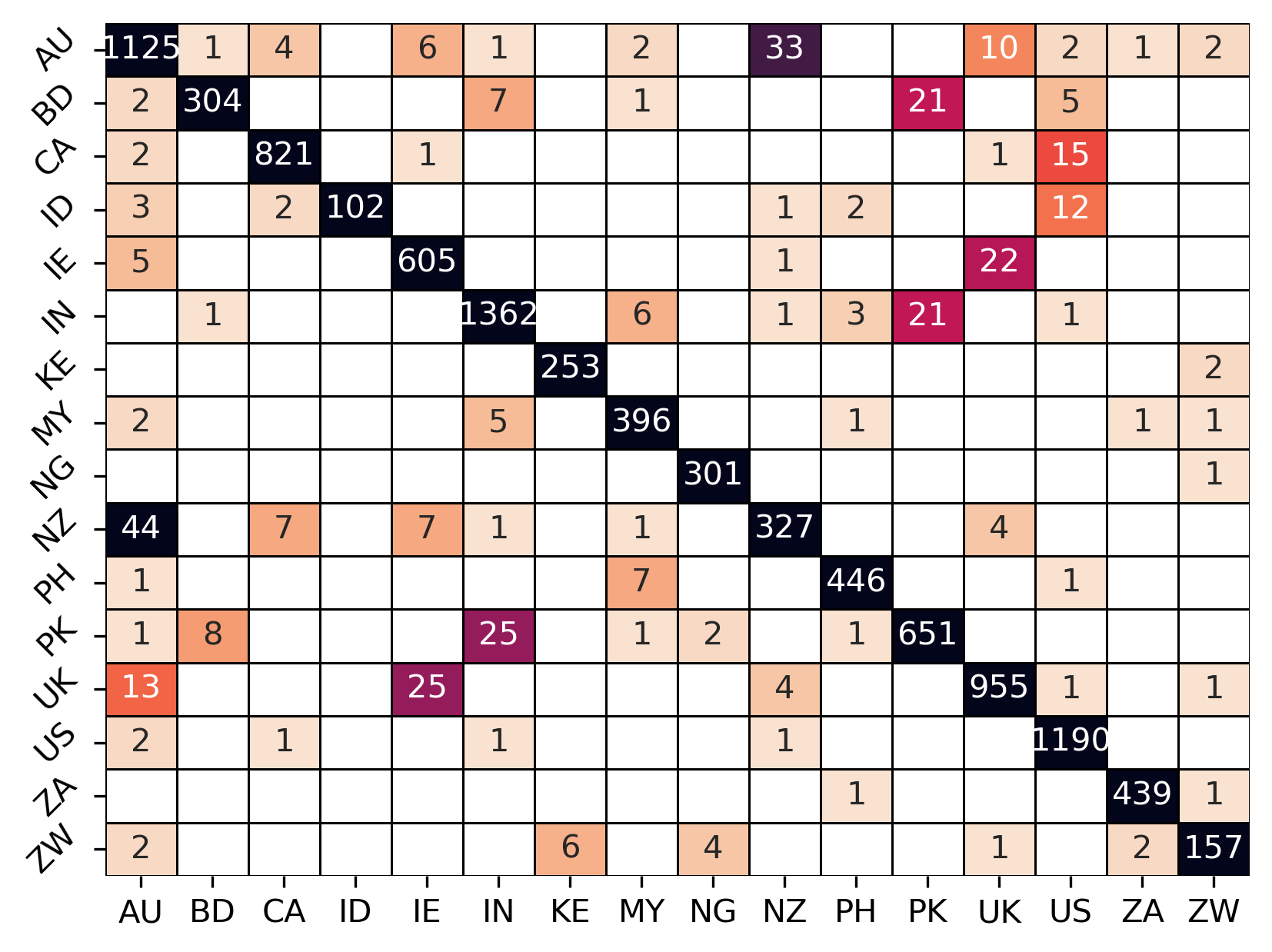}
\caption{Distribution of Errors in Country-Level Dialect Model with Late-Stage Grammar}
\label{fig:errors}
\end{figure*}

We can further explore each dialect model using a confusion matrix to examine the types of errors made. For instance, Figure \ref{fig:errors} shows the distribution of errors for the country-level classification of dialects using the late-stage grammar. Correct predictions occur on the diagonal; given the weighted f-score of 0.97, most predictions are correct in this case. Yet the number of errors for each pair of dialects reflects their similarity. Thus, the most similar countries are (i) New Zealand and Australia with 35+28 errors, (ii) Ireland and the UK with 26+21 errors, (iii) Pakistan and India with 16+21 errors, and (iv) Canada and the US with 17+1 errors. The confusion matrix also reveals the dominant variety, in the sense that only one sample of American English is mistakenly predicted to be Canadian (rows represent the true labels) while 17 samples of Canadian English are mistaken for American English. Thus, these are asymmetrical errors. The point here is that these models allow us to measure not only the overall quality of the dialect classifier (its prediction accuracy represented by the f-score) but also determine which dialects are the most similar. This, in turn, means that we can measure the stability of dialect similarity across different nodes within the grammar. Note that this error-based similarity is different from feature-based similarity measures \citep{10.3389/frai.2019.00023, 10.3389/frai.2019.00015} which instead operate on the grammatical representations themselves. However, the more reliable a classifier is (i.e., the higher its prediction accuracy), the more its errors can be used directly as a similarity measure. Our focus here is on understanding the dialect model itself by examining its false positive and false negative errors.

\section{Results}

The basic question in this paper is whether different nodes within the grammar equally capture dialectal variation and whether the resulting picture of dialectal relations is the same regardless of which node we examine. Now that we have discussed the grammatical features used (from computational CxG) and the means of observing syntactic variation (a dialect classifier), this section analyzes the results. We start with region-level dialects before moving to country-level and then local dialects.

\subsection{Regional Dialects}

The 16 countries used in this study are grouped into regions as shown in Table \ref{tab:1}, resulting in seven larger macro-dialects as shown in Table \ref{tab:results_region}. The table shows three measures: precision (false positive errors), recall (false negative errors), and f-score. On the left the table shows the dialect performance with the late-stage grammar (i.e., with constructions containing all three types of slot-constraints) and on the right the early-stage  grammar (i.e., with constructions containing only syntactic slot constraints). The late-stage grammar performs better (0.99 f-score vs 0.93) but both perform well. In the early-stage  grammar, two dialects are a particular source of the lower performance: Southern African English (South Africa and Zimbabwe) and Oceanic English (Australia and New Zealand).

\begin{table*}
\centering
\begin{tabular}{|Sl|ccc|ccc|}
\hline
~ & \multicolumn{3}{Sc|}{\textbf{Late-Stage Grammar}} & \multicolumn{3}{Sc|}{\textbf{Early-Stage Grammar}} \\
\textbf{Region} & \textit{Precision} & \textit{Recall} & \textit{F-Score} & \textit{Precision} & \textit{Recall} & \textit{F-Score} \\
\hline
Africa, Southern & 0.99 & 0.97 & 0.98 & 0.88 & 0.81 & 0.84 \\
Africa, Sub & 0.98 & 0.99 & 0.99 & 0.92 & 0.91 & 0.91 \\
North America & 0.99 & 1.00 & 0.99 & 0.93 & 0.95 & 0.94 \\
Asia, South & 1.00 & 1.00 & 1.00 & 0.99 & 0.98 & 0.98 \\
Asia, Southeast & 0.99 & 0.98 & 0.98 & 0.92 & 0.92 & 0.92 \\
Western Europe & 0.98 & 0.99 & 0.99 & 0.94 & 0.94 & 0.94 \\
Oceania & 0.98 & 0.98 & 0.98 & 0.88 & 0.89 & 0.88 \\
\hline
\textbf{Weighted Avg} & \textbf{0.99} & \textbf{0.99} & \textbf{0.99} & \textbf{0.93} & \textbf{0.93} & \textbf{0.93} \\
\hline
  \end{tabular}
  \caption{Performance of Dialect Classifier With Regional Dialects: Late-Stage Constructions (Left, with all constraint types) and Early-Stage Constructions (Right, with only syntactic constraints)}
  \label{tab:results_region}
\end{table*}

\begin{figure*}
\centering
\includegraphics[width = 500pt]{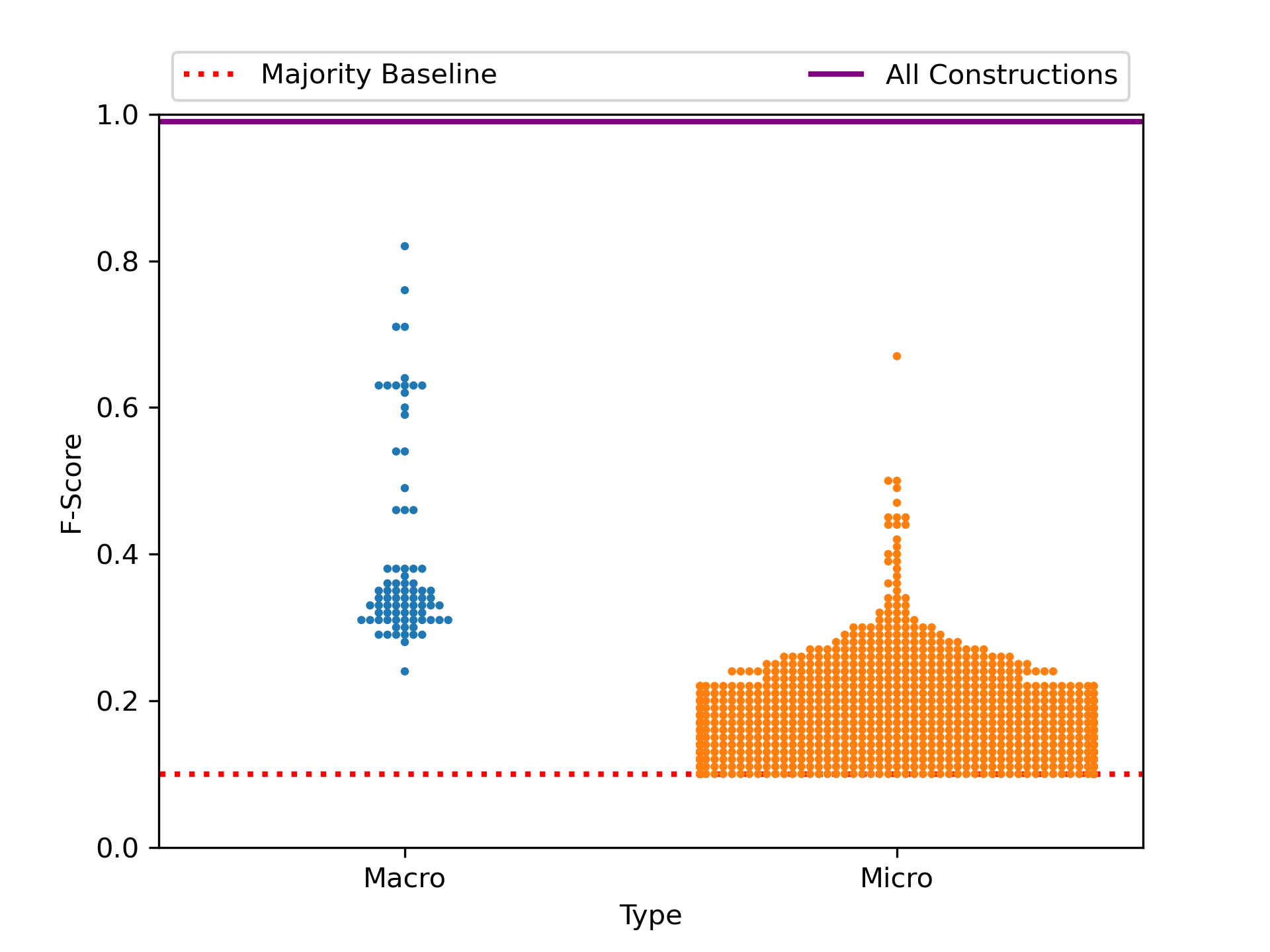}
\caption{Distribution of Classification Performance Across Sub-Sets of the Grammar, Regional Dialect with the Late-Stage Grammar. Each macro-cluster and micro-cluster of constructions is plotted with its f-score on the dialect classification task, with both the performance of the entire late-stage grammar and the majority baseline also shown.}
\label{fig:results_region_full}
\end{figure*}

The overall f-score of the late-stage grammar (0.99) tells us that, at this level of spatial granularity, the grammar as a single network is able to make almost perfect distinctions between the syntax of these regional dialects. But how does this ability to capture variation spread across nodes within the grammar? This is explored in Figure \ref{fig:results_region_full}, which contrasts the f-score of individual nodes using the macro-clusters and micro-clusters discussed above. A macro-cluster is a fairly large group of constructions based on pairwise similarity between the constructions themselves. A micro-cluster is a smaller group within a macro-cluster, based on pairwise similarity between instances or tokens of constructions in a test corpus \citep{DunnInRevision}. Each point in this figure is a single node of the grammar (blue for macro-clusters and orange for micro-clusters). The position of the point on the y-axis reflects the prediction performance for regional dialect classification using only that portion of the grammar. The dotted horizontal line represents the majority baseline, at which classification performance is no better than chance.

This figure shows us, first, that the grammar as a whole is always better able to describe syntactic variation than individual nodes within that grammar. This is important in itself because, if language is a complex adaptive system, then it follows that variation in language is at least in part an emergent phenomenon in which the interaction between elements (here constructions) cannot be characterized by observing those elements in isolation. The second finding which this figure shows is that individual nodes vary greatly in the amount of dialectal variation which they are subject to. Thus, several nodes in isolation are able to characterize between 70\% and 80\% of the overall variation, but others characterize very little. Within micro-clusters, especially, we see that many nodes within the grammar are essentially not subject to variation and thus provide little predictive power on their own.

\begin{figure*}
\centering
\includegraphics[width = 500pt]{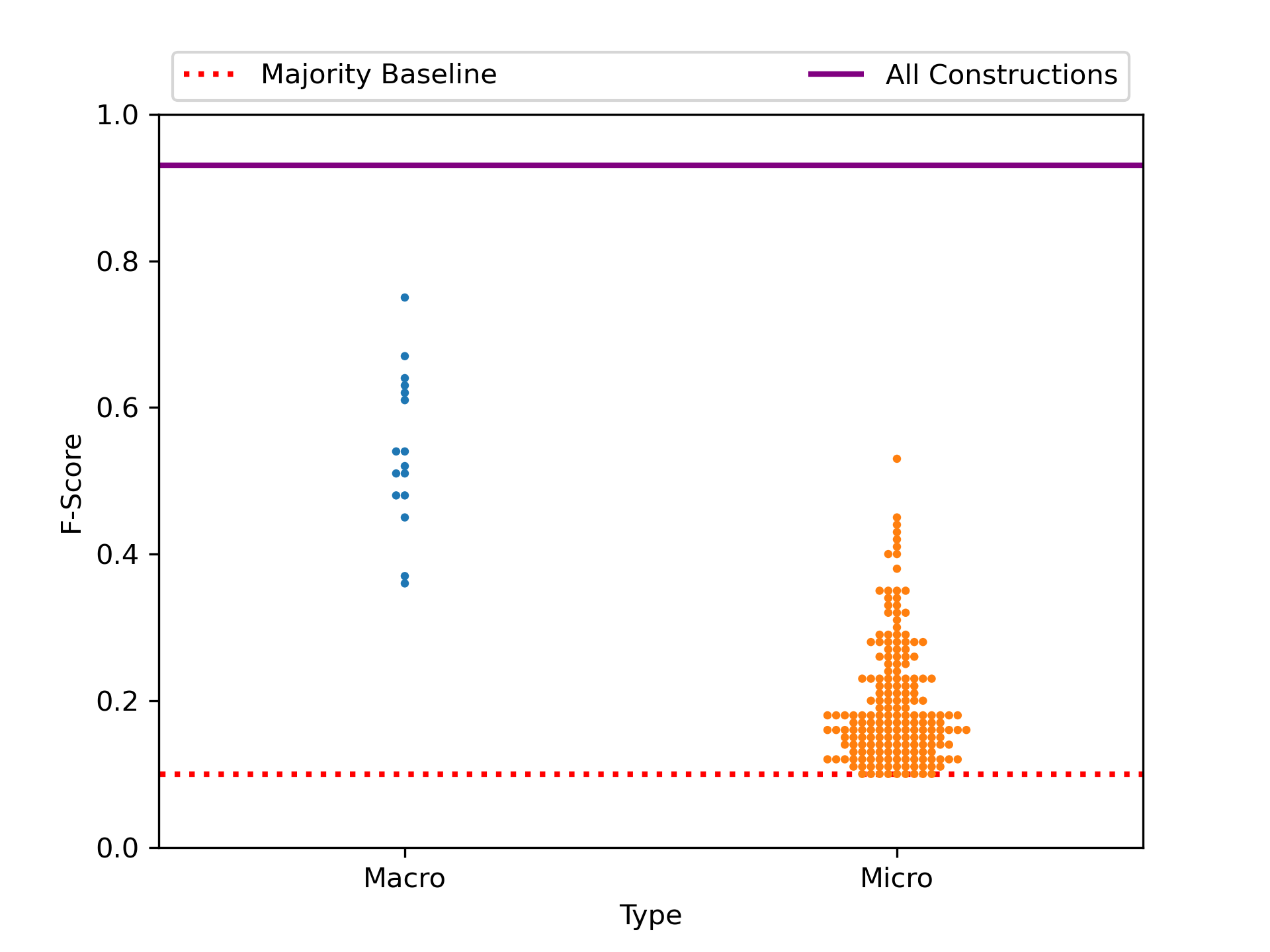}
\caption{Distribution of Classification Performance Across Sub-Sets of the Grammar, Regional Dialect with the Early-Stage Grammar}
\label{fig:results_region_syn}
\end{figure*}

The same type of graph is shown again in Figure \ref{fig:results_region_syn}, now for the early-stage grammar with only local constraints. This figure replicates the same findings: First, no individual node is able to capture the overall variation as well as the complete network. Second, there is a wide range across macro-clusters and micro-clusters in terms of that node's ability to characterize dialectal variation. What this tells us, in short, is that the complete picture of dialectal variation cannot be seen by observing small portions of the grammar in isolation. And yet this is precisely what almost all work on syntactic variation does.

The next question is which regional dialects are similar according to the representations learned in this classifier. This is shown in Table \ref{tab:errors_region} using the relative number of errors between regional dialects. Thus, for example, European English (UK and Ireland) and Oceanic English (Australia and New Zealand) account for a plurality of errors: 37.5\% in the late-stage grammar and 23.74\% in the early-stage  grammar. Thus, these are the most similar dialects because they are the most likely to be confused. Given the colonial spread of English, this is of course not surprising. What this table also shows is that the similarity between dialects differs to some degree depending on which part of the overall grammar we are observing: for example, North American English and Southeast Asian English are much more similar if we observe constructions from one part of the grammar (late-stage, with 10.83\% of the errors) than the other (early-stage, with only 5.93\% of the errors). This is important because it means that observing part of the grammar in isolation will not only inadequately characterize the amount of dialectal variation but will also provide different characterizations of the dialects relative to one another.

\begin{table*}
\centering
\begin{tabular}{|ScSr|ScSr|}
\hline
\multicolumn{2}{|Sc|}{\textbf{Late-Stage Grammar}} & \multicolumn{2}{Sc|}{\textbf{Early-Stage Grammar}} \\
\textit{Region Pairs} & \textit{\% of Errors} & \textit{Region Pairs} & \textit{\% of Errors} \\
\hline
Europe + Oceania & 37.50\% & Europe + Oceania & 23.74\% \\
America + Asia (Southeast) & 10.83\% & America + Oceania & 20.33\% \\
Africa (Southern) + Africa (Sub) & 10.83\% & Africa (Southern) + Africa (Sub) & 10.09\% \\
America + Oceania & 10.00\% & Asia (South) + Asia (Southeast) & 6.38\% \\
Asia (South) + Asia (Southeast) & 8.33\% & America + Asia (southeast) & 5.93\% \\
Africa (Southern) + Oceania & 5.83\% & Africa (Southern) + America & 5.34\% \\
Asia (Southeast) + Oceania & 5.00\% & Africa (Southern) + Asia (Southeast) & 5.19\% \\
\hline
  \end{tabular}
  \caption{Distribution of Errors in Dialect Classifier With Regional Dialects: Late-Stage Constructions (Left) and Early-Stage Constructions (Right)}
  \label{tab:errors_region}
\end{table*}

\begin{figure*}[!h]
\centering
\includegraphics[width = 500pt]{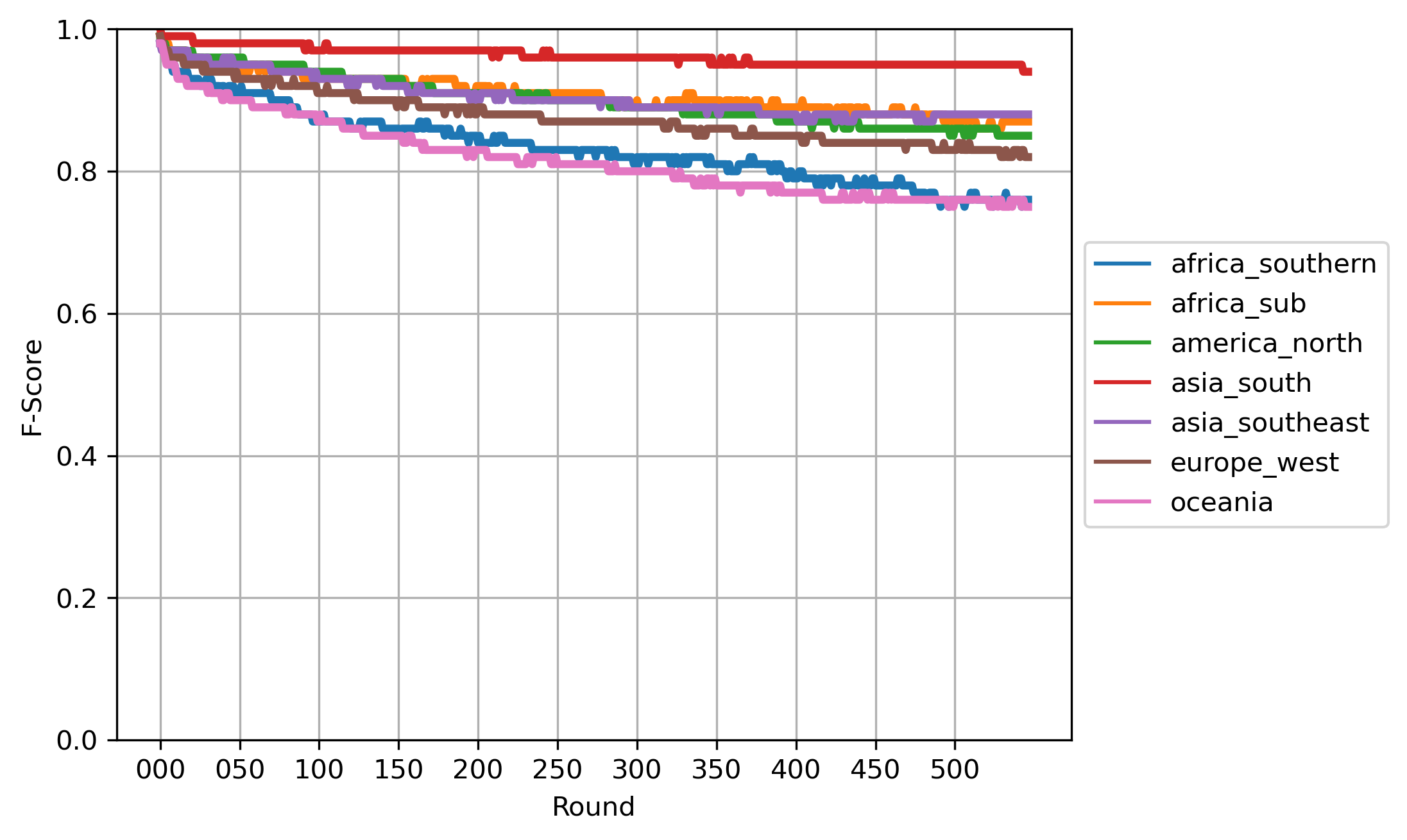}
\caption{Unmasking of Regional Dialects with the Late-Stage Grammar. Each 50 rounds removes approximately 2.5\% of the grammar, so that round 500 includes only 75\% of the original grammar.}
\label{fig:unmasking_region_full}
\end{figure*}

Before we investigate changes in dialect similarity across nodes within the grammar, we first evaluate the degree to which the classifiers depend on a small number of very predictive features. A possible confound with a prediction-based model of dialect is that a small number of constructions within each node could contribute almost all of the predictive power. This would distort our view of variation by implying that many constructions are in variation when, in fact, only a small number are. This is shown in Figure \ref{fig:unmasking_region_full} for the late-stage grammar using the unmasking method from forensic linguistics \citep{ksb07}. The basic idea behind this method is to train a linear classifier (in this case an SVM) over many rounds, removing the most predictive features for each dialect on each round. In the figure the f-score for each region is plotted on the y-axis over the course of 500 rounds of feature pruning, where each round removes the top features for each dialect. Overall, the prediction accuracy at round 500 represents the ability to characterize dialects when the top 25\% of construction have been removed. A robust classification model exhibits a gentle downward curve on an unmasking plot while a brittle model (depending on just a few constructions) would exhibit a sharp decline in the first several rounds. This figure, then, shows that the regional dialect model is quite robust overall. This supports our further analysis by showing that the predictive power is not constrained to only a few constructions within each node.

\begin{figure*}
\centering
\includegraphics[width = 500pt]{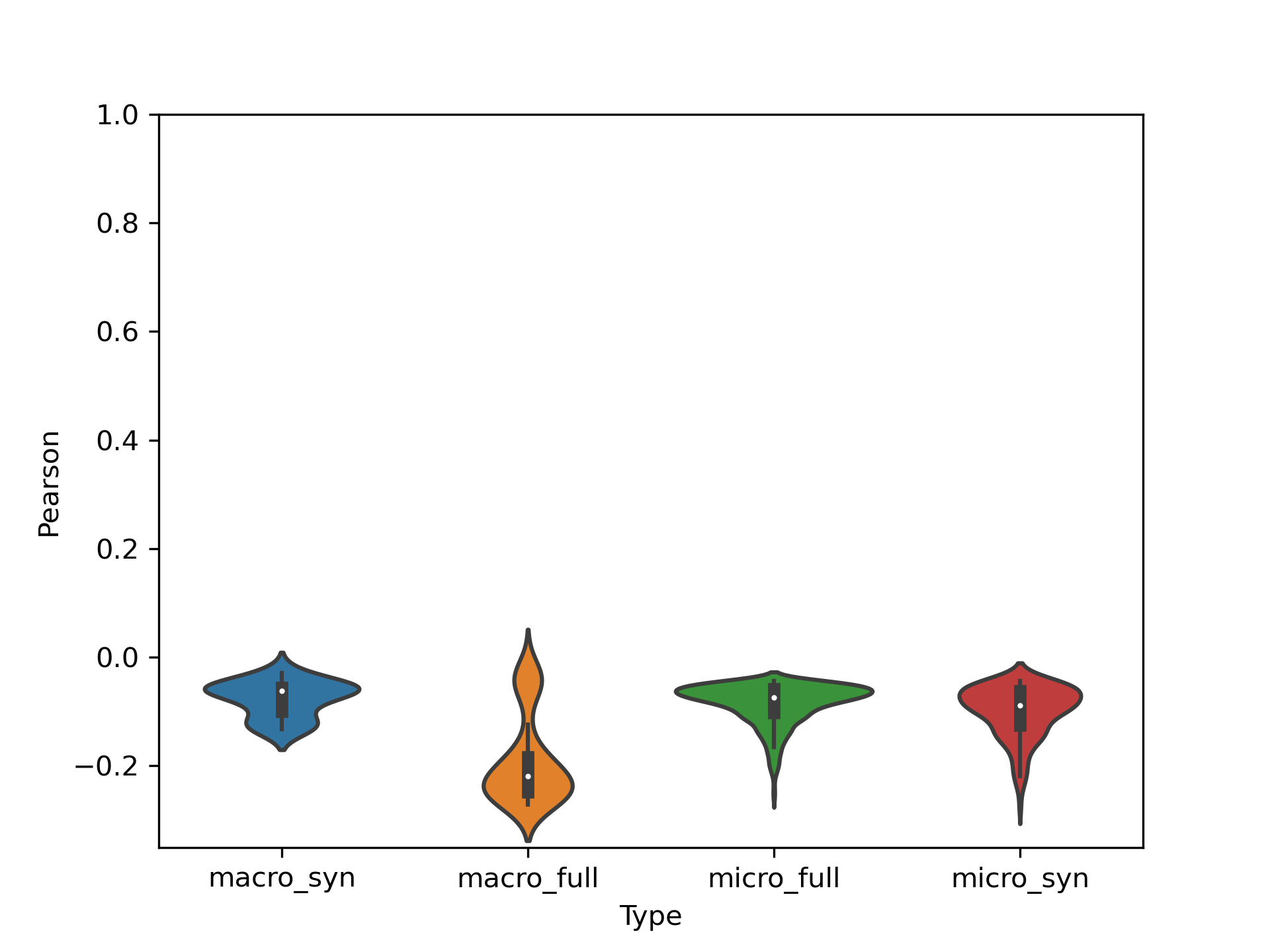}
\caption{Correlation of Error Distribution Between Late-Stage Grammar and Nodes within the Grammar for Regional Dialects. High Correlation indicates that the same dialects are similar in each model type while low correlation indicates that the relationships between dialects for a given component of the grammar differ from the late-stage grammar.}
\label{fig:error_distribution_region}
\end{figure*}

Given that these dialect models do not implicitly depend on only a few constructions, we further examine the variation in similarity relationships across nodes within the grammar in Figure \ref{fig:error_distribution_region}. This figure shows a violin plot of the distribution of correlation scores between (i) the similarity relations derived from a node within the grammar and (ii) the similarity relations derived from the high-performing late-stage grammar. The full late-stage grammar serves as a sort of ground-truth here because its prediction accuracies are nearly perfect. Thus, this measure can be seen as an indication of how close we would get to the actual relationships between dialects by observing only one node within the grammar (a macro-cluster or micro-cluster). The figure shows that, in all cases, there is no meaningful relationship between the two. If our goal is to characterize syntactic dialect variation as a whole, this means that studies which observe only isolated sets of features will not be able to make predictions beyond those features. In short, language is a complex adaptive system and observing only small pieces of this system is inadequate for capturing emergent interactions between constructions.

\subsection{National Dialects}

The previous section analyzed variation across regional dialects in order to show that variation is spread across many nodes within the grammar and that our view of syntactic variation depends heavily on the specific portions of the grammar under observation. This section continues this analysis with finer-grained country-level dialects in order to determine whether these same patterns emerge. The same methods of analysis are used, starting with the classification performance in Table \ref{tab:results_country}. As before, this is divided into late-stage constructions and early-stage  constructions. This table represents a single dialect model but for convenience it is divided into inner-circle varieties on the top and outer-circle varieties on the bottom \citep{k90}. As before, the late-stage grammar provides a better characterization of dialects than the early-stage grammar, 0.96 f-score vs. 0.83 f-score. While the late-stage grammar's performance is comparable to the regional-dialect model, the early-stage  model shows a sharp decline. Dialects with lower performance are less distinct and thus more influenced by other dialects; for example, we see that New Zealand English in both models is much less distinct or entrenched than other country-level dialects. It is thus more difficult to identify a sample of New Zealand English (c.f., \citealt{dunn-wong-2022-stability}).

\begin{table*}
\centering
\begin{tabular}{|Sl|ccc|ccc|}
\hline
~ & \multicolumn{3}{Sc|}{\textbf{Late-Stage Grammar}} & \multicolumn{3}{Sc|}{\textbf{Early-Stage Grammar}} \\
\textbf{Country} & \textit{Precision} & \textit{Recall} & \textit{F-Score} & \textit{Precision} & \textit{Recall} & \textit{F-Score} \\
\hline
\multicolumn{7}{|Sc|}{\textbf{Inner-Circle Varieties}} \\
\hline
Australia (au) & 0.94 & 0.95 & 0.94 & 0.81 & 0.83 & 0.82 \\
Canada (ca) & 0.98 & 0.98 & 0.98 & 0.71 & 0.73 & 0.72 \\
Ireland (ie) & 0.94 & 0.96 & 0.95 & 0.88 & 0.88 & 0.88 \\
New Zealand (nz) & 0.89 & 0.84 & 0.86 & 0.64 & 0.59 & 0.61 \\
United Kingdom (uk) & 0.96 & 0.96 & 0.96 & 0.85 & 0.86 & 0.86 \\
United States (us) & 0.97 & 1.00 & 0.98 & 0.87 & 0.87 & 0.87 \\
south Africa (za) & 0.99 & 1.00 & 0.99 & 0.80 & 0.81 & 0.80 \\
\hline
\multicolumn{7}{|Sc|}{\textbf{Outer-Circle Varieties}} \\
\hline
Bangladesh (bd) & 0.97 & 0.89 & 0.93 & 0.76 & 0.69 & 0.72 \\
Indonesian (id) & 1.00 & 0.84 & 0.91 & 0.57 & 0.49 & 0.53 \\
India (in) & 0.97 & 0.98 & 0.97 & 0.94 & 0.96 & 0.95 \\
Kenya (ke) & 0.98 & 0.99 & 0.98 & 0.92 & 0.87 & 0.89 \\
Malaysia (my) & 0.96 & 0.98 & 0.97 & 0.82 & 0.74 & 0.78 \\
Nigeria (ng) & 0.98 & 1.00 & 0.99 & 0.90 & 0.83 & 0.87 \\
Philippines (ph) & 0.98 & 0.98 & 0.98 & 0.83 & 0.87 & 0.85 \\
Pakistan (pk) & 0.94 & 0.94 & 0.94 & 0.84 & 0.89 & 0.87 \\
Zimbabwe (zw) & 0.95 & 0.91 & 0.93 & 0.48 & 0.53 & 0.51 \\
\hline
\textbf{Weighted Avg} & \textbf{0.96} & \textbf{0.96} & \textbf{0.96} & \textbf{0.83} & \textbf{0.83} & \textbf{0.83} \\
\hline
  \end{tabular}
  \caption{Performance of Dialect Classifier With National Dialects: Late-Stage Constructions (Left) and Early-Stage Constructions (Right). This is a single model for all dialects, with Inner-Circle Varieties shown at the top and Outer-Circle Varieties at the bottom.}
  \label{tab:results_country}
\end{table*}

Continuing with the error analysis in Table \ref{tab:errors_country}, we see that almost 20\% of the errors in the late-stage model are confusions between New Zealand and Australia and over 12\% between Ireland and the UK (which includes Northern Ireland here). As before, the similarity between dialects derived from classification errors reflects the similarity between these countries in geographic, historical, and social terms.

\begin{table*}[!h]
\centering
\begin{tabular}{|ScSr|ScSr|}
\hline
\multicolumn{2}{|Sc|}{\textbf{Late-Stage Grammar}} & \multicolumn{2}{Sc|}{\textbf{Early-Stage Grammar}} \\
\textit{Country Pairs} & \textit{\% of Errors} & \textit{Country Pairs} & \textit{\% of Errors} \\
\hline
Australia + New Zealand & 19.85\% & Canada + United States & 11.76\% \\
Ireland + United Kingdom & 12.11\% & Australia + New Zealand & 6.77\% \\
India + Pakistan & 11.86\% & Bangladesh + Pakistan & 5.58\% \\
Bangladesh + Pakistan & 7.47\% & Australia + Canada & 5.34\% \\
Australia + United Kingdom & 5.93\% & Ireland + United Kingdom & 5.29\% \\
Canada + United States & 4.12\% & Australia + United Kingdom & 4.69\% \\
\hline
  \end{tabular}
  \caption{Distribution of Errors in Dialect Classifier With National Dialects: Late-Stage Constructions (Left) and Early-Stage Constructions (Right)}
  \label{tab:errors_country}
\end{table*}

\begin{figure*}
\centering
\includegraphics[width = 500pt]{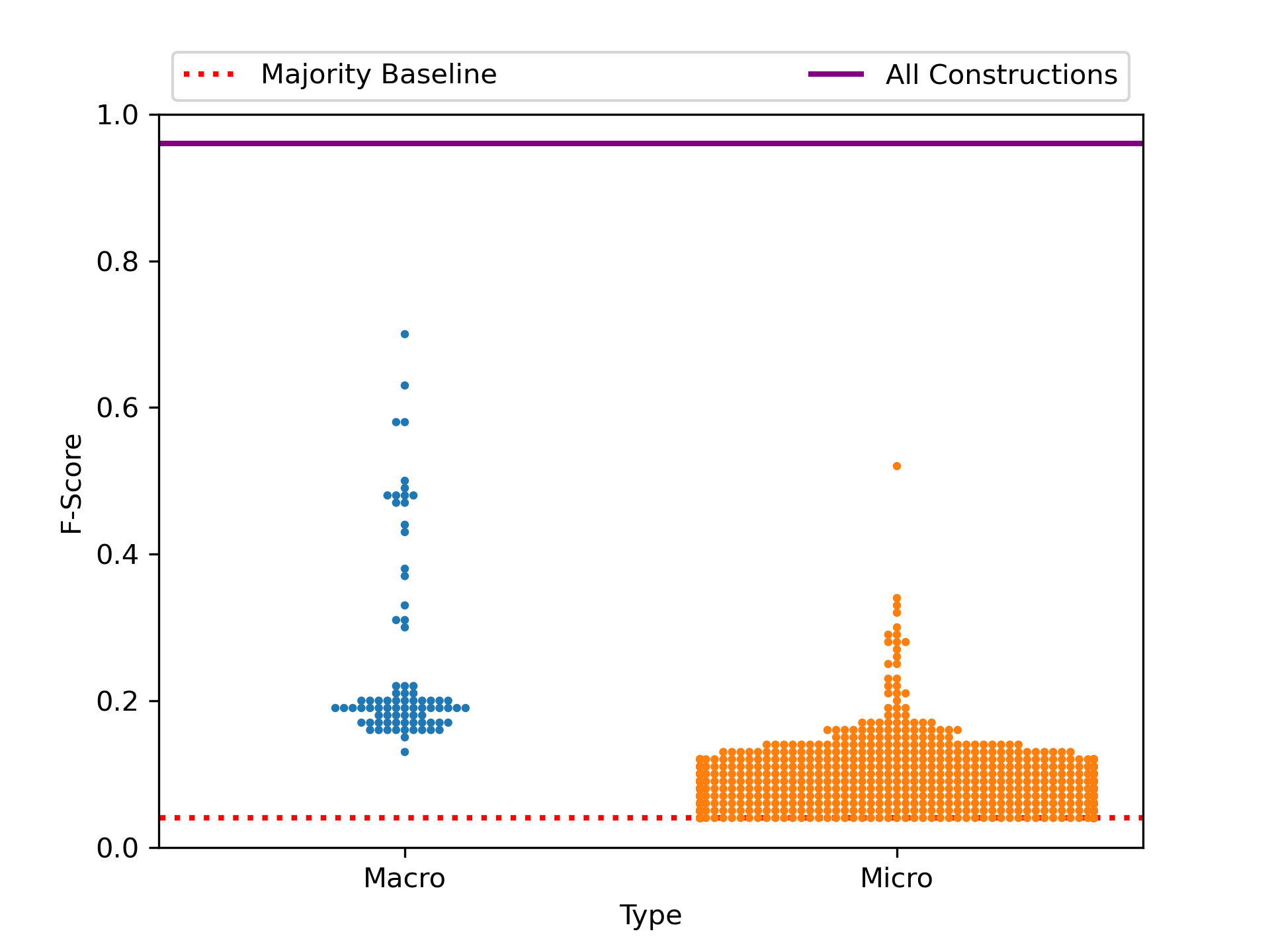}
\caption{Distribution of Classification Performance Across Sub-Sets of the Grammar, National Dialect with the Late-Stage Grammar}
\label{fig:results_country_full}
\end{figure*}

The distribution of predictive accuracy across nodes within the grammar is shown in Figure \ref{fig:results_country_full}. Once again we see that a few portions of the grammar have relatively high performance on their own (capturing upwards of 70\% of the predictive power), but that no individual nodes perform as well as the grammar as a whole. As before this means that variation is spread throughout the grammar and that interactions between nodes is important for characterizing syntactic variation at the country level.

\begin{figure*}[!h]
\centering
\includegraphics[width = 500pt]{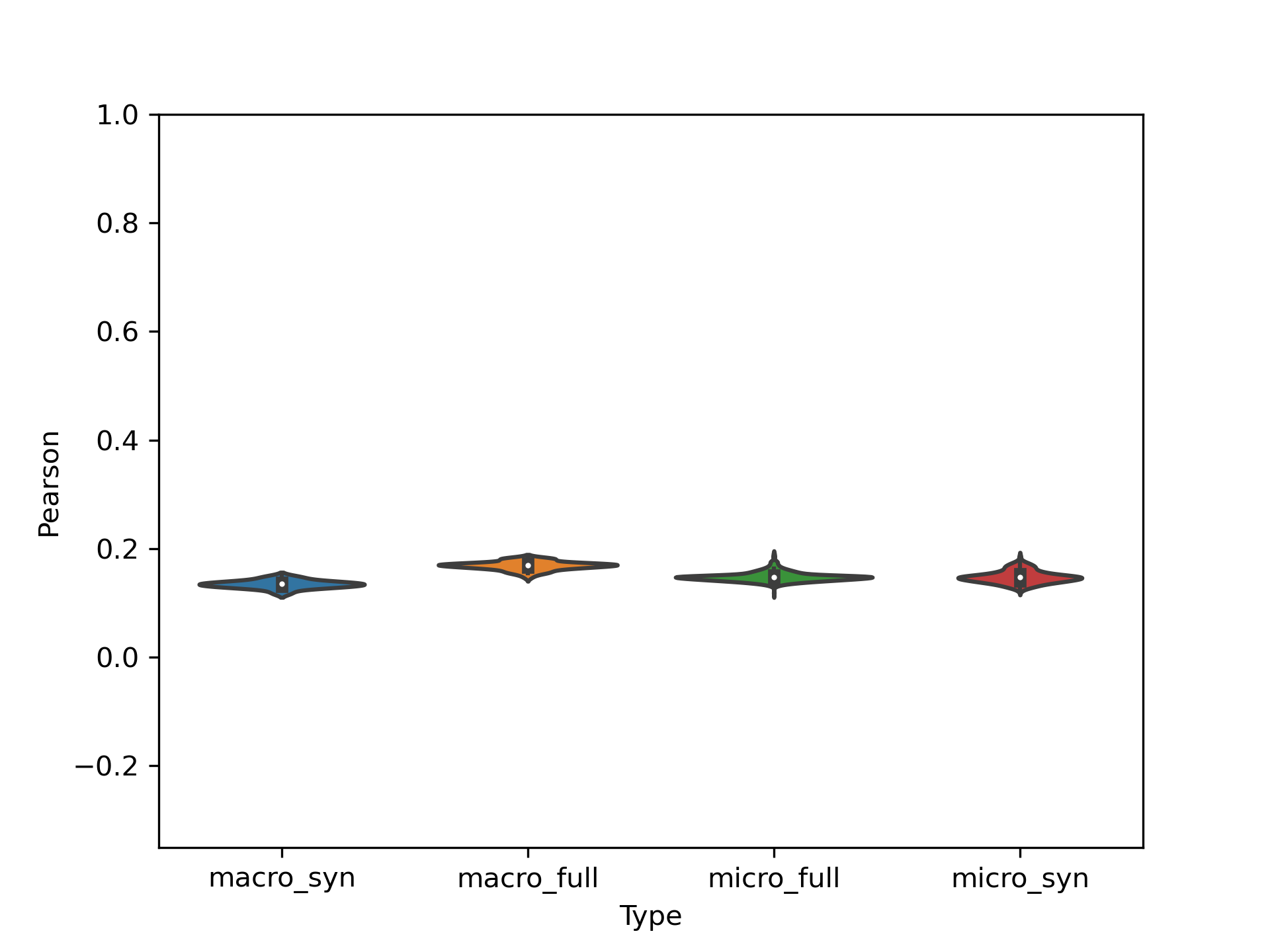}
\caption{Correlation of Error Distribution Between Late-Stage Grammar and Nodes within the Grammar for National Dialects. High Correlation indicates that the same dialects are similar in each model type.}
\label{fig:error_distribution_country}
\end{figure*}

Continuing with the distribution of similarity values across sub-sets of the grammar, Figure \ref{fig:error_distribution_country} shows the correlation between macro- and micro-clusters and the ground-truth of the high-performing late-stage grammar. Here the correlation is above chance but remains quite low (below 0.2). This again indicates that different nodes of the grammar provide different views of the similarity between country-level dialects, agreeing with the different errors ranks shown in Table \ref{tab:errors_country}. For instance, New Zealand English might be close to Australian in one part of the grammar but close to UK English in another. As a high-dimensional and complex system, grammatical variation must be viewed from the perspective of the entire grammar.




National dialects are more fine-grained than regional dialects and also have a higher number of categories (from 7 to 16), making the classification task more difficult because it must now distinguish between similar dialects (like American and Canadian English). Given the importance of the nation-state in modern mobility (i.e., in terms of ease of travel and immigration), these country-level dialects are more reflective of the underlying population than cross-country aggregations. In other words, the social network of countries is more coherent than that of larger regions because it is the nation which creates boundaries around human mobility. Since we are viewing both the grammar and the population of speakers of complex networks, it is important to go further and analyze local populations in the form of dialect areas within countries.

\subsection{Local Dialects}

This section takes a closer look at dialectal variation by modelling the differences between local populations within the same region. As discussed above, data is collected from areas around airports as a consistent proxy for urban areas. These urban areas are then clustered into local groups using spatial but not linguistic information. Thus, as shown in the map of North American dialect areas in Figure \ref{fig:north_america}, a country-level dialect is divided into many smaller areas and then the differences between these local dialects are modelled using the same classification methods. This section examines the North American, European, and South Asian models in more detail. The full results are available in the supplementary material.

\subsubsection{North America}

We start with North American dialects, mapped in Figure \ref{fig:north_america} and listed by name in Table \ref{tab:results_north_america}, again with the late-stage grammar on the left and the early-stage grammar on the right. The distinction between same-country dialects is much smaller which means that the prediction task is much harder: here the f-scores drop to 0.69 and 0.40 (both of which remain much higher than the majority baseline). A classification model thus also provides a measure of the amount of dialectal variation between varieties: here there is much variation overall because the local populations are more similar. Local dialects with lower performance are again less distinguishable and thus less unique in their grammar: for example, Midwestern and Plains American English have the lowest f-scores at 0.57 and 0.52, respectively.

\begin{table*}
\centering
\begin{tabular}{|Sl|ccc|ccc|}
\hline
~ & \multicolumn{3}{Sc|}{\textbf{Late-Stage Grammar}} & \multicolumn{3}{Sc|}{\textbf{Early-Stage Grammar}} \\
\textbf{Local Area} & \textit{Precision} & \textit{Recall} & \textit{F-Score} & \textit{Precision} & \textit{Recall} & \textit{F-Score} \\
\hline
Western Canada (CA-1) & 0.82 & 0.83 & 0.82 & 0.64 & 0.58 & 0.61 \\
Ontario (CA-2) & 0.65 & 0.69 & 0.67 & 0.43 & 0.45 & 0.44 \\
Quebec (CA-3) & 0.65 & 0.61 & 0.63 & 0.35 & 0.33 & 0.34 \\
Nova Scotia (CA-4) & 0.57 & 0.56 & 0.57 & 0.19 & 0.24 & 0.21 \\
\hline
Midwest (US-1) & 0.57 & 0.57 & 0.57 & 0.32 & 0.26 & 0.29 \\
Central California (US-2) & 0.80 & 0.81 & 0.81 & 0.35 & 0.34 & 0.35 \\
Texas (US-3) & 0.66 & 0.67 & 0.67 & 0.30 & 0.32 & 0.31 \\
Southern (US-4) & 0.68 & 0.56 & 0.61 & 0.26 & 0.27 & 0.27 \\
Florida (US-5) & 0.77 & 0.78 & 0.78 & 0.23 & 0.30 & 0.26 \\
West Texas (US-6) & 0.60 & 0.55 & 0.57 & 0.25 & 0.24 & 0.25 \\
East Coast (US-7) & 0.77 & 0.82 & 0.79 & 0.57 & 0.56 & 0.56 \\
Plains (US-8) & 0.55 & 0.49 & 0.52 & 0.11 & 0.14 & 0.12 \\
Southwestern (US-9) & 0.66 & 0.71 & 0.68 & 0.39 & 0.41 & 0.40 \\
\hline
\textbf{Weighted Avg} & \textbf{0.69} & \textbf{0.69} & \textbf{0.69} & \textbf{0.40} & \textbf{0.40} & \textbf{0.40} \\
\hline
  \end{tabular}
  \caption{Performance of Dialect Classifier With Local Dialects: Late-Stage Constructions (Left) and Early-Stage Constructions (Right). This is a single model for all dialects, with American Varieties shown at the bottom and Canadian at the top.}
  \label{tab:results_north_america}
\end{table*}

The distribution of classification errors is shown in Table \ref{tab:errors_north_america}. All major sources of errors in the late-stage grammar are within the same country and close geographically. For example, over 10\% of the errors are between Ontario and Quebec. The distribution of errors also shows that the lower performance of the Midwest and Plains areas is a result of their similarity to one another and, in the case of the Midwest, to the Eastern dialect. Thus, the types of errors here are as important as the number of errors.

\begin{table*}[!h]
\centering
\begin{tabular}{|ScSr|ScSr|}
\hline
\multicolumn{2}{|Sc|}{\textbf{Late-Stage Grammar}} & \multicolumn{2}{Sc|}{\textbf{Early-Stage Grammar}} \\
\textit{Country Pairs} & \textit{\% of Errors} & \textit{Country Pairs} & \textit{\% of Errors} \\
\hline
Ontario (CA-2) + Quebec (CA-3) & 10.31\% & Western (CA-1) + Quebec (CA-3) & 6.09\% \\
Western (CA-1) + Ontario (CA-2) & 8.05\% & Ontario (CA-2) + Quebec (CA-3) & 5.12\% \\
Midwest (US-1) + East (US-7) & 7.73\% & Ontario (CA-2) + East (US-7) & 4.47\% \\
Western (CA-1) + Quebec (CA-3) & 6.44\% & Western (CA-1) + Ontario (CA-2) & 4.39\% \\
Midwest (US-1) + Plains (US-8) & 5.48\% & Midwest (US-1) + Plains (US-8) & 3.82\% \\
\hline
  \end{tabular}
  \caption{Distribution of Errors in Dialect Classifier With North American Dialects: Late-Stage Constructions (Left) and Early-Stage Constructions (Right)}
  \label{tab:errors_north_america}
\end{table*}

\begin{figure*}
\centering
\includegraphics[width = 500pt]{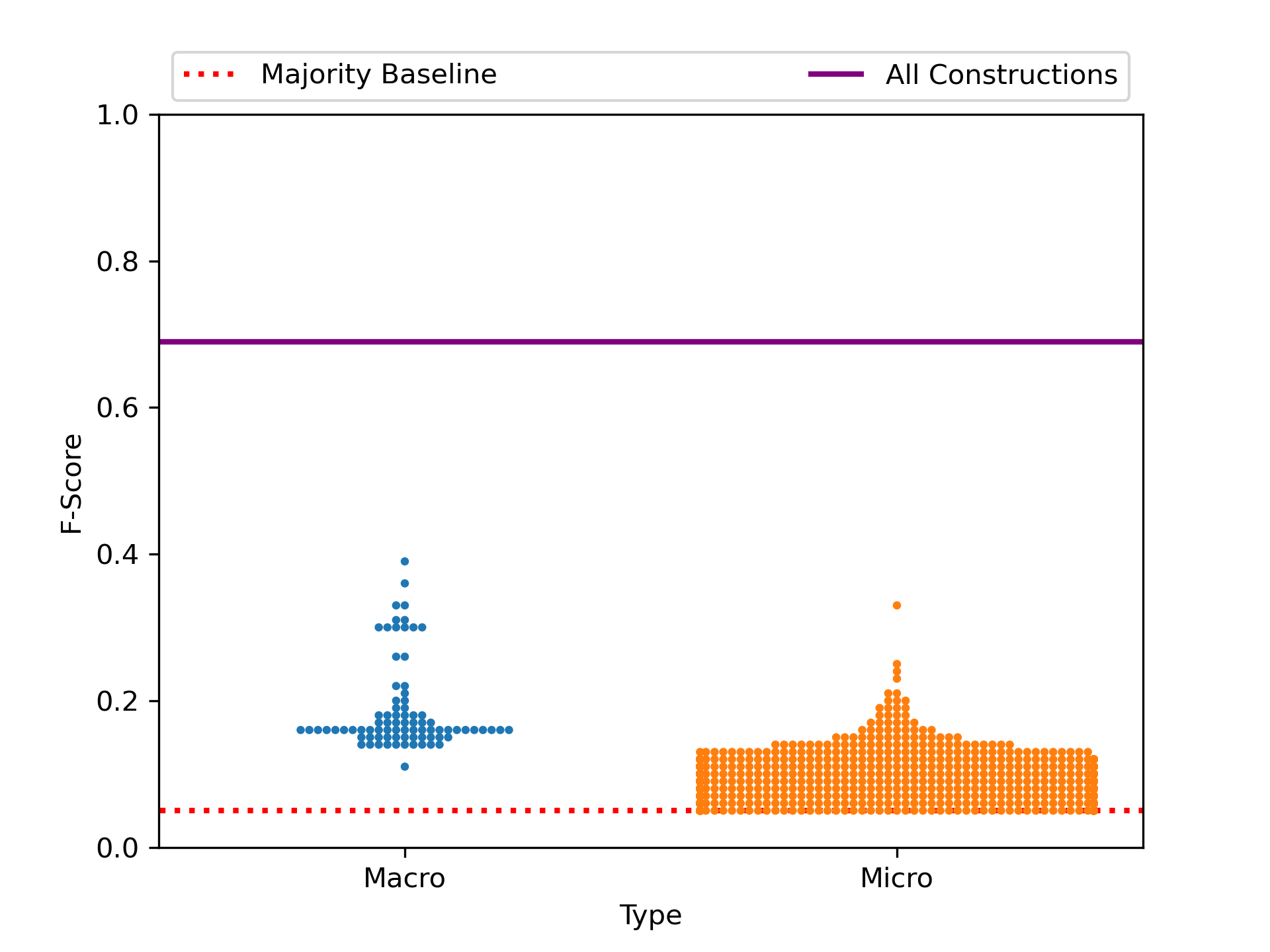}
\caption{Distribution of Performance across the Late-Stage grammar, North America}
\label{fig:results_north_america_full}
\end{figure*}

\begin{figure*}[!h]
\centering
\includegraphics[width = 500pt]{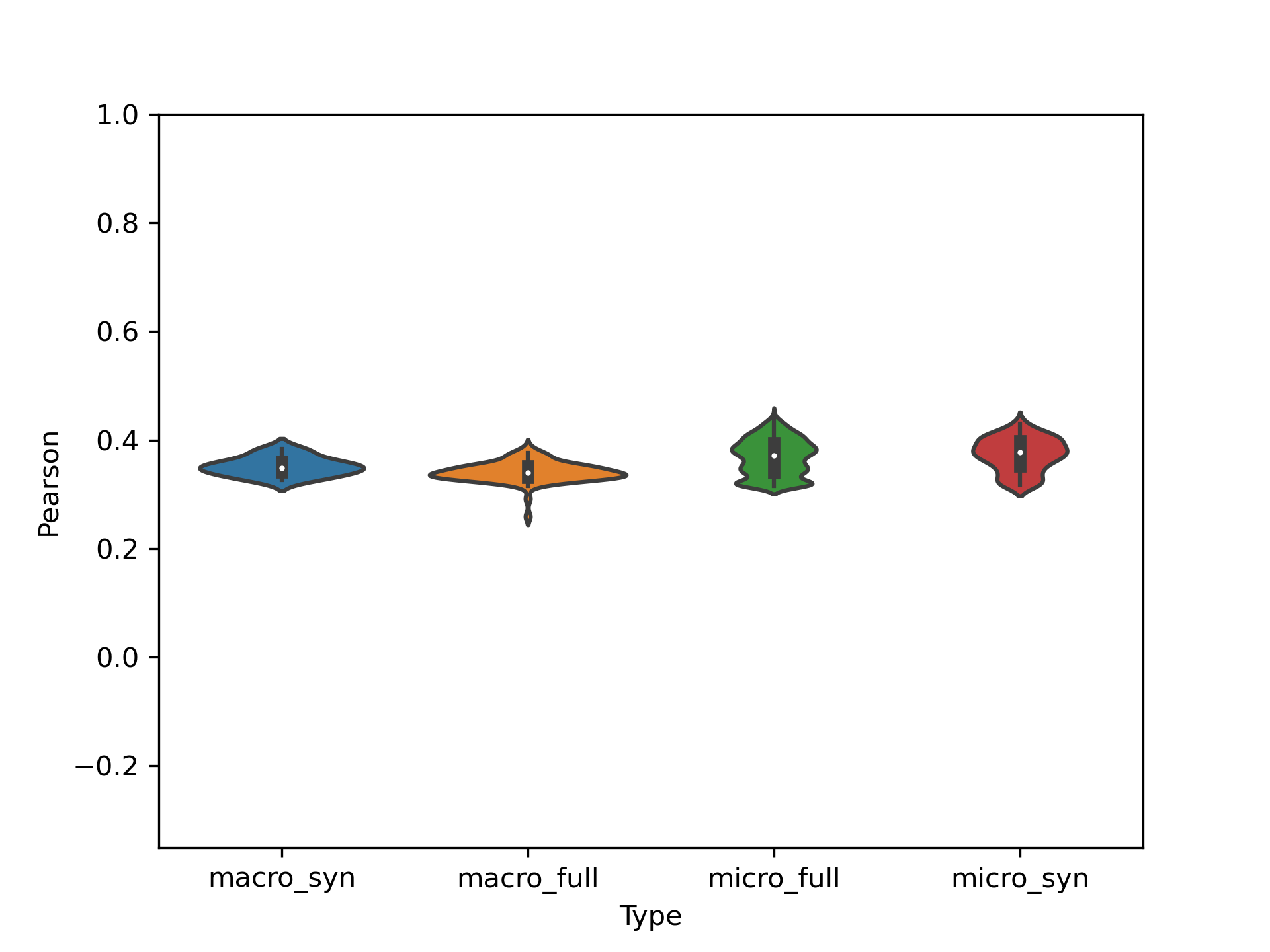}
\caption{Correlation of Error Distribution Between Late-Stage Grammar and Nodes within the Grammar for North American Dialects. High Correlation indicates that the same dialects are similar in each model.}
\label{fig:error_distribution_north_america}
\end{figure*}

The distribution of nodes within the grammar in terms of classification performance is shown in Figure \ref{fig:results_north_america_full}; here most micro-clusters have minimal predictive adequacy on their own but some macro-clusters retain meaningful predictive power. The correlation between similarity relations across nodes within the grammar is visualized in Figure \ref{fig:error_distribution_north_america}. While the overall prediction accuracy is lower, the similarity relationships are more stable. In other words, the representation of which dialects have similar grammars is less dependent here on the sub-set of the grammar being observed.


\subsubsection{Europe}

\begin{figure*}
\centering
\fbox{\includegraphics[width = 500pt]{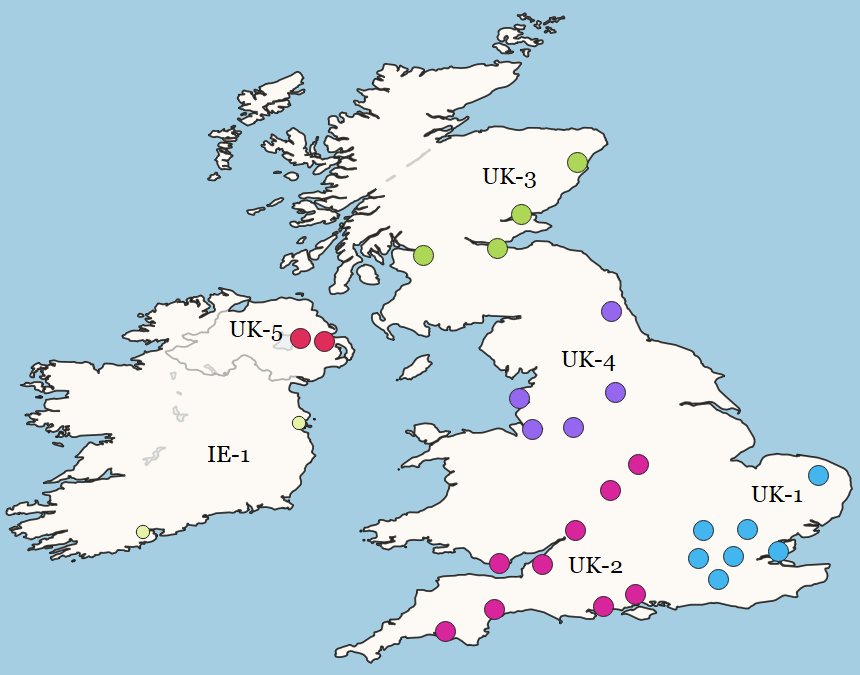}}
\caption{Distribution of Local Dialects in Europe}
\label{fig:europe}
\end{figure*}

The map of European dialects (within the UK and Ireland) is shown in Figure \ref{fig:europe}. As before, these areas are formed using spatial clustering with the dialect classifier then used to characterize the syntactic differences between dialect areas. As listed in Table \ref{tab:results_europe}, there are six local dialects with an f-score of 0.77 (late-stage grammar) and 0.64 (early-stage grammar). The range of f-scores across dialect areas is wider than in the North American model, with a high of 0.94 (Ireland) and a low of 0.40 (Scotland). This is partly driven by the number of samples per dialect, with Scotland particularly under-represented. This means, then, that the characterizations this model makes about Irish English are much more reliable than the characterizations made about Scottish English.

\begin{table*}
\centering
\begin{tabular}{|Sl|ccc|ccc|}
\hline
~ & \multicolumn{3}{Sc|}{\textbf{Late-Stage Grammar}} & \multicolumn{3}{Sc|}{\textbf{Early-Stage Grammar}} \\
\textbf{Local Area} & \textit{Precision} & \textit{Recall} & \textit{F-Score} & \textit{Precision} & \textit{Recall} & \textit{F-Score} \\
\hline
Ireland (IE-1) & 0.92 & 0.97 & 0.94 & 0.84 & 0.88 & 0.86 \\
\hline
Southeast England (UK-1) & 0.70 & 0.74 & 0.72 & 0.59 & 0.61 & 0.60 \\
Southwest England (UK-2) & 0.60 & 0.58 & 0.59 & 0.43 & 0.43 & 0.43 \\
Scotland (UK-3) & 0.54 & 0.32 & 0.40 & 0.23 & 0.23 & 0.23 \\
Northern England (UK-4) & 0.75 & 0.71 & 0.73 & 0.59 & 0.57 & 0.58 \\
Northern Ireland (UK-5) & 0.75 & 0.46 & 0.57 & 0.38 & 0.23 & 0.29 \\
\hline
\textbf{Weighted Avg} & \textbf{0.77} & \textbf{0.78} & \textbf{0.77} & \textbf{0.64} & \textbf{0.65} & \textbf{0.64} \\
\hline
  \end{tabular}
  \caption{Performance of Dialect Classifier With Local Dialects: Late-Stage Constructions (Left) and Early-Stage Constructions (Right). This is a single model for all local dialects.}
  \label{tab:results_europe}
\end{table*}

\begin{figure*}
\centering
\includegraphics[width = 500pt]{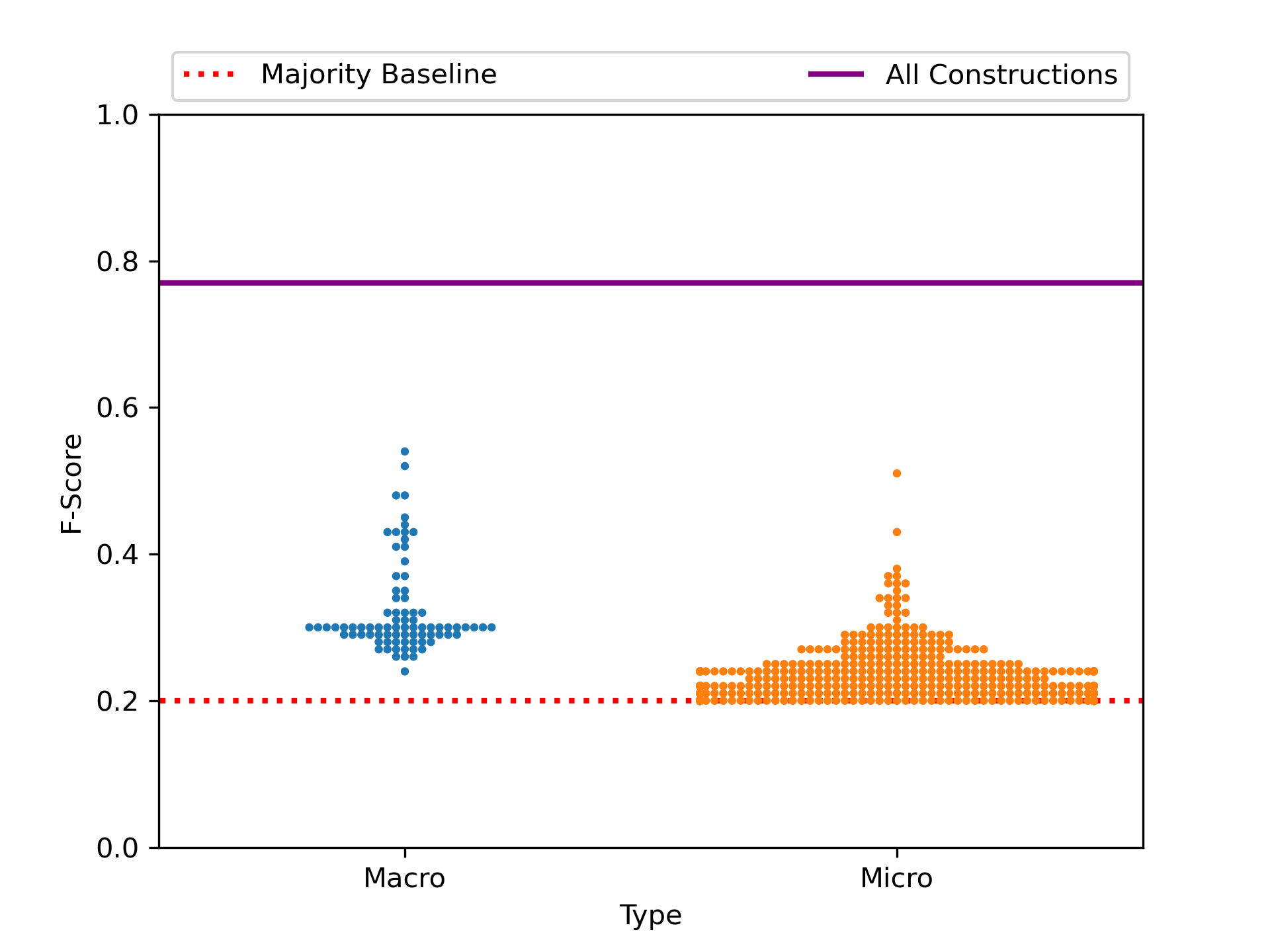}
\caption{Distribution of Performance across the Late-Stage grammar, Europe}
\label{fig:results_europe_full}
\end{figure*}

The distribution of performance across nodes in the grammar is shown in Figure \ref{fig:results_europe_full}. The figure shows that many nodes are meaningfully above the majority baseline in terms of predictive power but all fall short of the grammar as a whole. In fact, some micro-clusters have nearly as much power as the best macro-clusters. This indicates that the variation in UK and Irish English is spread throughout many individual parts of the grammar, a fact that would be overlooked if we had focused instead on a few constructions in isolation.




The main take-away from the European model, then, is once again that an accurate characterization of the grammatical differences between these local dialects requires access to the entire grammar. Focusing on smaller nodes within the grammar does not provide strong predictive power. As before, this has two implications: first, that variation is distributed across the grammar and, second, that emergent relationships between constructions are essential for depicting syntactic variation in the aggregate.

\subsubsection{South Asia}

\begin{figure*}
\centering
\fbox{\includegraphics[width = 500pt]{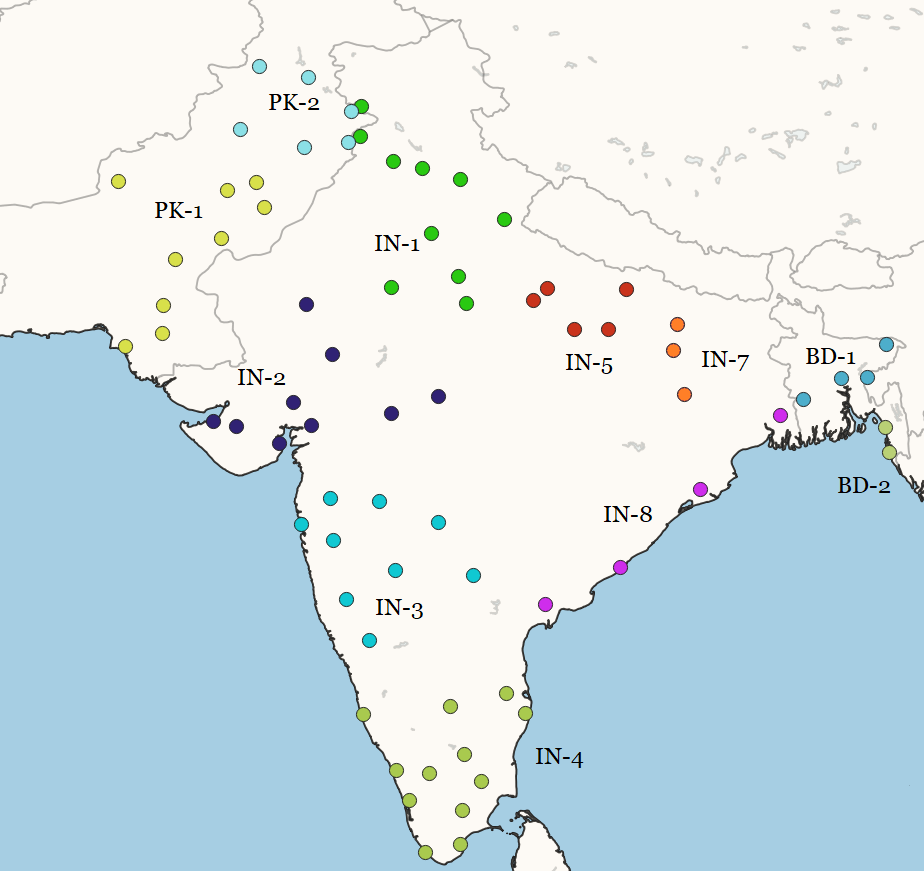}}
\caption{Distribution of Local Dialects in South Asia}
\label{fig:south_asia}
\end{figure*}

The final model of local dialects we will investigate is the outer-circle varieties from South Asia. These differ from the inner-circle dialects in a number of ways. In the first case, these speakers use English for different purposes than inner-circle speakers, almost always being highly multi-lingual with other languages used in the home (while, for example, North American speakers of English are often monolingual). In the second case, there is a socio-economic skew in the sense that higher class speakers are more likely to use English. The impact of socio-economic status is even more important when we consider that this data is collected from digital sources (tweets) and that the population of Twitter users in South Asia is less representative of the larger population than it is in North America and Europe. Thus, we use South Asia as a case-study in variation within outer-circle dialects. The map of local dialect areas is shown in Figure \ref{fig:south_asia}. This encompasses three countries: India, Pakistan, and Bangladesh.

The performance of the dialect model by local dialect is shown in Table \ref{tab:results_south_asia}, divided by country and by grammar type. The overall f-scores here are more similar to the North American model, at 0.68 and 0.56. As before, it is more difficult to distinguish between these local dialects because they are more similar overall. There is also a larger range of f-scores than before: the Bangladesh dialects are the highest performing, at 0.97 and 0.82. But within India, with many adjacent dialect areas, the f-scores fall as low as 0.14 (Uttar Pradesh) and 0.24 (Bihar). One conclusion to be drawn from these low values for specific local dialects is that the areas posited by the spatial clustering step do not contain unique and predictable dialectal variants. This is a clear case where a joint spatial/linguistic approach to forming local areas would be preferable. In other words, the spatial organization suggests a boundary which the linguistic features do not support.

\begin{table*}
\centering
\begin{tabular}{|Sl|ccc|ccc|}
\hline
~ & \multicolumn{3}{Sc|}{\textbf{Late-Stage Grammar}} & \multicolumn{3}{Sc|}{\textbf{Early-Stage Grammar}} \\
\textbf{Local Area} & \textit{Precision} & \textit{Recall} & \textit{F-Score} & \textit{Precision} & \textit{Recall} & \textit{F-Score} \\
\hline
Dhaka (BD-1) & 0.96 & 0.98 & 0.97 & 0.88 & 0.89 & 0.89 \\
Chattogram (BD-2) & 0.87 & 0.79 & 0.82 & 0.61 & 0.58 & 0.60 \\
\hline
New Delhi (IN-1) & 0.56 & 0.64 & 0.60 & 0.51 & 0.53 & 0.52 \\
Gujarat (IN-2) & 0.55 & 0.55 & 0.55 & 0.33 & 0.39 & 0.36 \\
Maharashtra (IN-3) & 0.67 & 0.71 & 0.69 & 0.54 & 0.53 & 0.54 \\
Tamil Nadu (IN-4) & 0.72 & 0.71 & 0.71 & 0.55 & 0.58 & 0.56 \\
Uttar Pradesh (IN-5) & 0.17 & 0.11 & 0.14 & 0.42 & 0.22 & 0.29 \\
Bihar (IN-7) & 0.39 & 0.17 & 0.24 & 0.28 & 0.17 & 0.21 \\
Andhra Pradesh + Odisha  (IN-8) & 0.69 & 0.61 & 0.65 & 0.54 & 0.49 & 0.51 \\
\hline
Southern (PK-1) & 0.56 & 0.51 & 0.53 & 0.43 & 0.40 & 0.42 \\
Northern (PK-2) & 0.72 & 0.75 & 0.74 & 0.63 & 0.66 & 0.65 \\
\hline
\textbf{Weighted Avg} & \textbf{0.68} & \textbf{0.68} & \textbf{0.68} & \textbf{0.56} & \textbf{0.56} & \textbf{0.56} \\
\hline
  \end{tabular}
  \caption{Performance of Dialect Classifier With Local Dialects in South Asia: Late-Stage Constructions (Left) and Early-Stage Constructions (Right). This is a single model for all local dialects.}
  \label{tab:results_south_asia}
\end{table*}

\begin{figure*}[!h]
\centering
\includegraphics[width = 500pt]{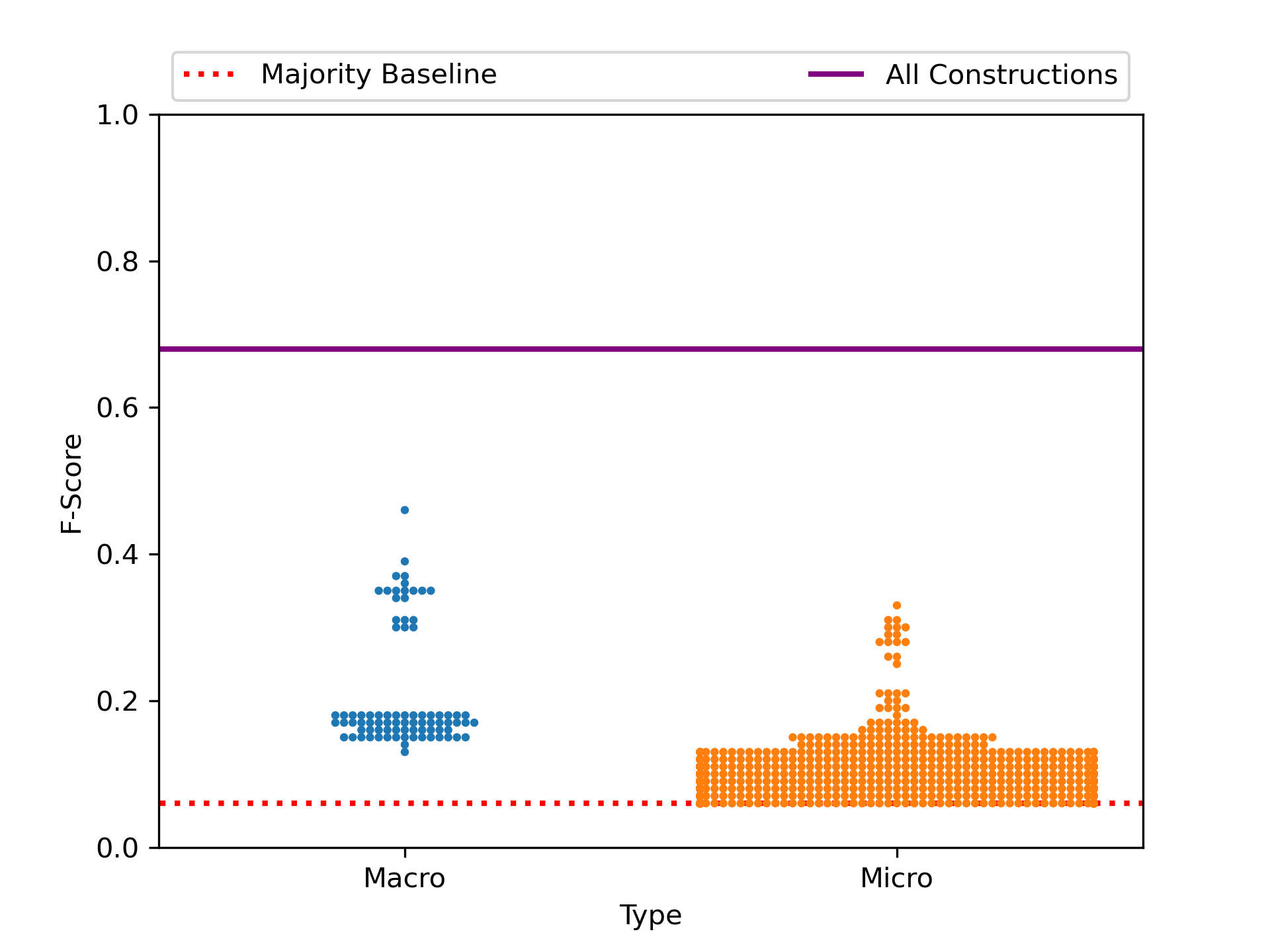}
\caption{Distribution of Performance across the Late-Stage grammar, South Asia}
\label{fig:results_asia_south_full}
\end{figure*}

The distribution of predictive power across the grammar is shown in Figure \ref{fig:results_asia_south_full}. As before, predictive power is distributed across nodes within the grammar and no single node nears the performance of the grammar as a whole. This means that interactions between constructions are an important part of variation and that the variability in different nodes is not simply redundant information.




\section{Discussion and Conclusions}

The main focus of this paper has been to investigate the assumption, common to most studies of grammatical variation across dialects, that individual constructions within the grammar can be viewed in isolation. The results show conclusively that this is simply not the case: language is a complex adaptive system and observing variation in an arbitrary sub-set of that system leads to incomplete models. The paper has repeated the same experimental paradigm at three levels of spatial granularity (regional, national, and local dialects) in order to show that the inadequacy of individual nodes within the grammar is a consistent phenomenon. Across all levels of spatial granularity, these experiments tell us three important things about variation within a complex system:

\textbf{Finding 1: No individual node within the grammar captures dialectal variation as accurately as the grammar as a whole.} The basic approach taken here is to first cluster constructions within the grammar into macro- and micro-clusters using similarity relations between the constructions themselves (thus, independently of their patterns of variation). As shown in Figure \ref{fig:grammar}, this leads to 1,941 micro-clusters in the late-stage grammar and 191 in the early-stage grammar. Even larger nodes are captured by macro-clusters, including 81 in the late-stage grammar and 16 in the early-stage grammar. The dialect classification experiments are repeated with each of these clusters alone, with the results shown in Figures \ref{fig:results_region_full}, \ref{fig:results_region_syn}, \ref{fig:results_country_full}, \ref{fig:results_north_america_full}, \ref{fig:results_europe_full}, and \ref{fig:results_asia_south_full}. Additional figures are available in the supplementary material. In each case the grammar as a whole performs much better than any individual sub-set or node within the grammar.

Why? Our interpretation is that there are emergent interactions between constructions in different nodes across the grammar. This means, for instance, that the use of plural nouns interacts with the use of certain adpositional phrases which interacts in turn with the use of phrasal verbs. Language is a complex adaptive system and the huge number of such minor interactions provides information about syntactic variation which is not redundant with the information contained in local nodes within the grammar. In other words, a significant part of syntactic variation is contained within emergent interactions which cannot be localized to specific families of constructions.

\textbf{Finding 2: Individual nodes within the grammar vary widely in the degree to which they are subject to dialectal variation.} Following from the above finding, the degree to which individual macro- and micro-clusters are able to predict dialect membership represents the degree to which they themselves are subject to spatial variation. Thus, the results shown in Figures \ref{fig:results_region_full}, \ref{fig:results_region_syn}, \ref{fig:results_country_full}, \ref{fig:results_north_america_full}, \ref{fig:results_europe_full}, and \ref{fig:results_asia_south_full} also mean that not all clusters of constructions are subject to the same level of variation.

Why? Our interpretation is that dialect variation is distributed across the grammar but in uneven portions. For instance, the unmasking experiment (Figure \ref{fig:unmasking_region_full} and the supplementary material) shows that, even if we disregard the network structure of the grammar, the performance of the model is distributed across many features: performance remains rather high even with the top 25\% of constructions removed. Thus, we know that variation is distributed widely across the grammar (regardless of cluster assignments) and that macro- and micro-clusters also vary widely in predictive power. This is important because it means that a grammar's variation is not simply the sum of the variation of its component constructions.

\textbf{Finding 3: Similarity relations between dialects diverge widely from the best-performing model depending on the sub-set of the grammar being observed.} Given that the entire late-stage grammar retains high classification performance, we extract similarity relationships between dialects by looking at the errors which the classifier makes. For example, our reasoning is that Midwestern American English and Plains American English have more errors precisely because their grammar is more similar and thus easier to confuse. But the question is whether all sub-sets of the grammar lead to the same similarity errors. The answer is a resounding no, as shown in Tables \ref{tab:errors_region}, \ref{tab:errors_country}, and\ref{tab:errors_north_america} and in Figures \ref{fig:error_distribution_region}, \ref{fig:error_distribution_country}, and \ref{fig:error_distribution_north_america}. Taking similarity measures as a characterization of the dialects relative to one another, this means that the overall story of variation depends heavily on which sub-set of the grammar we observe.

Why is this a fundamental problem for previous work? Our interpretation is that language is a complex adaptive system so that examining any sub-set in isolation leads to incorrect conclusions. Thus, if we observed a small part of the grammar and tried to predict the most similar grammars overall, we would rarely reach the same relationships. By treating parts of the grammar as independent and self-contained systems, we fail to capture the interactions and complexities of the grammar as a single functioning system. The implication of these findings are that all studies which are based on isolating one or two constructions are fundamentally unable to provide an accurate representation of syntactic variation.



\section*{Supplemental Data}
The full supplementary materials contain the raw classification results, the full grammar and composition of macro- and micro-clusters, and additional figures and maps for each level of spatial granularity. These additional results repeat the main findings of the analysis presented in the paper. The supplementary material can be found \href{https://osf.io/2akmy/}{\textbf{at this link}}.

\section*{Data Availability Statement}
While the corpora analyzed for this study cannot be shared directly, the constructional features used for the classification models can be found \href{https://osf.io/2akmy/}{\textbf{at this link}}. This allows the replication of these findings.


\appendix

\begin{table*}[t]
\centering
\begin{tabular}{|l|l|l|l|l|l|l|l|}
\hline
know & time & people & day & love & new & see & think \\
why & here & want & go & really & need & today & make \\
still & because & first & very & best & after & than & never \\
got & much & back & please & going & great & right & then \\
life & thank & well & way & always & year & over & world \\
most & take & man & say & last & let & into & work \\
where & other & look & many & said & off & same & years \\
which & game & video & better & come & something & happy & thanks \\
via & yes & down & hope & god & stop & give & ever \\
feel & everyone & big & team & help & live & getting & while \\
use & keep & things & another & long & week & sure & days \\
watch & real & looking & shit & against & actually & doing & money \\
free & show & since & home & lot & nothing & bad & find \\
already & read & through & part & tell & without & won & such \\
start & little & play & thought & everything & morning & old & support \\
person & call & done & check & mean & news & put & both \\
wait & women & end & believe & used & top & around & night \\
looks & family & name & country & yeah & anyone & between & gonna \\
trying & says & hard & guys & maybe & friends & point & beautiful \\
remember & win & full & sorry & follow & government & high & during \\
amazing & yet & making & school & under & anything & coming & state \\
post & away & guy & change & try & house & open & might \\
season & whole & makes & left & song & media & saying & few \\
using & president & different & enough & black & called & talk & trump \\
place & talking & friend & care & power & once & wrong & city \\
working & nice & ready & times & business & understand & set & music \\
join & buy & vote & hate & heart & future & girl & mind \\
wish & face & seen & tomorrow & found & needs & watching & party \\
though & playing & least & problem & stay & covid & project & head \\
kind & white & group & health & until & food & story & cause \\
literally & soon & men & congratulations & job & ask & police & human \\
saw & waiting & far & ~ & ~ & ~ & ~ & ~ \\
\hline
  \end{tabular}
  \caption{Keywords used to create lexically-balanced samples.}
  \label{tab:appendix}
\end{table*}

\bibliographystyle{Frontiers-Harvard} 
\bibliography{References}


\end{document}


\onecolumn
\firstpage{1}

\title {{\helveticaitalic{Supplementary Material}}}

\maketitle

\begin{figure*}
\centering
\fbox{\includegraphics[width = 500pt]{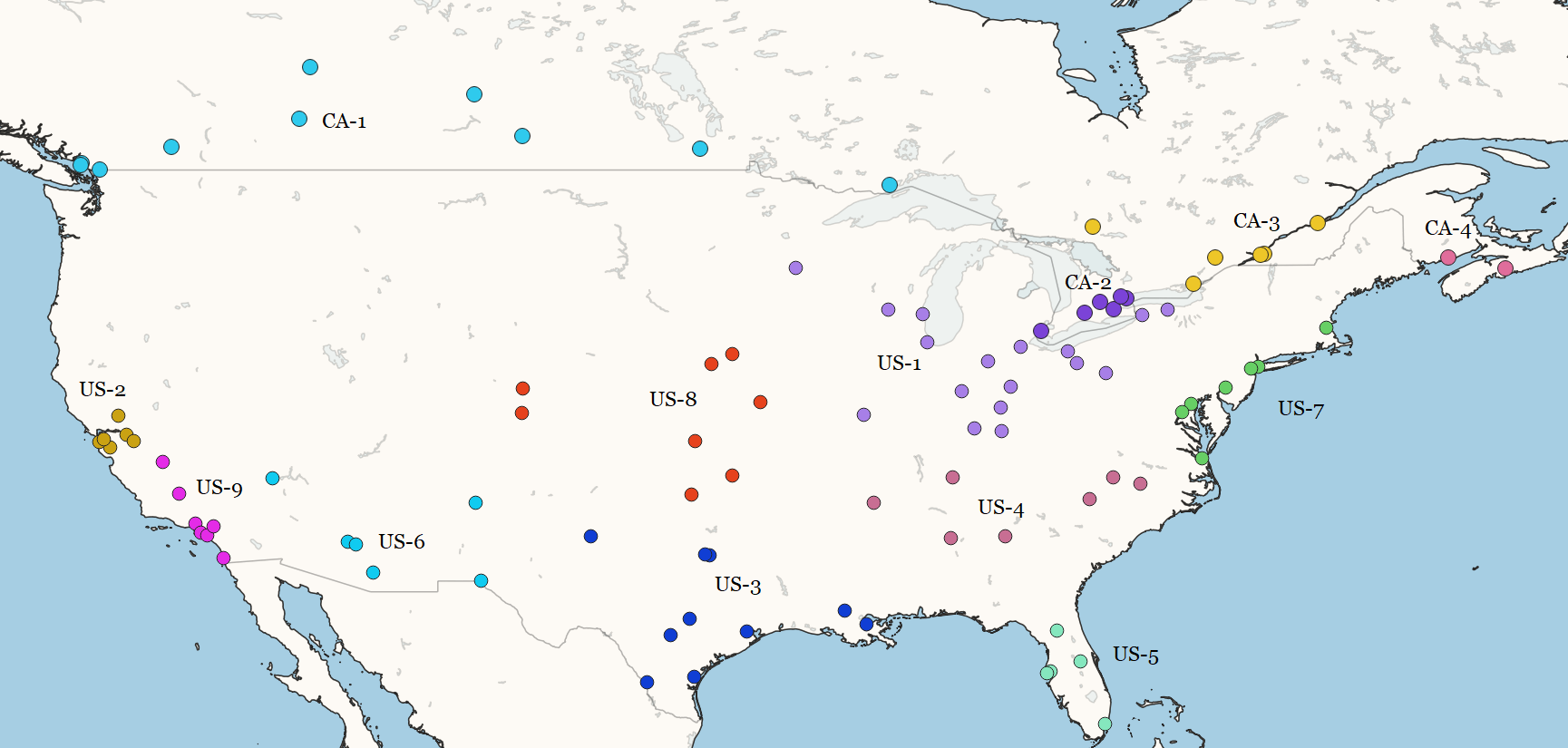}}
\caption{Distribution of Local Dialects in North America. Each point is an a collection location (an airport as a proxy for urban areas). The color of each point represents the local dialect to which it is assigned, as indicated by the labels.}
\end{figure*}

\begin{figure*}
\centering
\fbox{\includegraphics[width = 500pt]{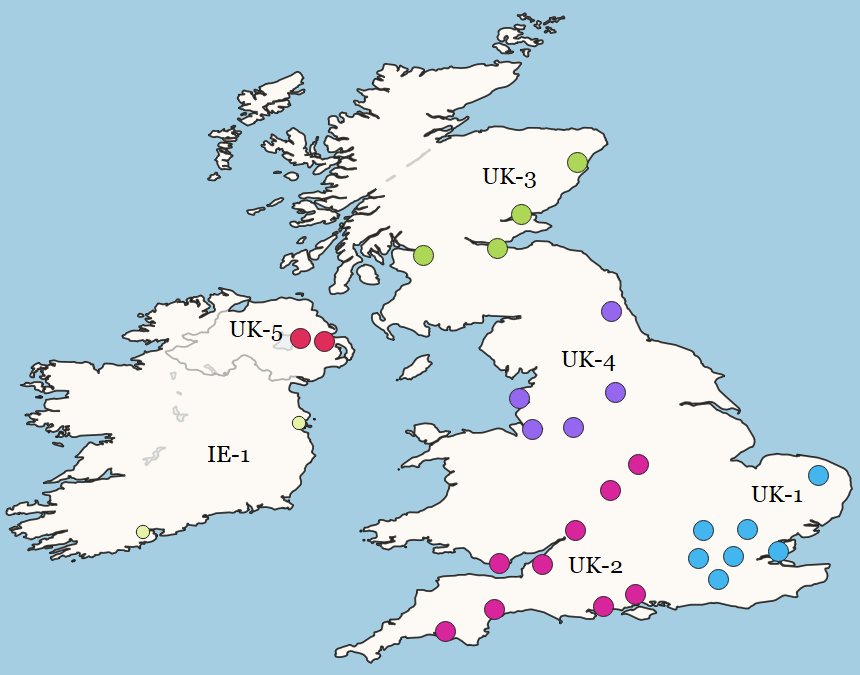}}
\caption{Distribution of Local Dialects in Europe. Each point is an a collection location (an airport as a proxy for urban areas). The color of each point represents the local dialect to which it is assigned, as indicated by the labels.}
\end{figure*}

\begin{figure*}
\centering
\fbox{\includegraphics[width = 500pt]{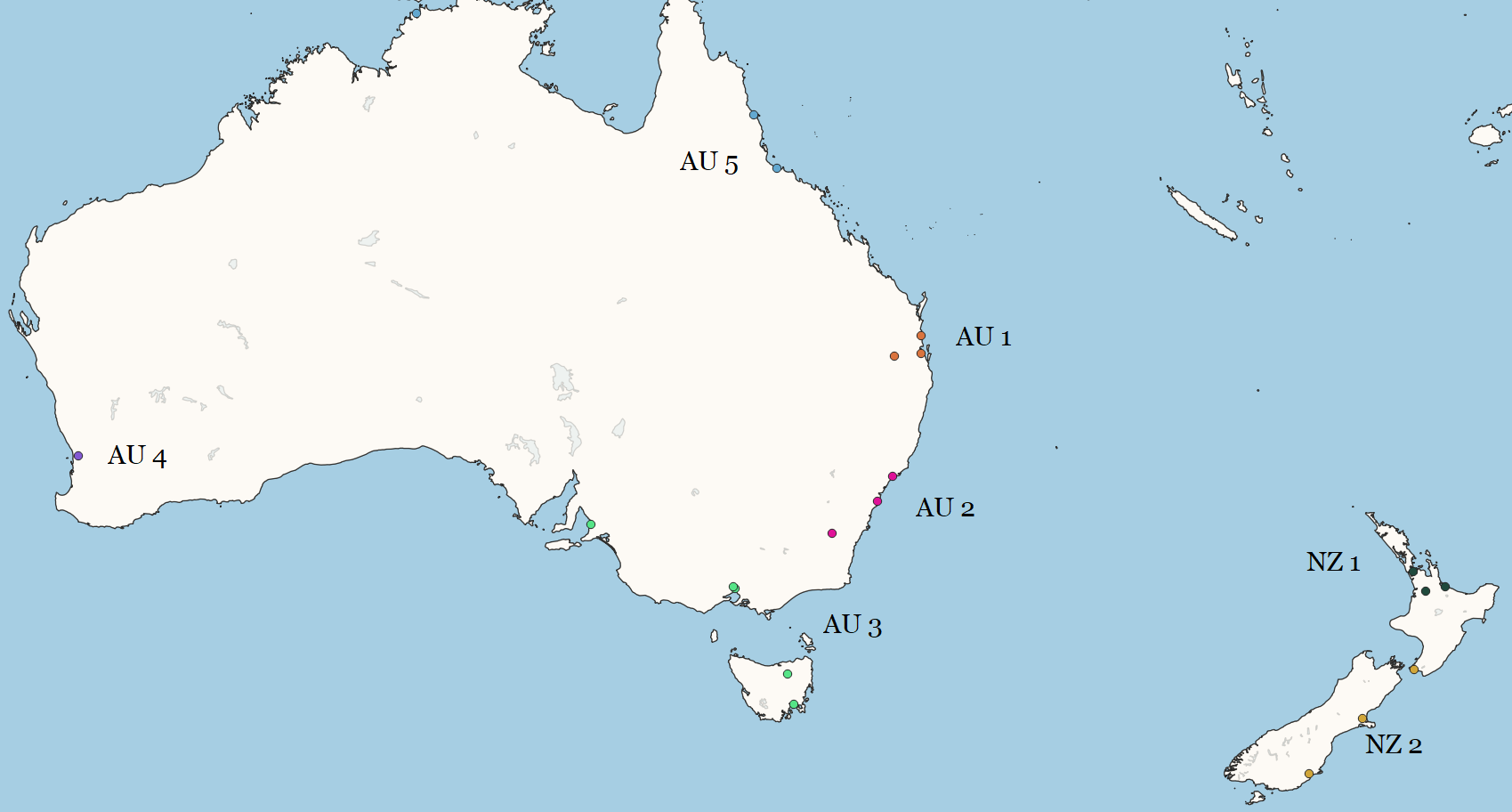}}
\caption{Distribution of Local Dialects in Oceania. Each point is an a collection location (an airport as a proxy for urban areas). The color of each point represents the local dialect to which it is assigned, as indicated by the labels.}
\end{figure*}

\begin{figure*}
\centering
\fbox{\includegraphics[width = 500pt]{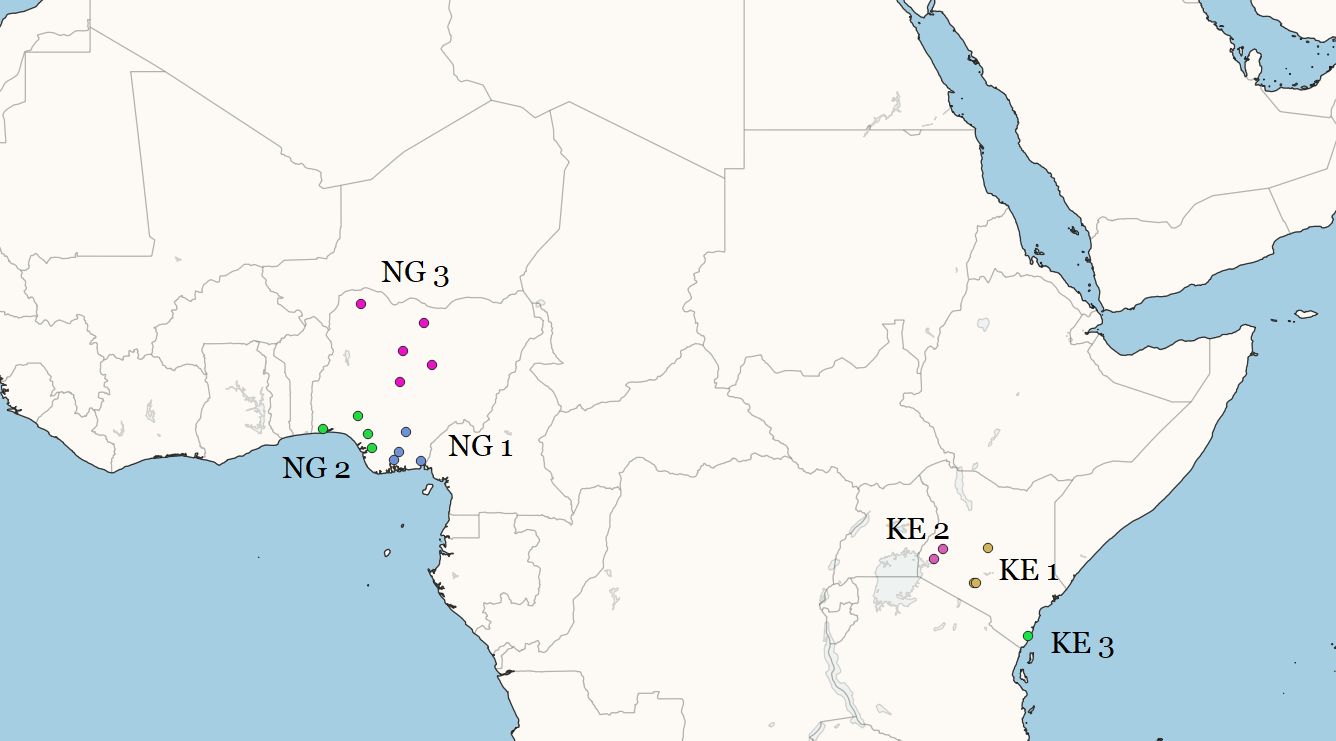}}
\caption{Distribution of Local Dialects in Sub-Saharan Africa. Each point is an a collection location (an airport as a proxy for urban areas). The color of each point represents the local dialect to which it is assigned, as indicated by the labels.}
\end{figure*}

\begin{figure*}
\centering
\fbox{\includegraphics[width = 500pt]{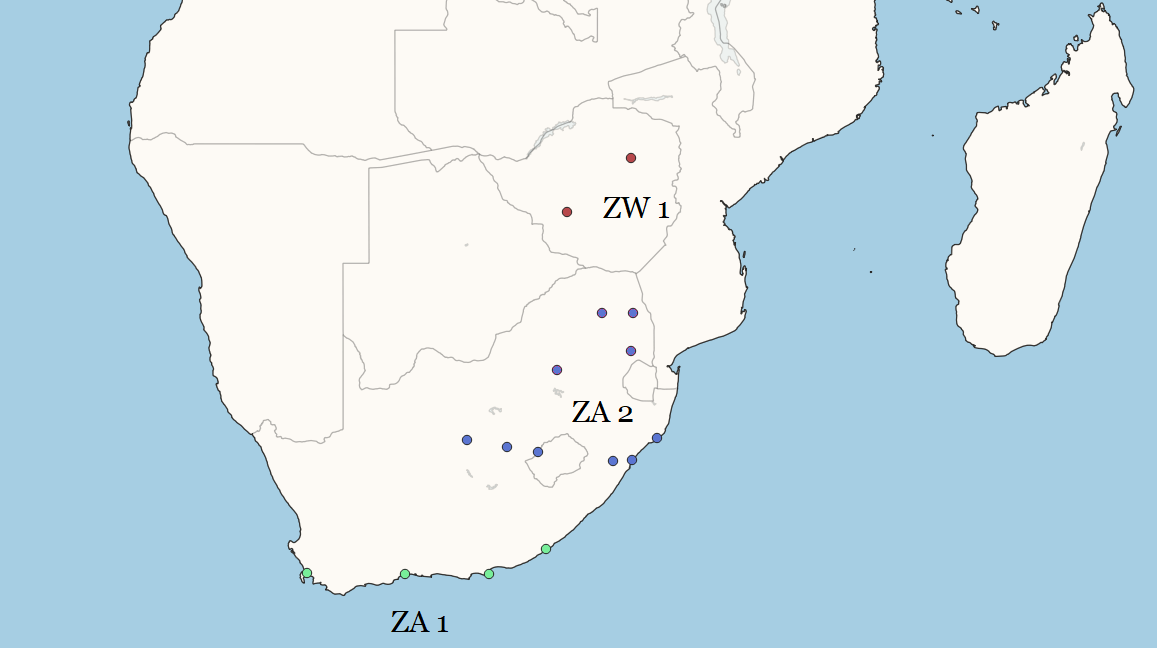}}
\caption{Distribution of Local Dialects in Southern Africa. Each point is an a collection location (an airport as a proxy for urban areas). The color of each point represents the local dialect to which it is assigned, as indicated by the labels.}
\label{fig:north_america}
\end{figure*}

\begin{figure*}
\centering
\fbox{\includegraphics[width = 500pt]{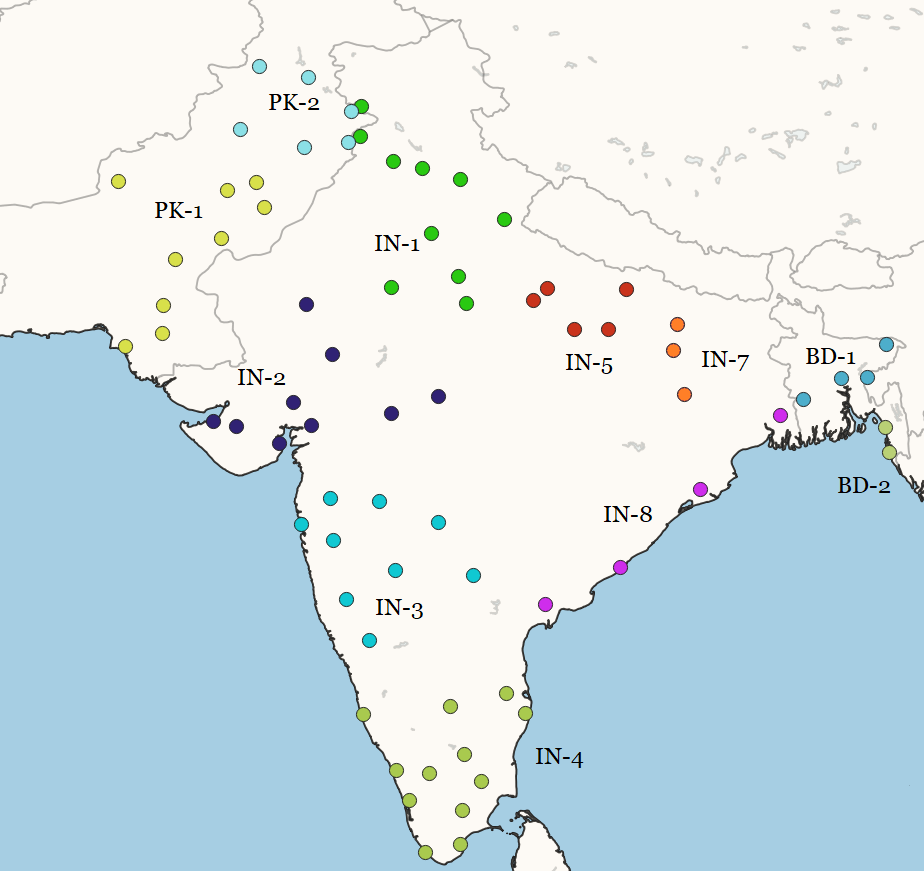}}
\caption{Distribution of Local Dialects in South Asia. Each point is an a collection location (an airport as a proxy for urban areas). The color of each point represents the local dialect to which it is assigned, as indicated by the labels.}
\end{figure*}

\begin{figure*}
\centering
\fbox{\includegraphics[width = 500pt]{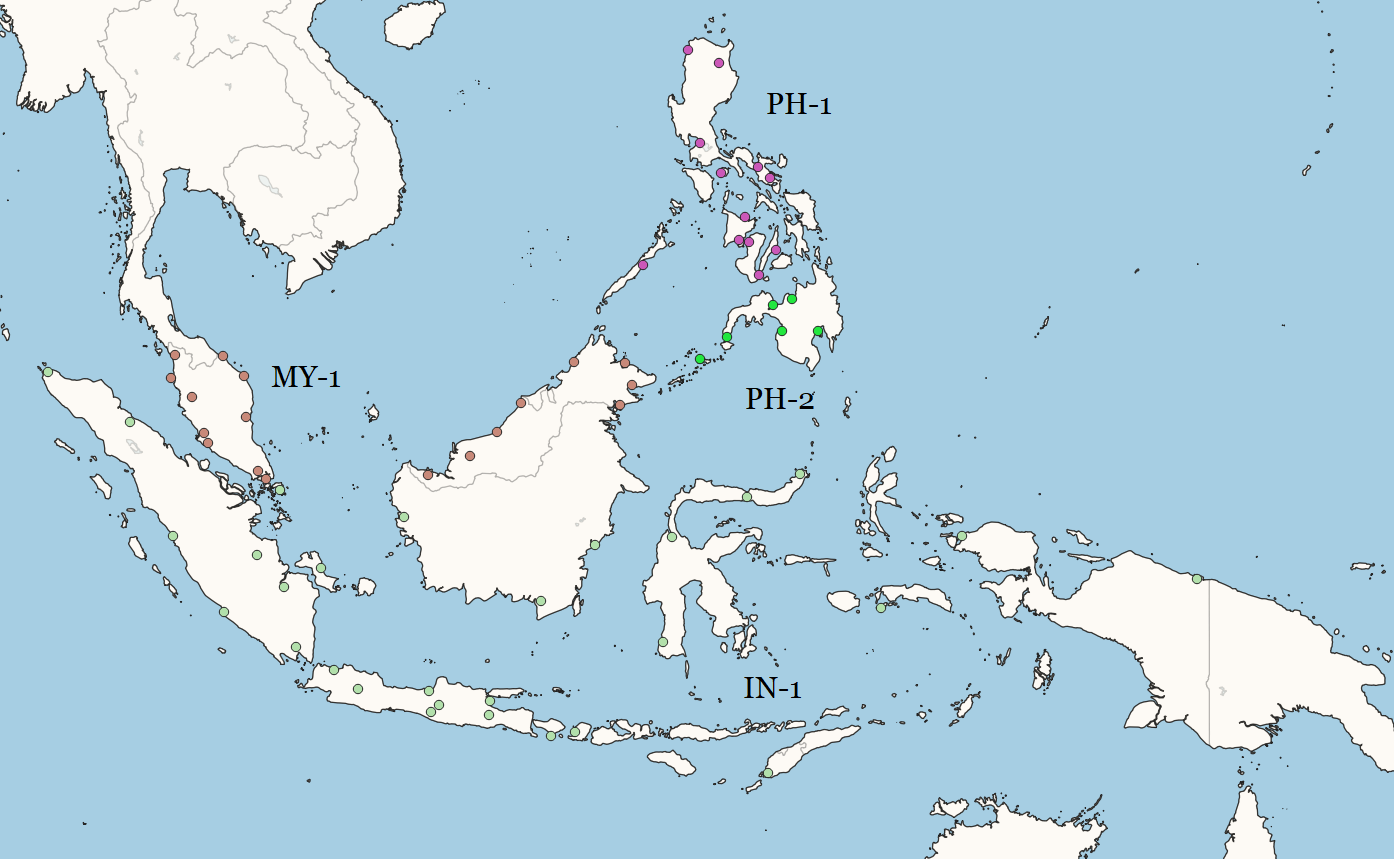}}
\caption{Distribution of Local Dialects in Southeast Asia. Each point is an a collection location (an airport as a proxy for urban areas). The color of each point represents the local dialect to which it is assigned, as indicated by the labels.}
\end{figure*}

\begin{figure*}
\centering
\fbox{\includegraphics[width = 500pt]{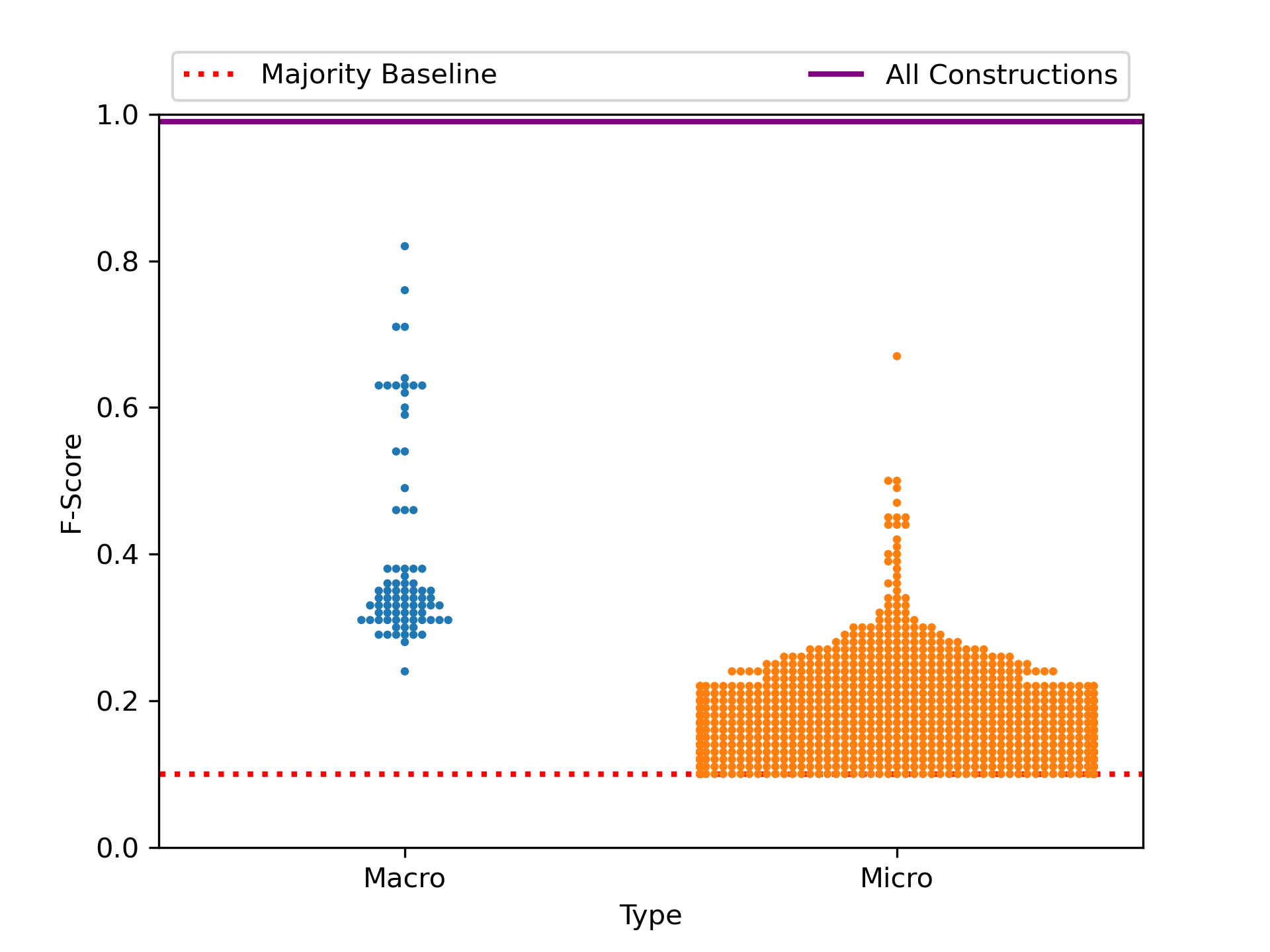}}
\caption{Distribution of Classification Performance Across Sub-Sets of the Grammar with the Late-Stage Grammar for \textbf{Regional Dialects}. Each macro-cluster and micro-cluster of constructions is plotted with its f-score on the dialect classification task, with both the performance of the entire late-stage grammar and the majority baseline also shown.}
\end{figure*}

\begin{figure*}
\centering
\fbox{\includegraphics[width = 500pt]{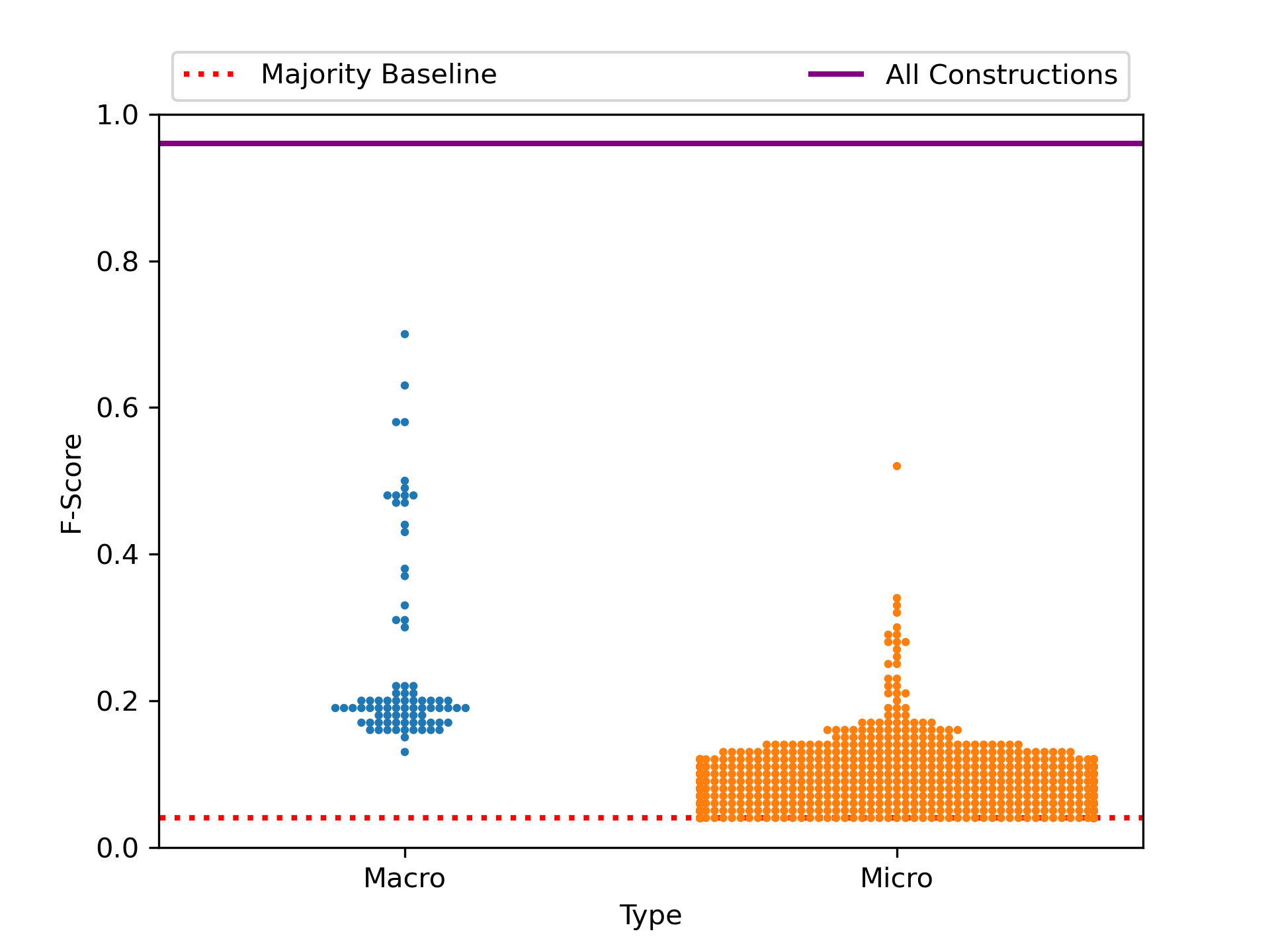}}
\caption{Distribution of Classification Performance Across Sub-Sets of the Grammar with the Late-Stage Grammar for \textbf{National Dialects}. Each macro-cluster and micro-cluster of constructions is plotted with its f-score on the dialect classification task, with both the performance of the entire late-stage grammar and the majority baseline also shown.}
\end{figure*}

\begin{figure*}
\centering
\fbox{\includegraphics[width = 500pt]{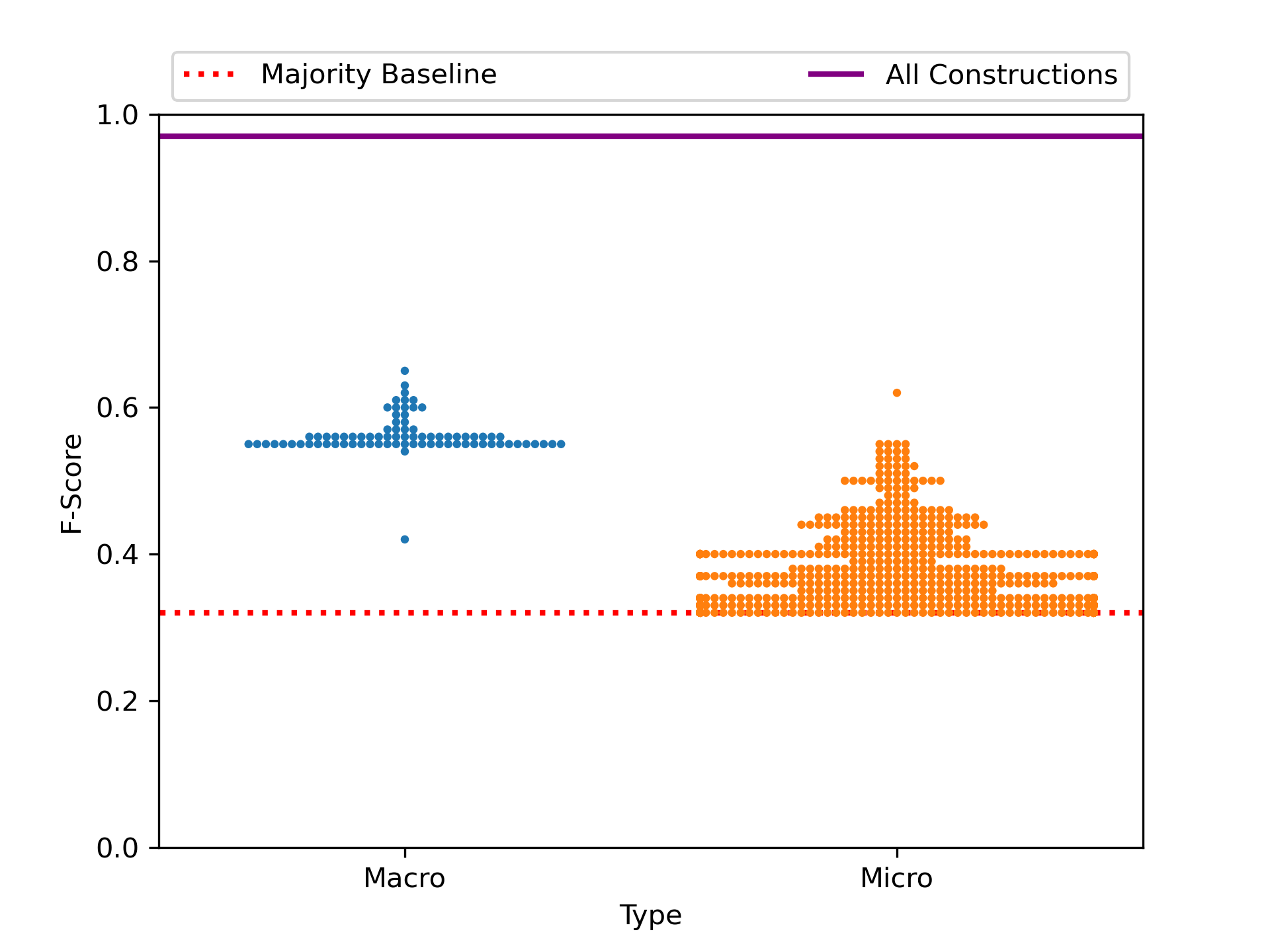}}
\caption{Distribution of Classification Performance Across Sub-Sets of the Grammar with the Late-Stage Grammar for \textbf{Local Dialects in Southern Africa}. Each macro-cluster and micro-cluster of constructions is plotted with its f-score on the dialect classification task, with both the performance of the entire late-stage grammar and the majority baseline also shown.}
\end{figure*}

\begin{figure*}
\centering
\fbox{\includegraphics[width = 500pt]{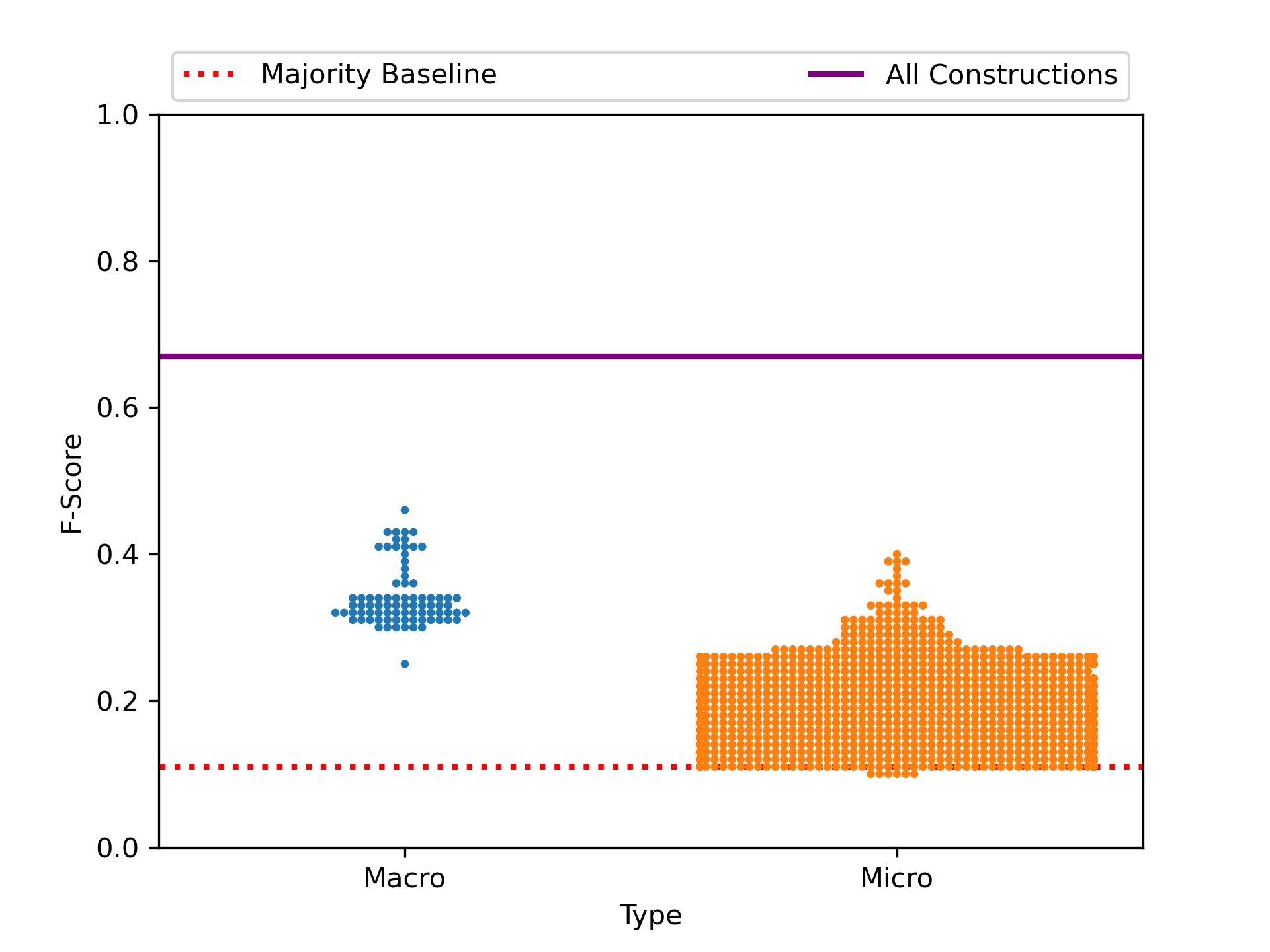}}
\caption{Distribution of Classification Performance Across Sub-Sets of the Grammar with the Late-Stage Grammar for \textbf{Local Dialects in Sub-Saharan Africa}. Each macro-cluster and micro-cluster of constructions is plotted with its f-score on the dialect classification task, with both the performance of the entire late-stage grammar and the majority baseline also shown.}
\end{figure*}

\begin{figure*}
\centering
\fbox{\includegraphics[width = 500pt]{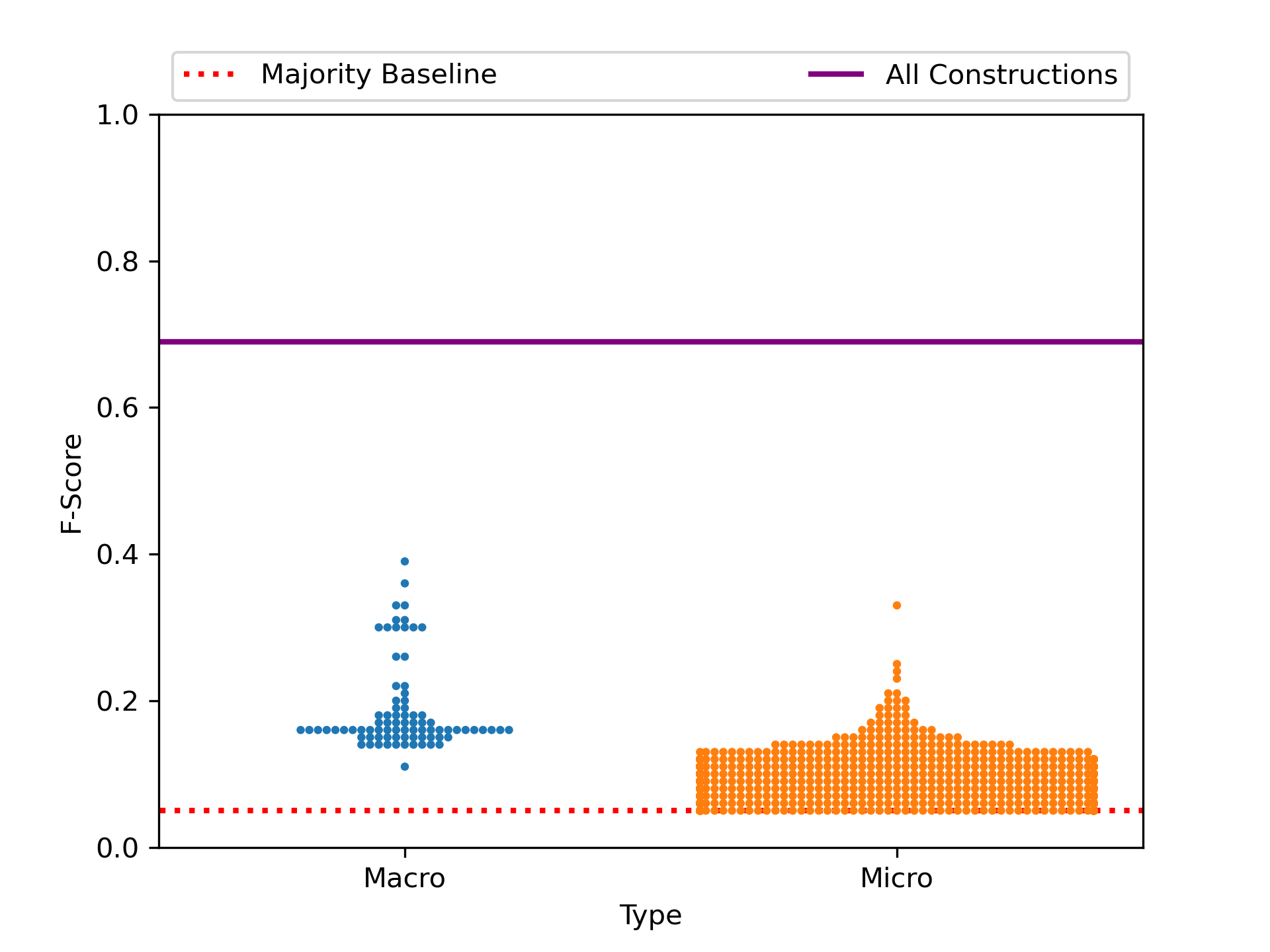}}
\caption{Distribution of Classification Performance Across Sub-Sets of the Grammar with the Late-Stage Grammar for \textbf{Local Dialects in North America}. Each macro-cluster and micro-cluster of constructions is plotted with its f-score on the dialect classification task, with both the performance of the entire late-stage grammar and the majority baseline also shown.}
\end{figure*}

\begin{figure*}
\centering
\fbox{\includegraphics[width = 500pt]{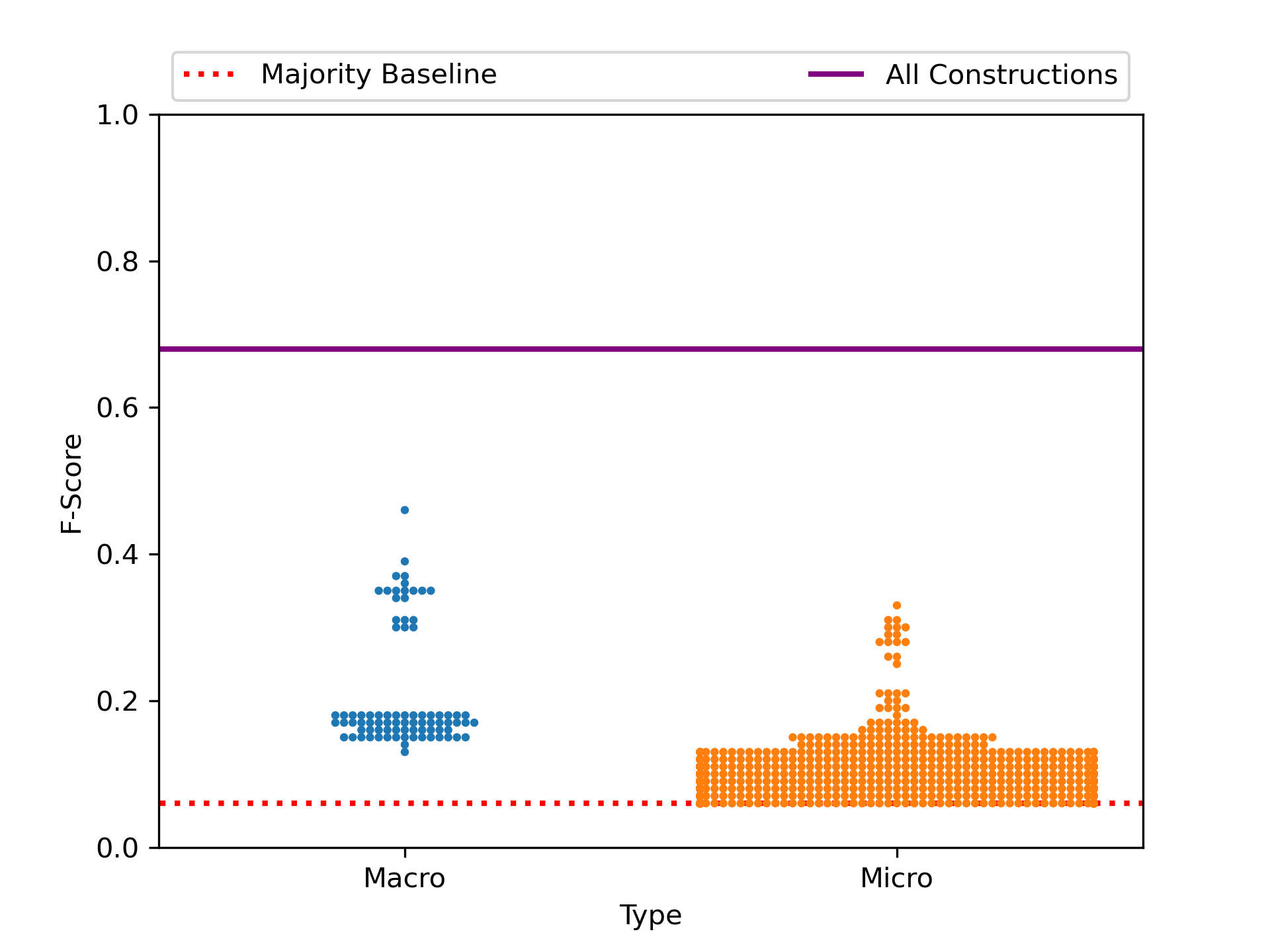}}
\caption{Distribution of Classification Performance Across Sub-Sets of the Grammar with the Late-Stage Grammar for \textbf{Local Dialects in South Asia}. Each macro-cluster and micro-cluster of constructions is plotted with its f-score on the dialect classification task, with both the performance of the entire late-stage grammar and the majority baseline also shown.}
\end{figure*}

\begin{figure*}
\centering
\fbox{\includegraphics[width = 500pt]{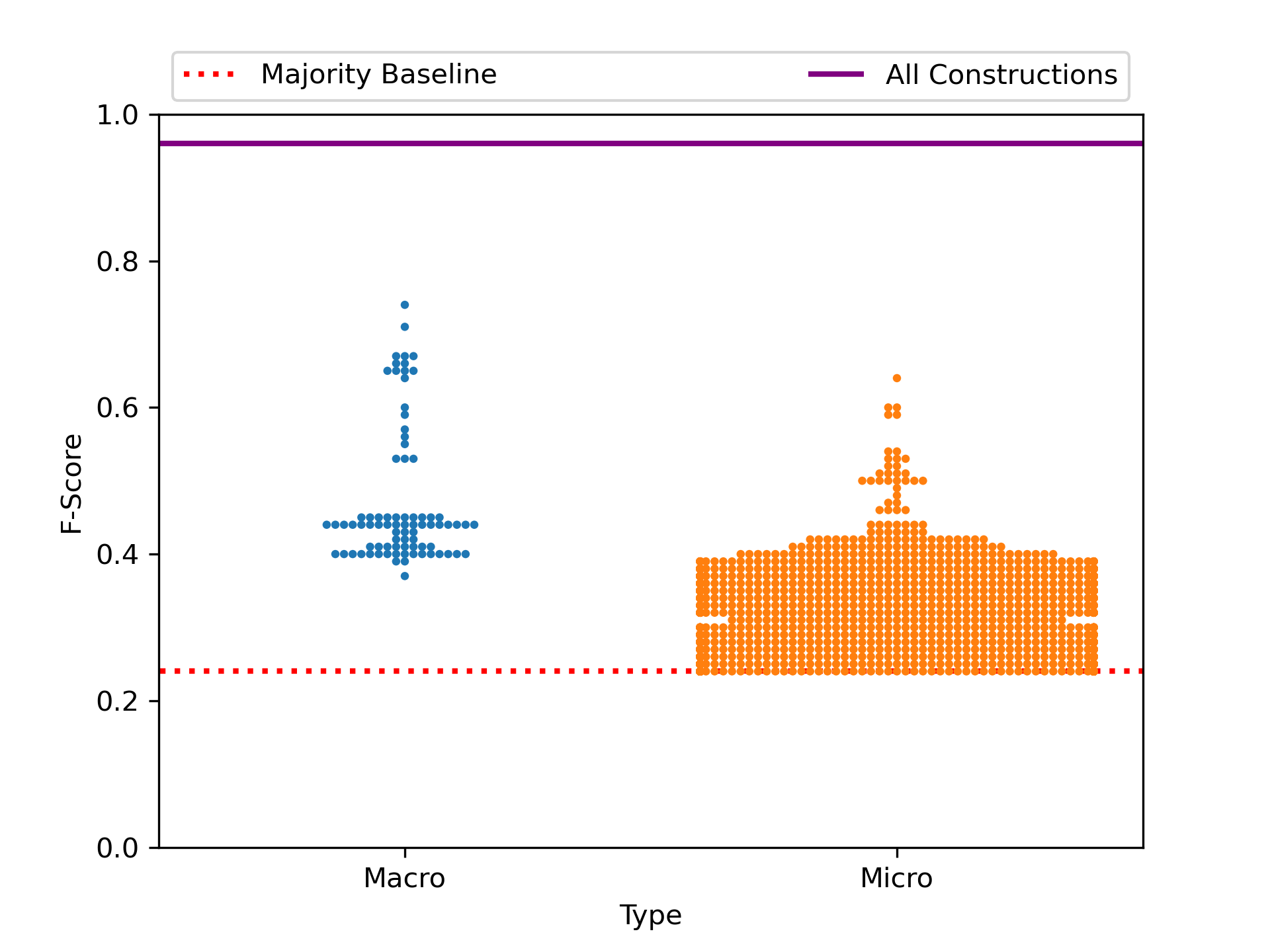}}
\caption{Distribution of Classification Performance Across Sub-Sets of the Grammar with the Late-Stage Grammar for \textbf{Local Dialects in Southeast Asia}. Each macro-cluster and micro-cluster of constructions is plotted with its f-score on the dialect classification task, with both the performance of the entire late-stage grammar and the majority baseline also shown.}
\end{figure*}

\begin{figure*}
\centering
\fbox{\includegraphics[width = 500pt]{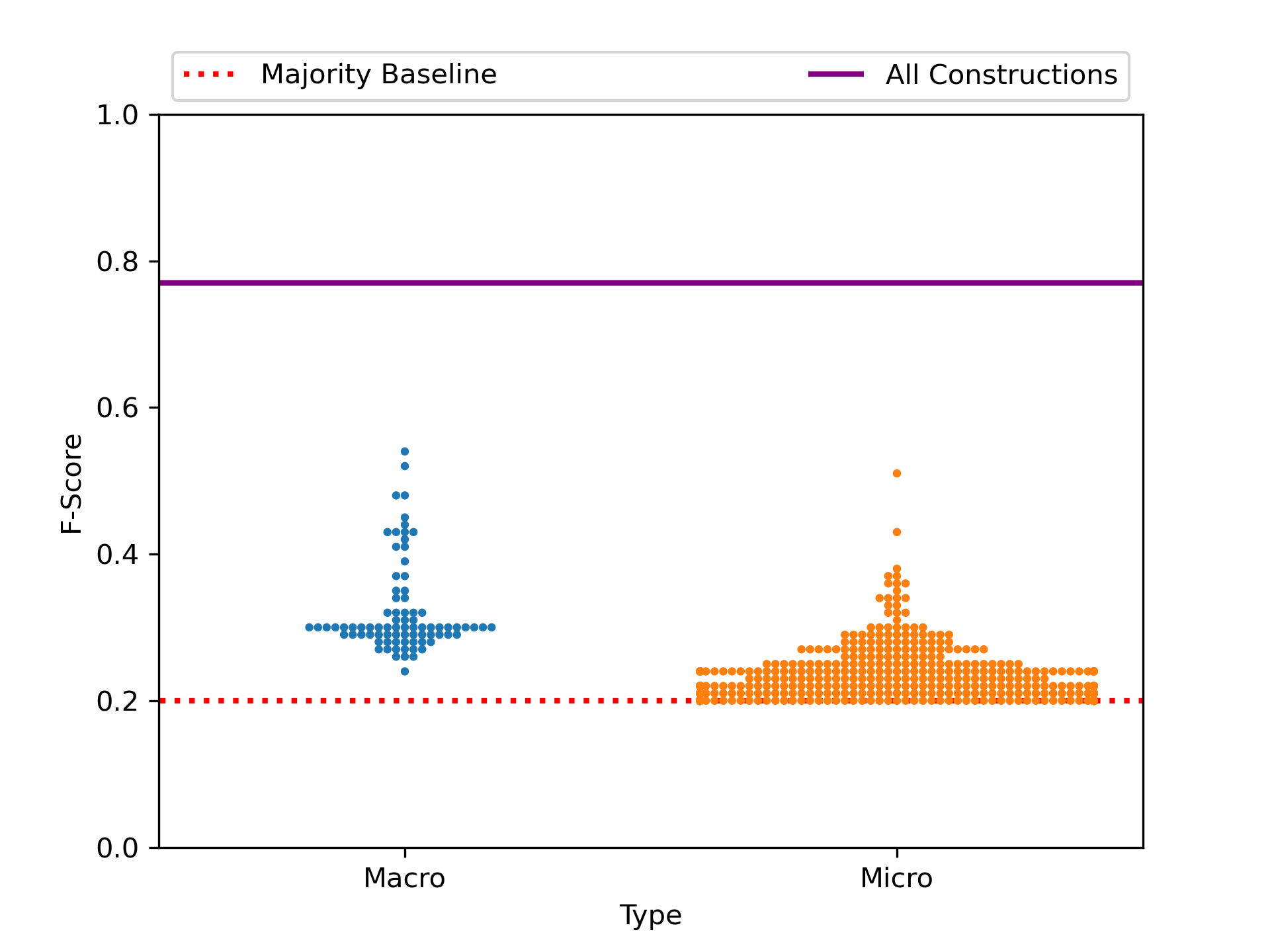}}
\caption{Distribution of Classification Performance Across Sub-Sets of the Grammar with the Late-Stage Grammar for \textbf{Local Dialects in Western Europe}. Each macro-cluster and micro-cluster of constructions is plotted with its f-score on the dialect classification task, with both the performance of the entire late-stage grammar and the majority baseline also shown.}
\end{figure*}

\begin{figure*}
\centering
\fbox{\includegraphics[width = 500pt]{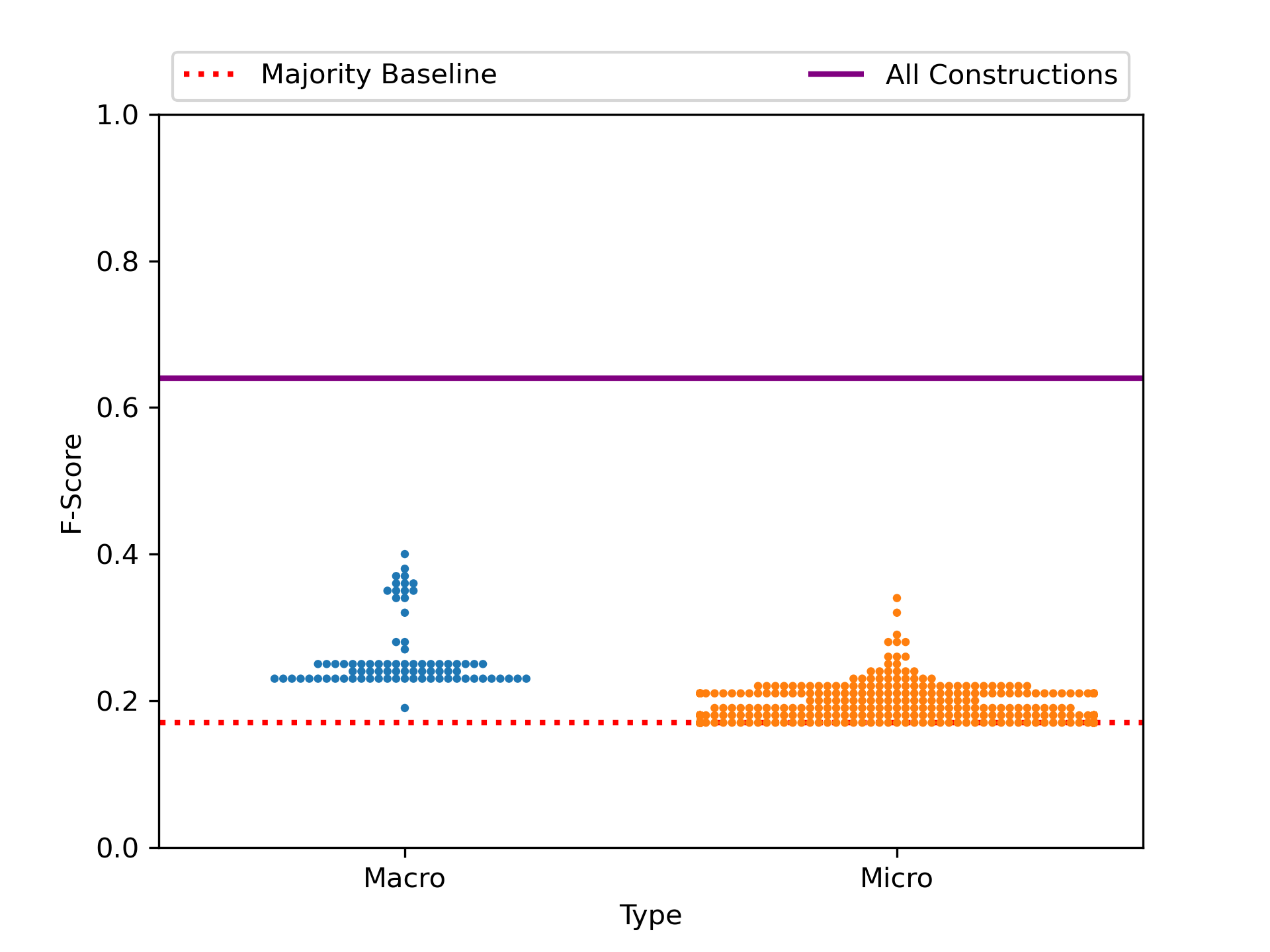}}
\caption{Distribution of Classification Performance Across Sub-Sets of the Grammar with the Late-Stage Grammar for \textbf{Local Dialects in Oceania}. Each macro-cluster and micro-cluster of constructions is plotted with its f-score on the dialect classification task, with both the performance of the entire late-stage grammar and the majority baseline also shown.}
\end{figure*}

\begin{figure*}
\centering
\fbox{\includegraphics[width = 500pt]{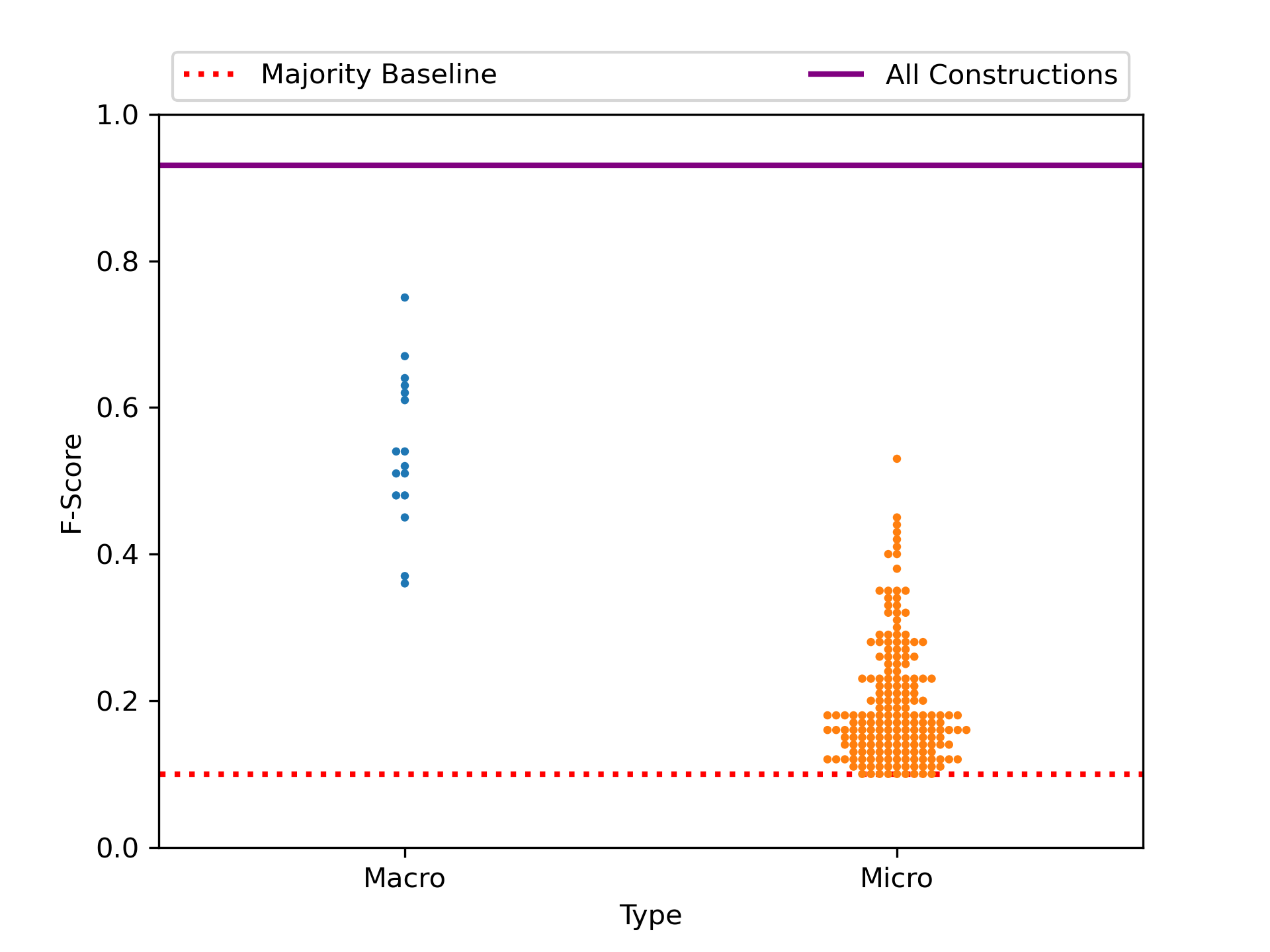}}
\caption{Distribution of Classification Performance Across Sub-Sets of the Grammar with the Early-Stage Grammar for \textbf{Regional Dialects}. Each macro-cluster and micro-cluster of constructions is plotted with its f-score on the dialect classification task, with both the performance of the entire late-stage grammar and the majority baseline also shown.}
\end{figure*}

\begin{figure*}
\centering
\fbox{\includegraphics[width = 500pt]{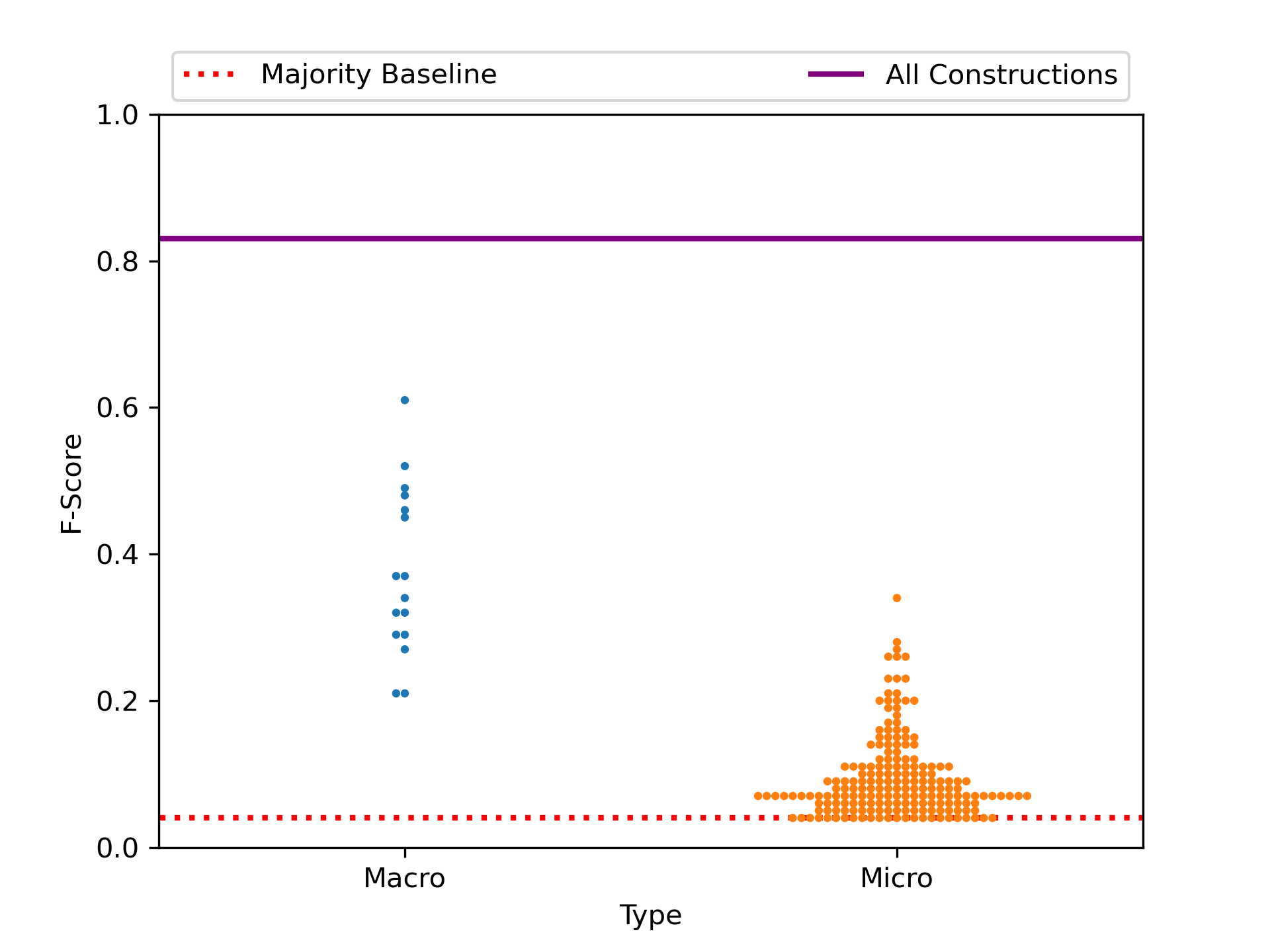}}
\caption{Distribution of Classification Performance Across Sub-Sets of the Grammar with the Early-Stage Grammar for \textbf{National Dialects}. Each macro-cluster and micro-cluster of constructions is plotted with its f-score on the dialect classification task, with both the performance of the entire late-stage grammar and the majority baseline also shown.}
\end{figure*}

\begin{figure*}
\centering
\fbox{\includegraphics[width = 500pt]{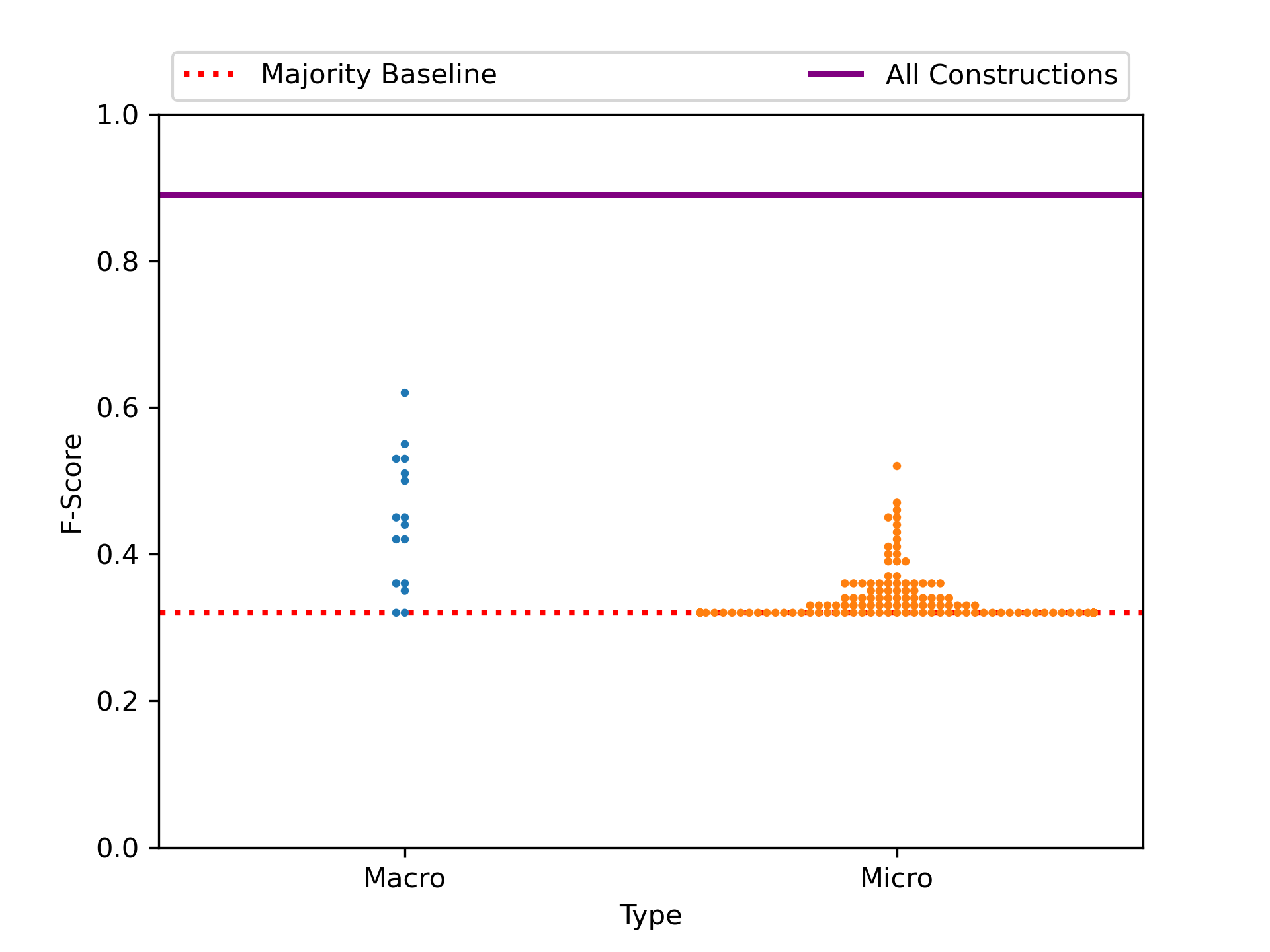}}
\caption{Distribution of Classification Performance Across Sub-Sets of the Grammar with the Early-Stage Grammar for \textbf{Local Dialects in Southern Africa}. Each macro-cluster and micro-cluster of constructions is plotted with its f-score on the dialect classification task, with both the performance of the entire late-stage grammar and the majority baseline also shown.}
\end{figure*}

\begin{figure*}
\centering
\fbox{\includegraphics[width = 500pt]{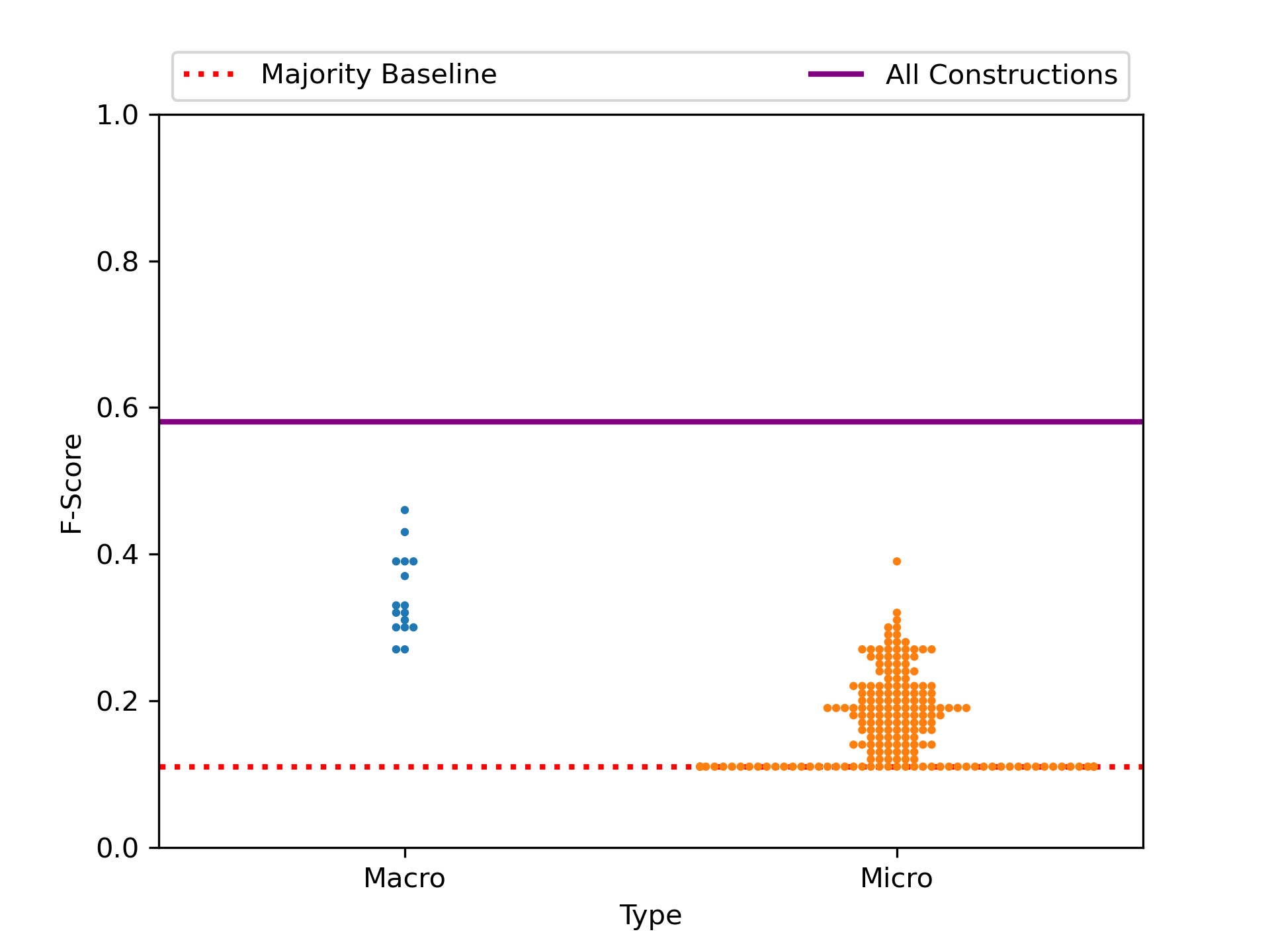}}
\caption{Distribution of Classification Performance Across Sub-Sets of the Grammar with the Early-Stage Grammar for \textbf{Local Dialects in Sub-Saharan Africa}. Each macro-cluster and micro-cluster of constructions is plotted with its f-score on the dialect classification task, with both the performance of the entire late-stage grammar and the majority baseline also shown.}
\end{figure*}

\begin{figure*}
\centering
\fbox{\includegraphics[width = 500pt]{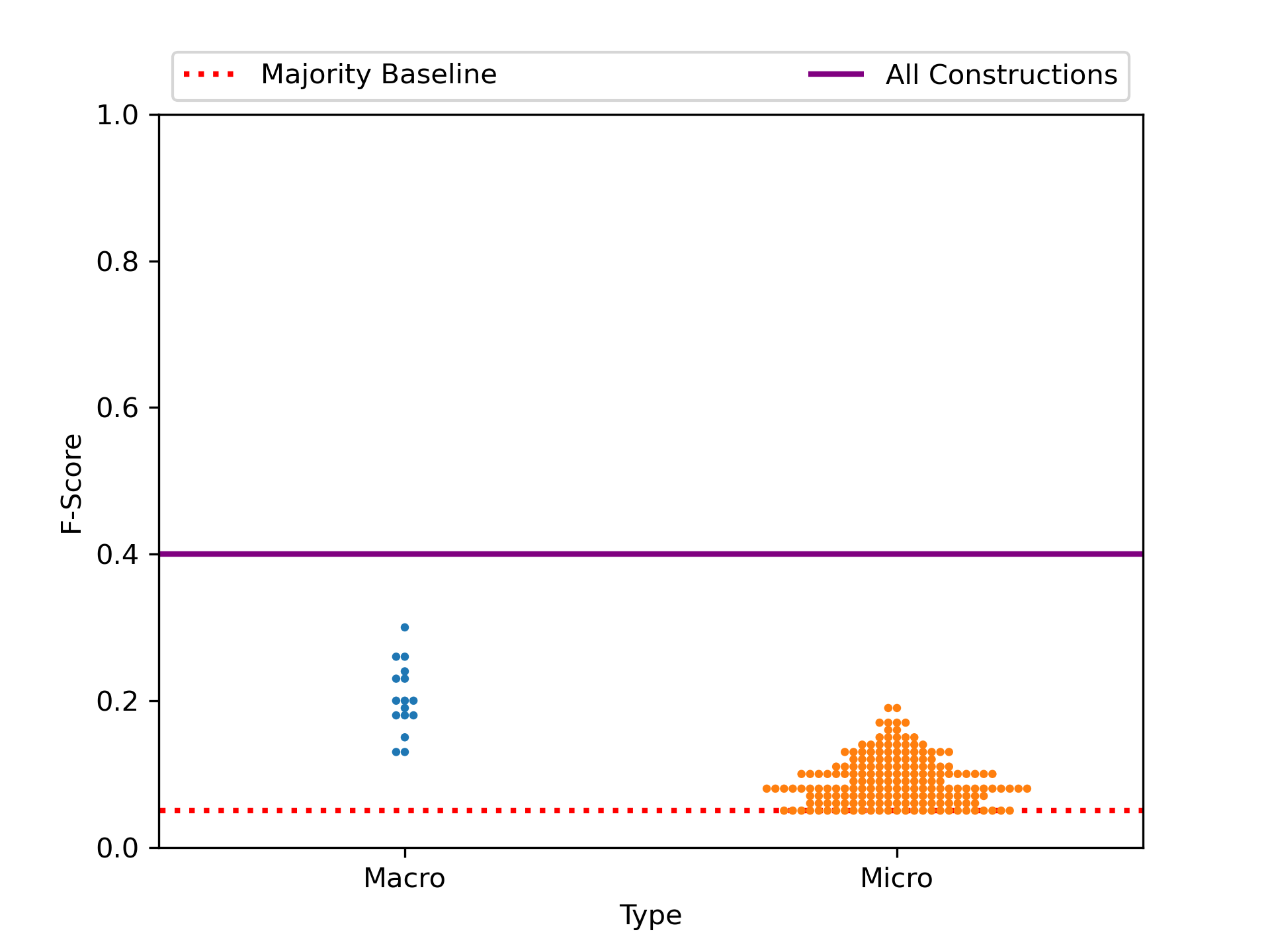}}
\caption{Distribution of Classification Performance Across Sub-Sets of the Grammar with the Early-Stage Grammar for \textbf{Local Dialects in North America}. Each macro-cluster and micro-cluster of constructions is plotted with its f-score on the dialect classification task, with both the performance of the entire late-stage grammar and the majority baseline also shown.}
\end{figure*}

\begin{figure*}
\centering
\fbox{\includegraphics[width = 500pt]{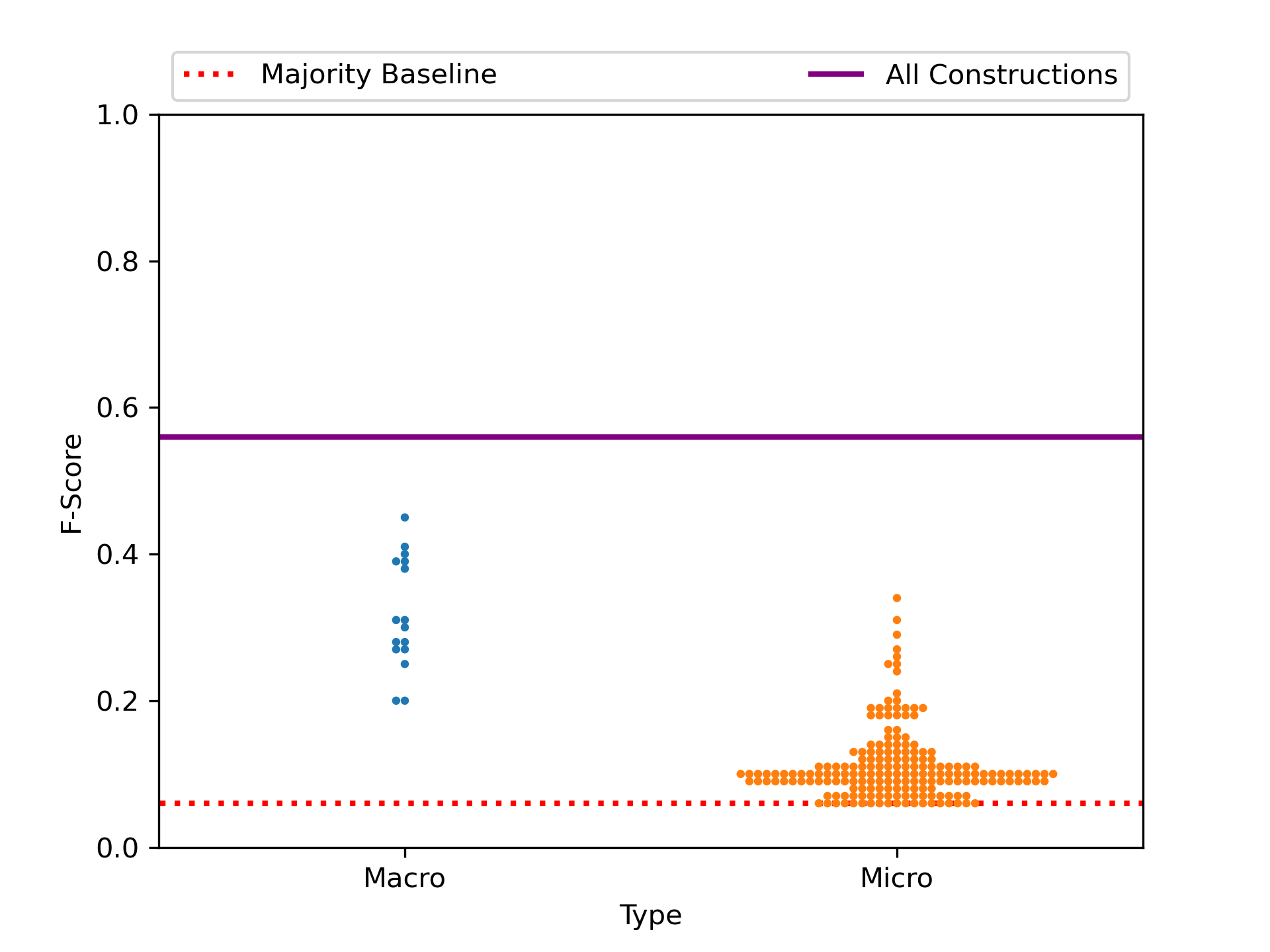}}
\caption{Distribution of Classification Performance Across Sub-Sets of the Grammar with the Early-Stage Grammar for \textbf{Local Dialects in South Asia}. Each macro-cluster and micro-cluster of constructions is plotted with its f-score on the dialect classification task, with both the performance of the entire late-stage grammar and the majority baseline also shown.}
\end{figure*}

\begin{figure*}
\centering
\fbox{\includegraphics[width = 500pt]{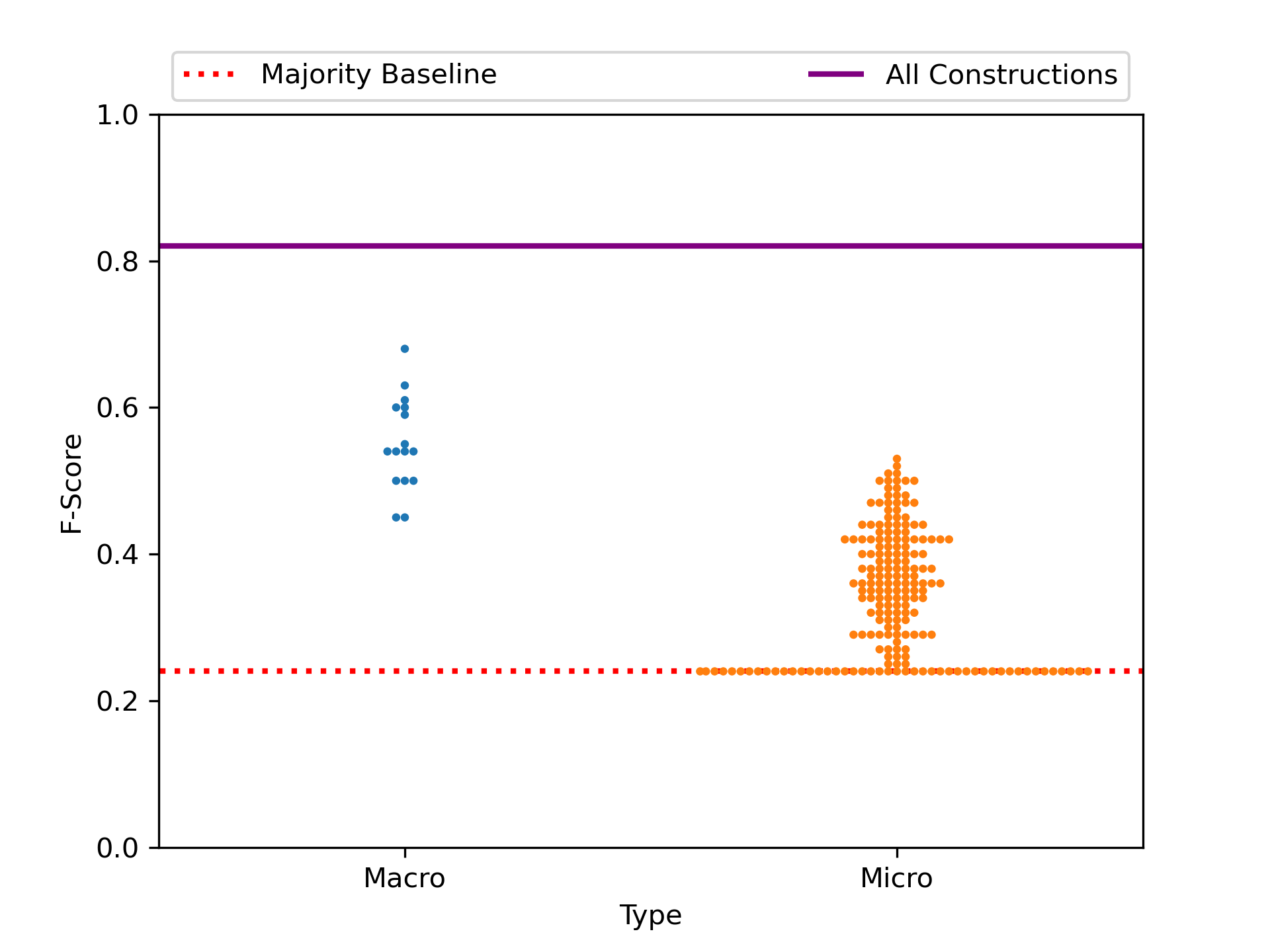}}
\caption{Distribution of Classification Performance Across Sub-Sets of the Grammar with the Early-Stage Grammar for \textbf{Local Dialects in Southeast Asia}. Each macro-cluster and micro-cluster of constructions is plotted with its f-score on the dialect classification task, with both the performance of the entire late-stage grammar and the majority baseline also shown.}
\end{figure*}

\begin{figure*}
\centering
\fbox{\includegraphics[width = 500pt]{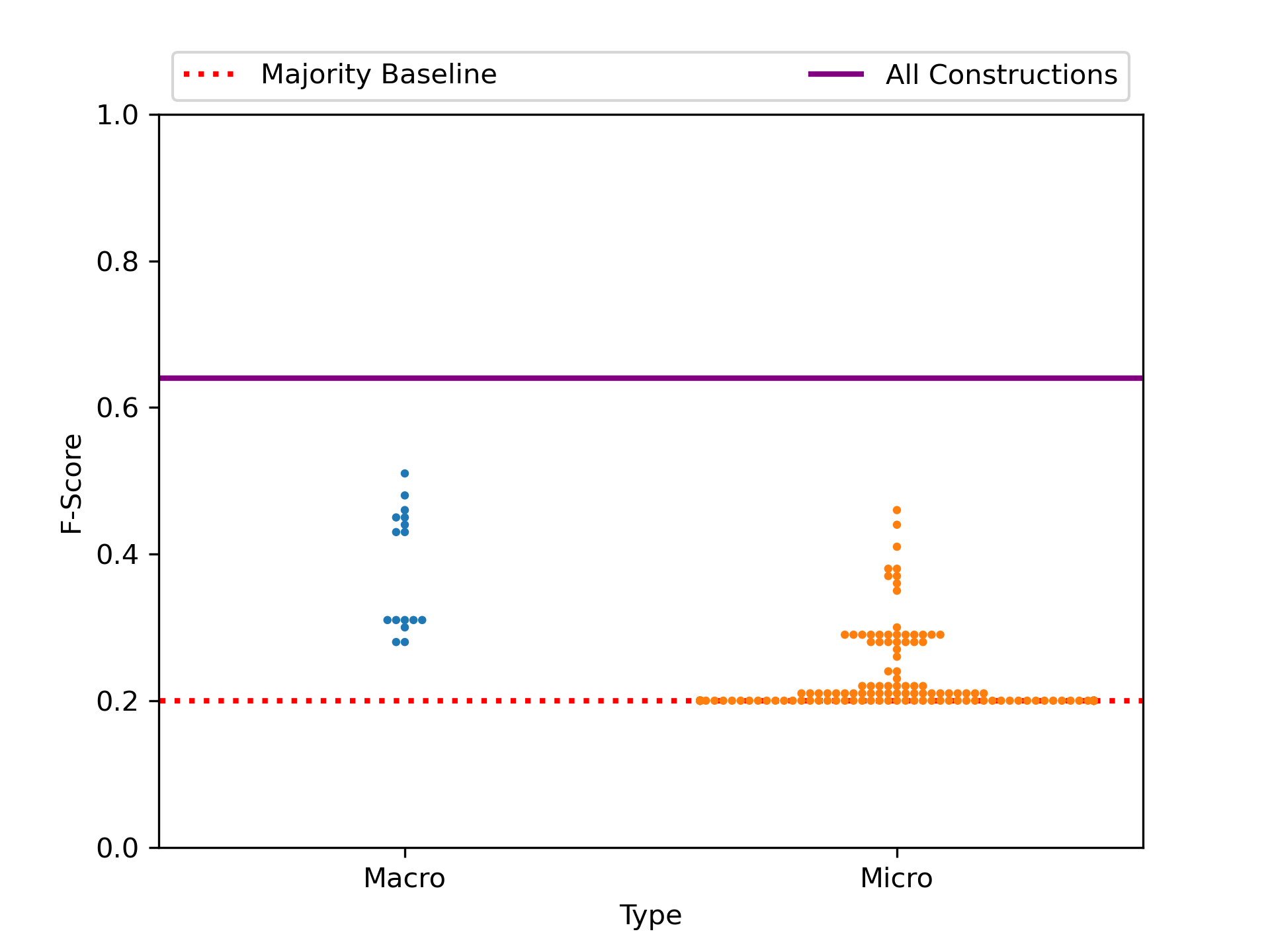}}
\caption{Distribution of Classification Performance Across Sub-Sets of the Grammar with the Early-Stage Grammar for \textbf{Local Dialects in Western Europe}. Each macro-cluster and micro-cluster of constructions is plotted with its f-score on the dialect classification task, with both the performance of the entire late-stage grammar and the majority baseline also shown.}
\end{figure*}

\begin{figure*}
\centering
\fbox{\includegraphics[width = 500pt]{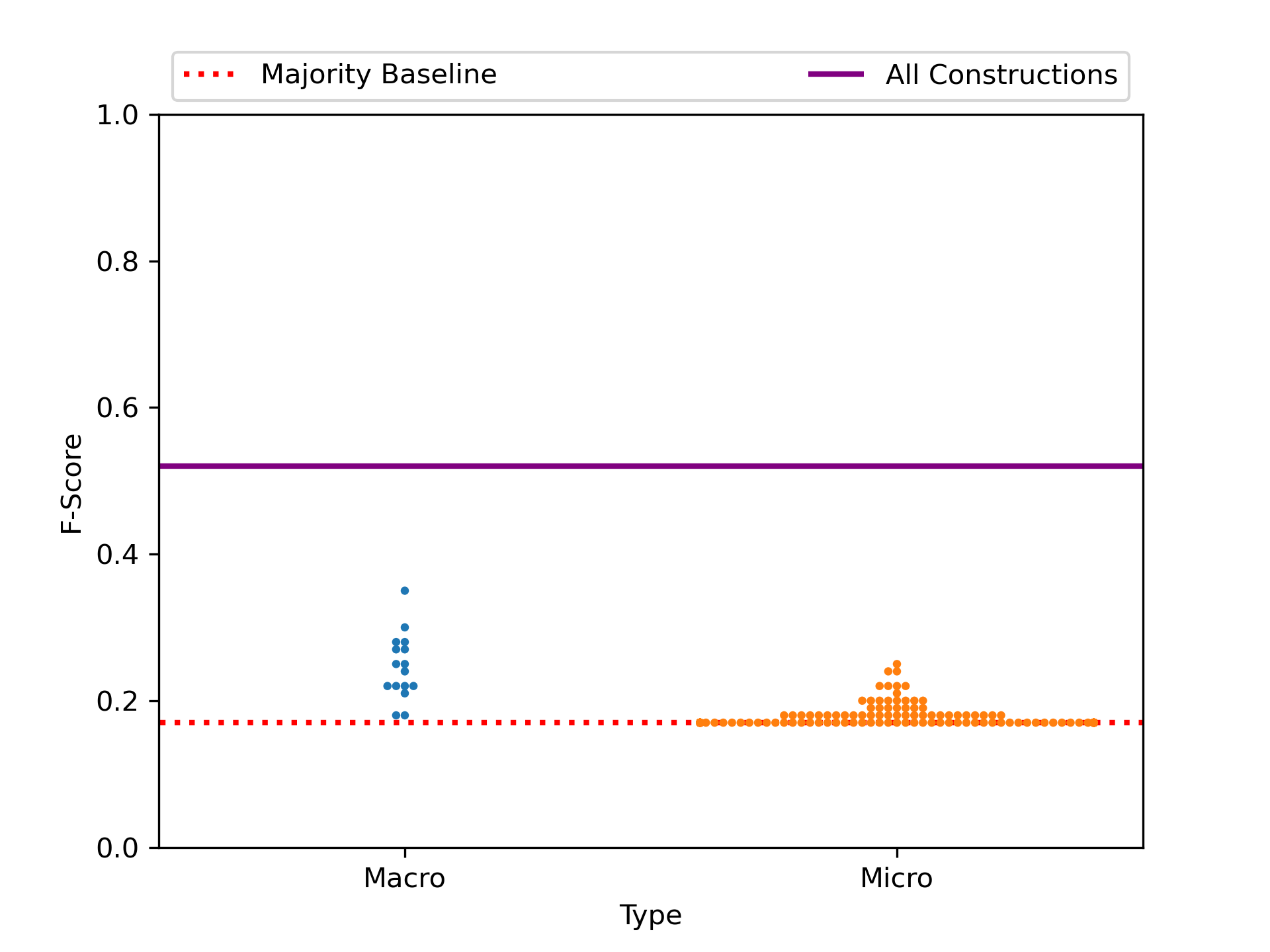}}
\caption{Distribution of Classification Performance Across Sub-Sets of the Grammar with the Early-Stage Grammar for \textbf{Local Dialects in Oceania}. Each macro-cluster and micro-cluster of constructions is plotted with its f-score on the dialect classification task, with both the performance of the entire late-stage grammar and the majority baseline also shown.}
\end{figure*}

\begin{figure*}
\centering
\fbox{\includegraphics[width = 500pt]{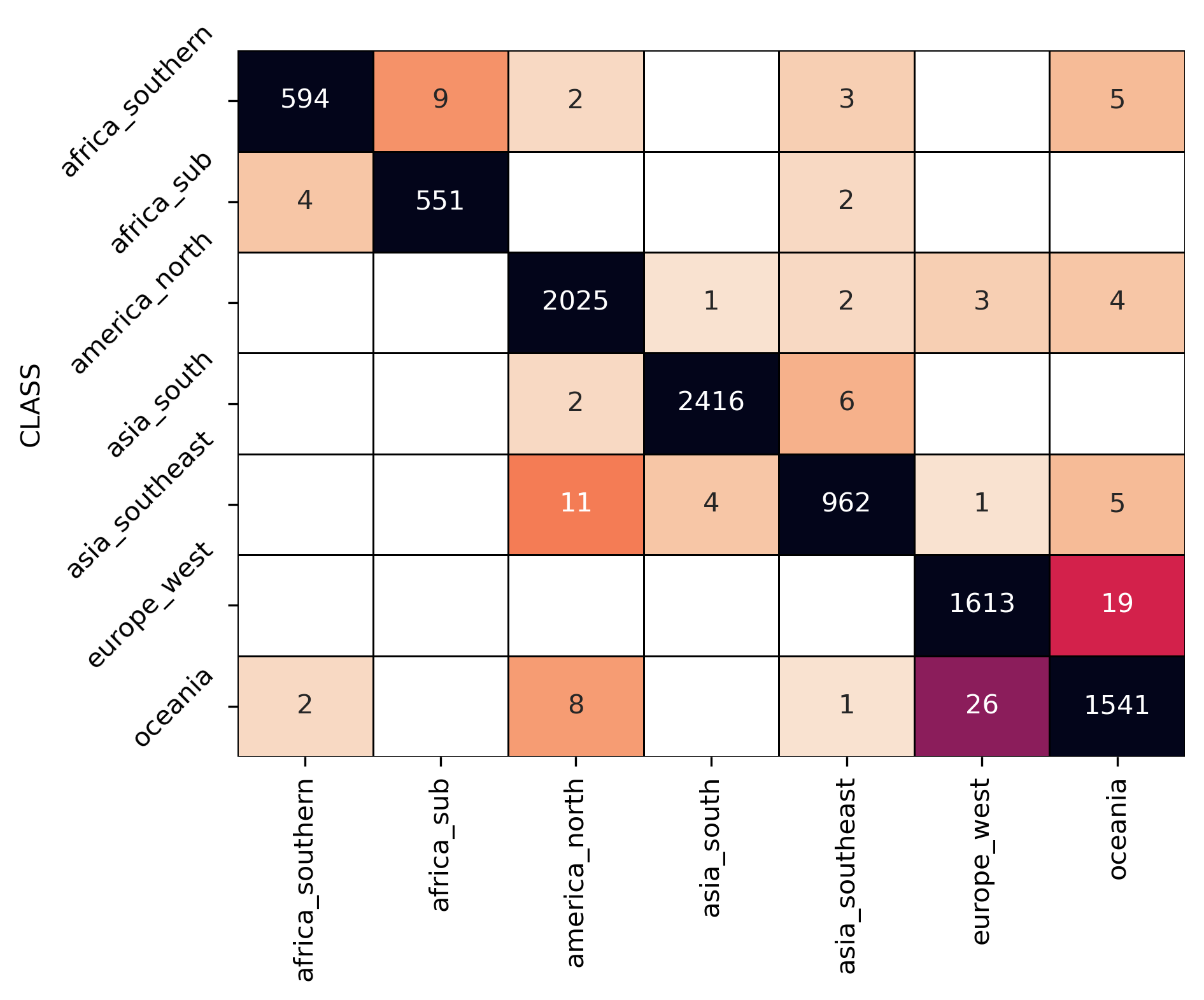}}
\caption{Distribution of Classification Errors for the Late-Stage Grammar with Regional Dialects. Each row represents samples from a place (the ground-truth) and each column represents the predicted label. Thus, the diagonal represents true positives.}
\end{figure*}

\begin{figure*}
\centering
\fbox{\includegraphics[width = 500pt]{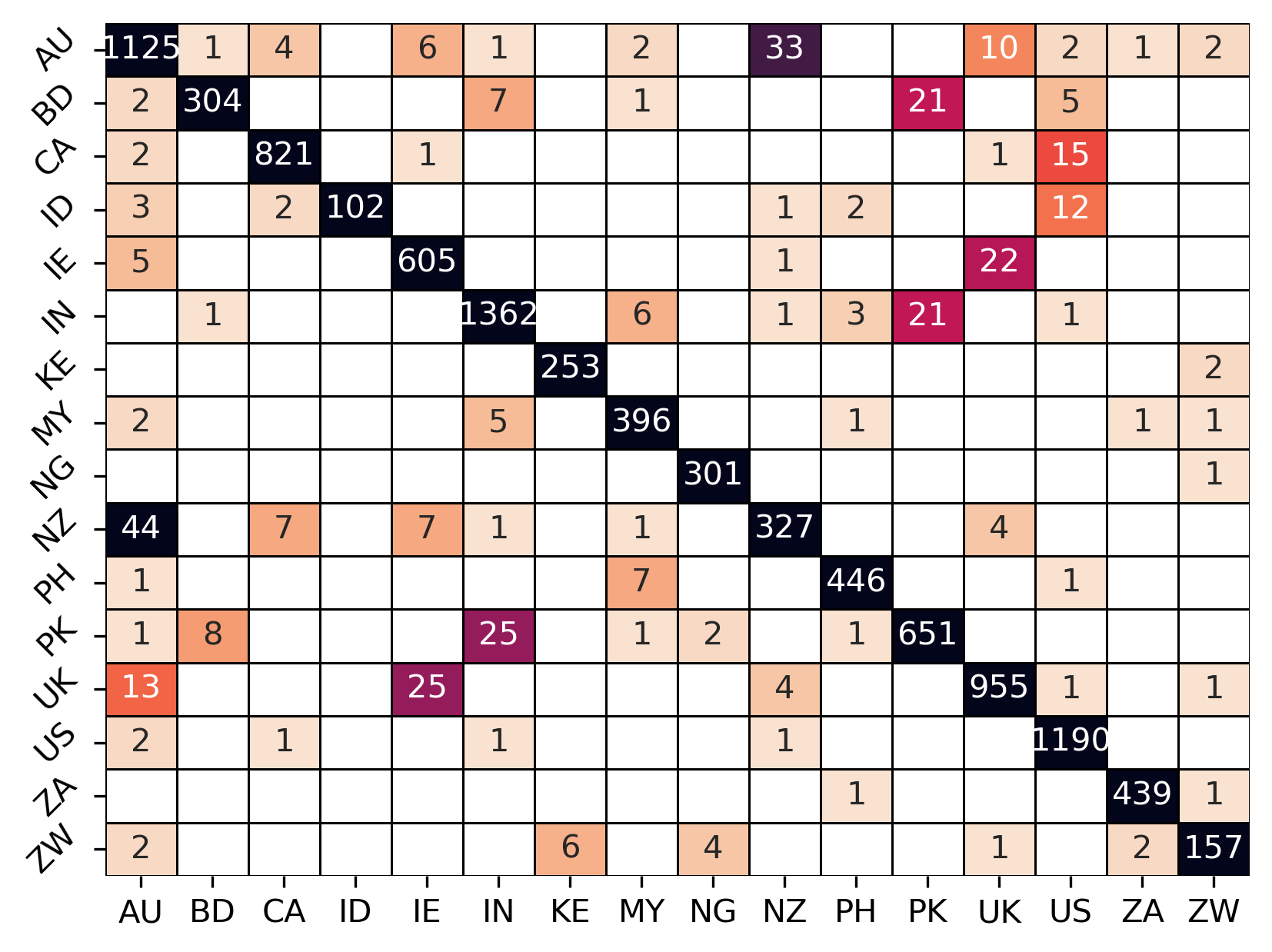}}
\caption{Distribution of Classification Errors for the Late-Stage Grammar with National Dialects. Each row represents samples from a place (the ground-truth) and each column represents the predicted label. Thus, the diagonal represents true positives.}
\end{figure*}

\begin{figure*}
\centering
\fbox{\includegraphics[width = 500pt]{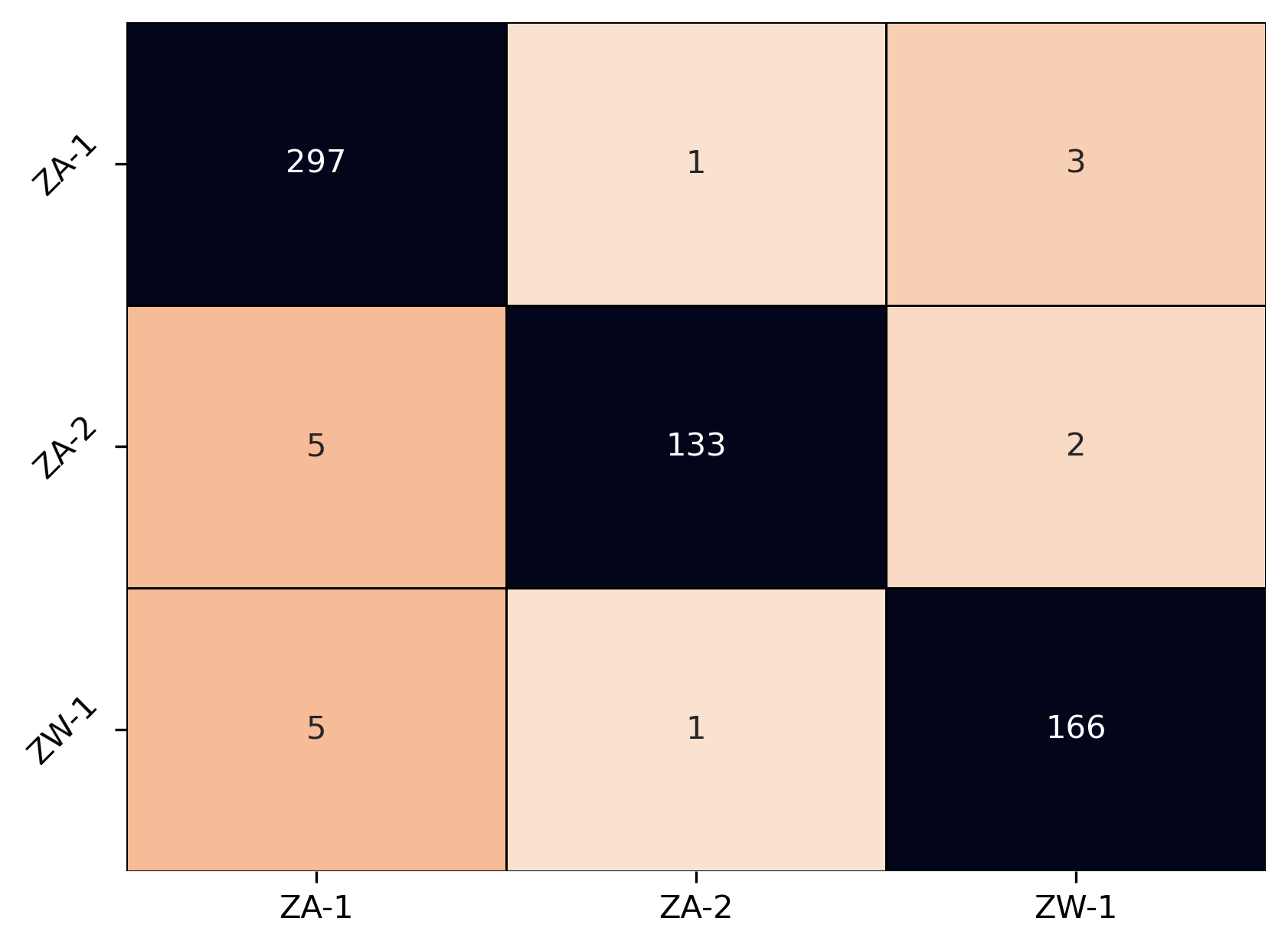}}
\caption{Distribution of Classification Errors for the Late-Stage Grammar with Local Dialects in Southern Africa. Each row represents samples from a place (the ground-truth) and each column represents the predicted label. Thus, the diagonal represents true positives.}
\end{figure*}

\begin{figure*}
\centering
\fbox{\includegraphics[width = 500pt]{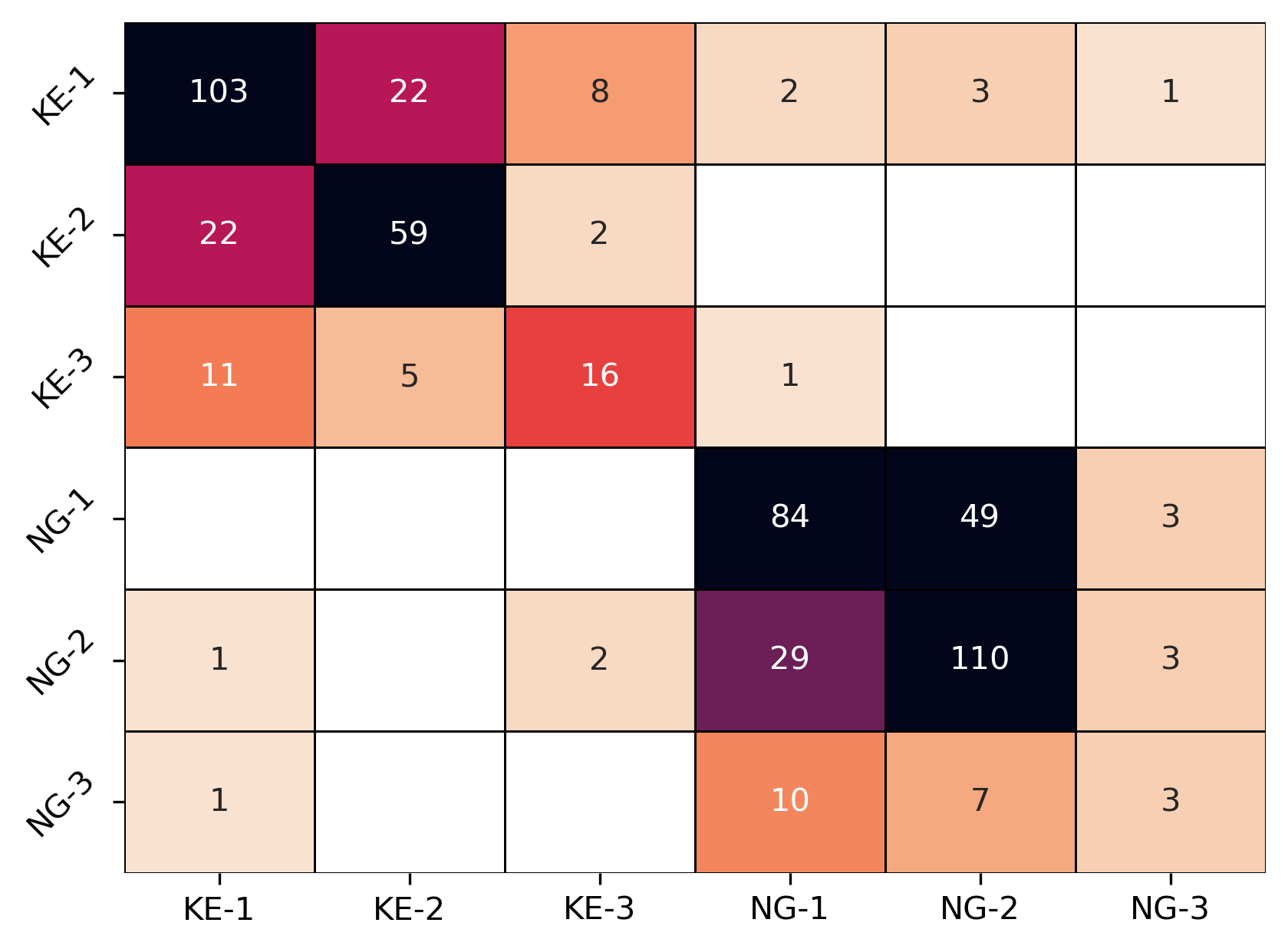}}
\caption{Distribution of Classification Errors for the Late-Stage Grammar with Local Dialects in Sub-Saharan Africa. Each row represents samples from a place (the ground-truth) and each column represents the predicted label. Thus, the diagonal represents true positives.}
\end{figure*}

\begin{figure*}
\centering
\fbox{\includegraphics[width = 500pt]{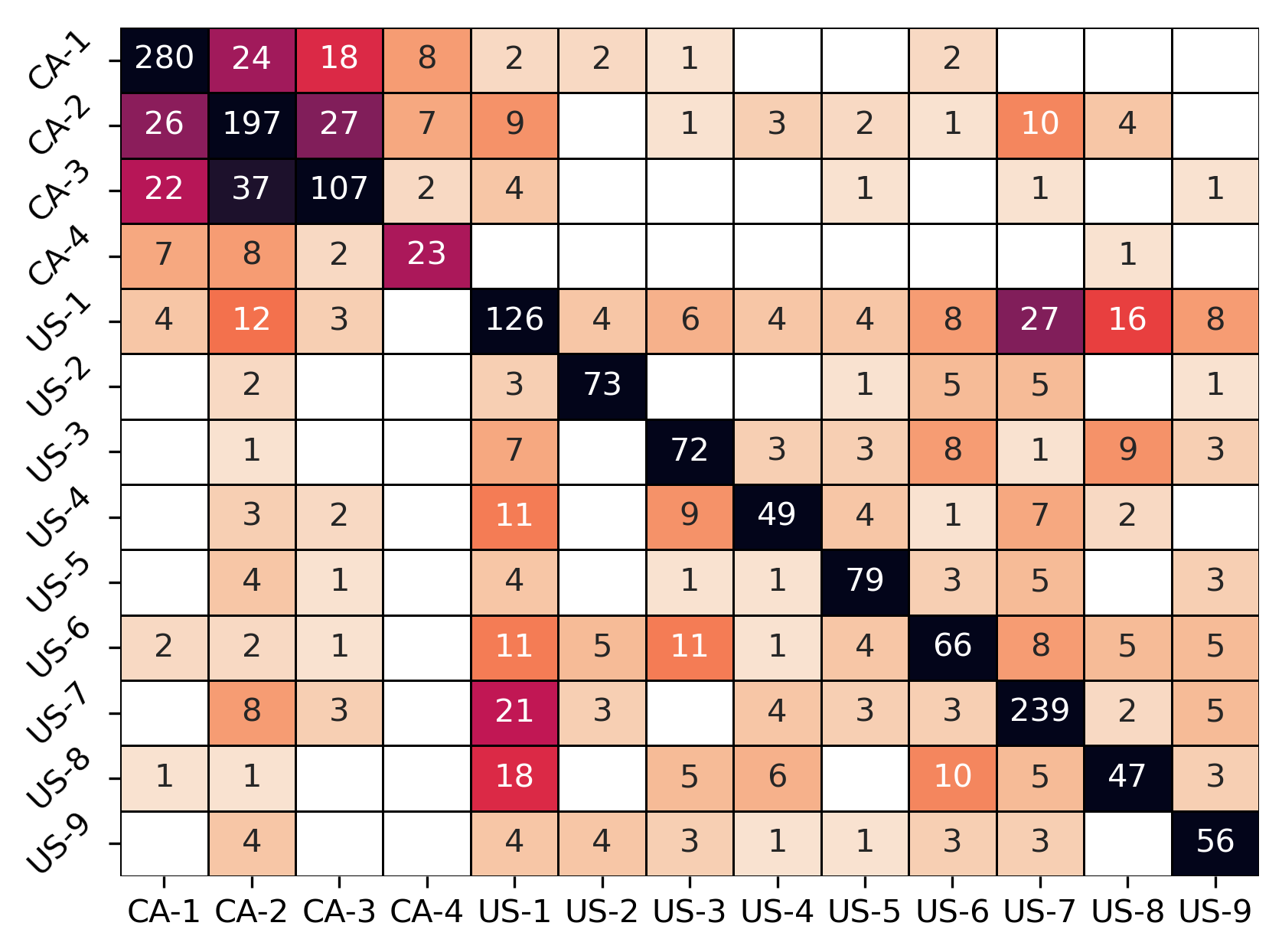}}
\caption{Distribution of Classification Errors for the Late-Stage Grammar with Local Dialects in North America. Each row represents samples from a place (the ground-truth) and each column represents the predicted label. Thus, the diagonal represents true positives.}
\end{figure*}

\begin{figure*}
\centering
\fbox{\includegraphics[width = 500pt]{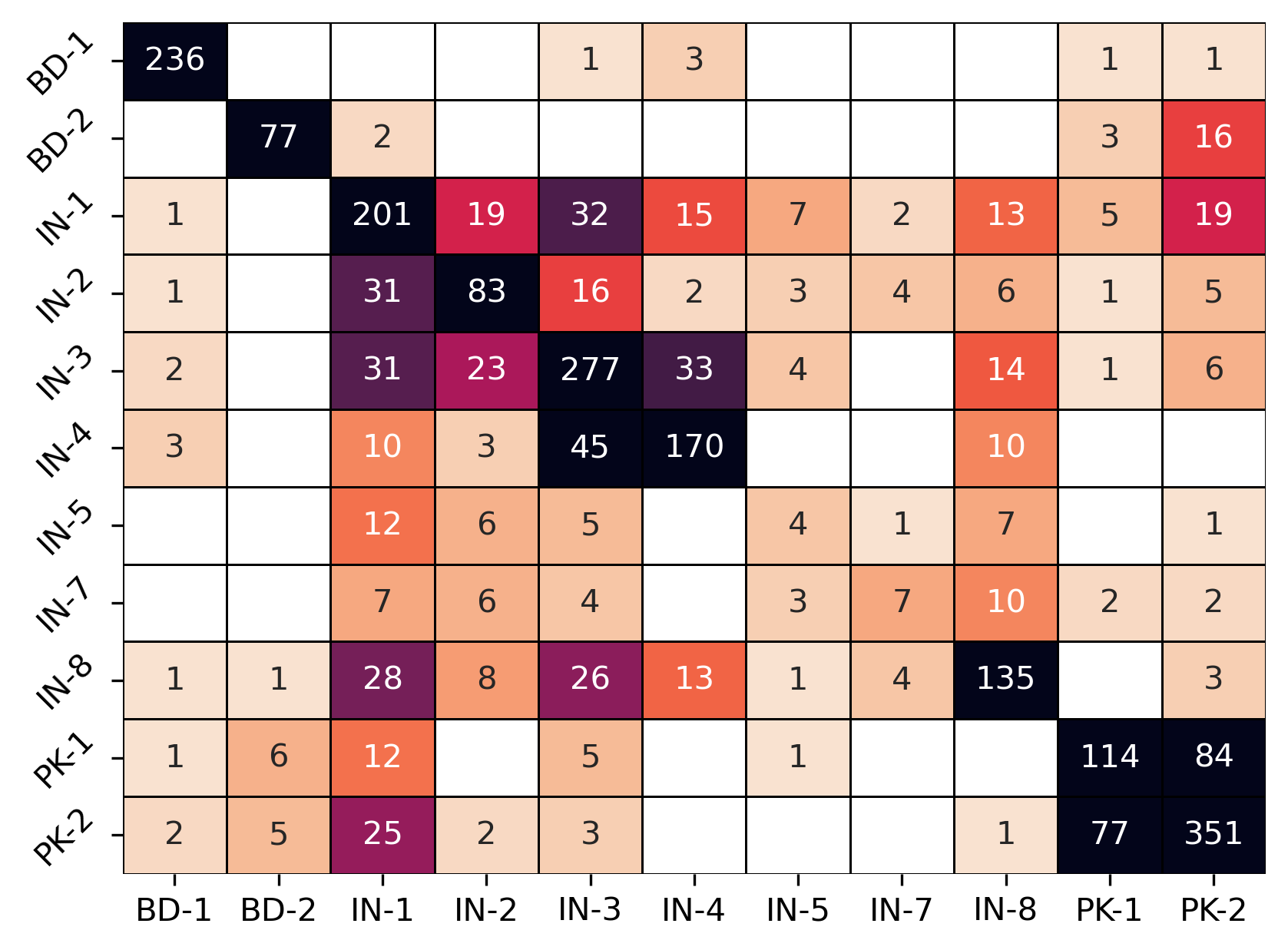}}
\caption{Distribution of Classification Errors for the Late-Stage Grammar with Local Dialects in South Asia. Each row represents samples from a place (the ground-truth) and each column represents the predicted label. Thus, the diagonal represents true positives.}
\end{figure*}

\begin{figure*}
\centering
\fbox{\includegraphics[width = 500pt]{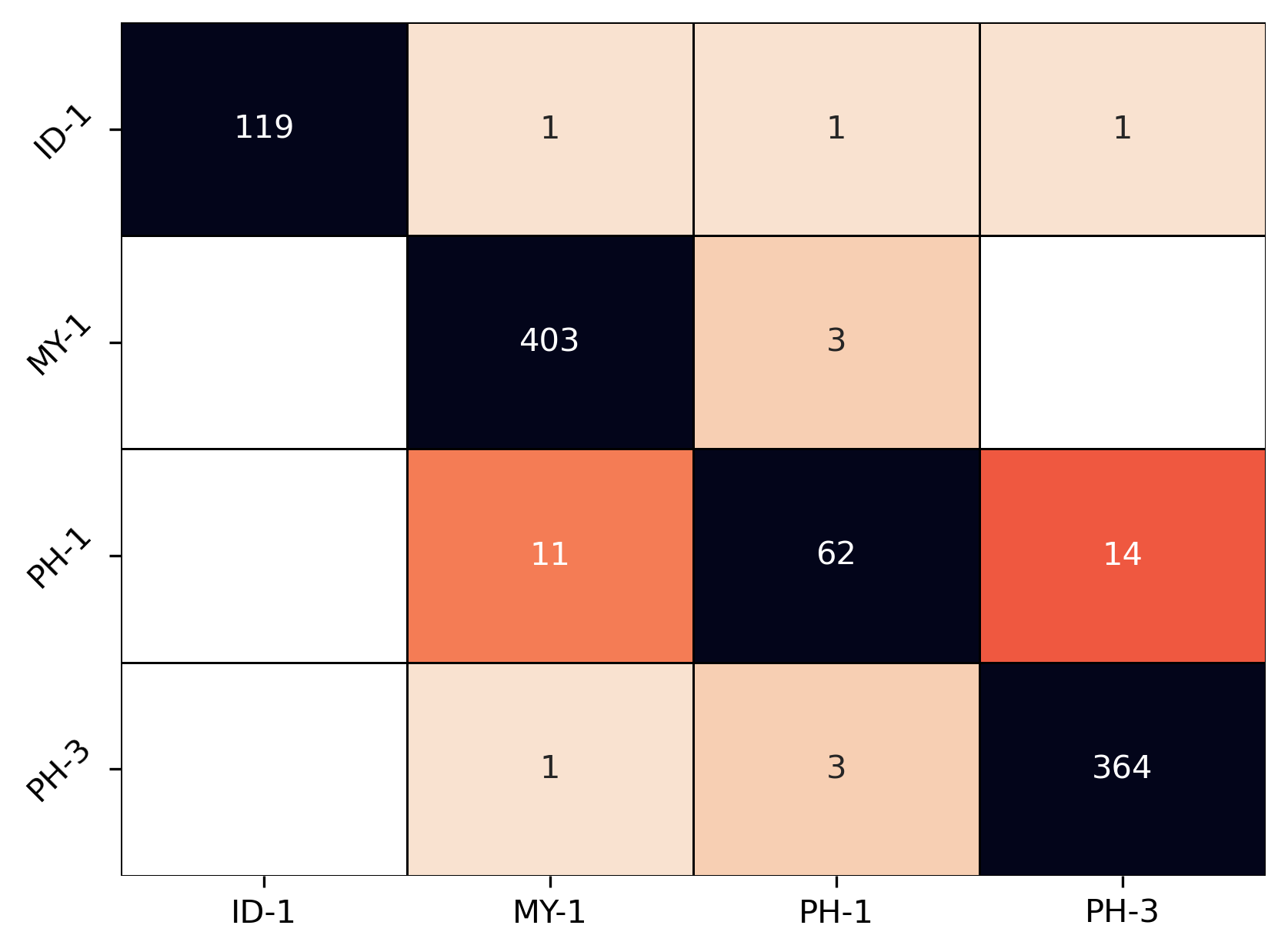}}
\caption{Distribution of Classification Errors for the Late-Stage Grammar with Local Dialects in Southeast Asia. Each row represents samples from a place (the ground-truth) and each column represents the predicted label. Thus, the diagonal represents true positives.}
\end{figure*}

\begin{figure*}
\centering
\fbox{\includegraphics[width = 500pt]{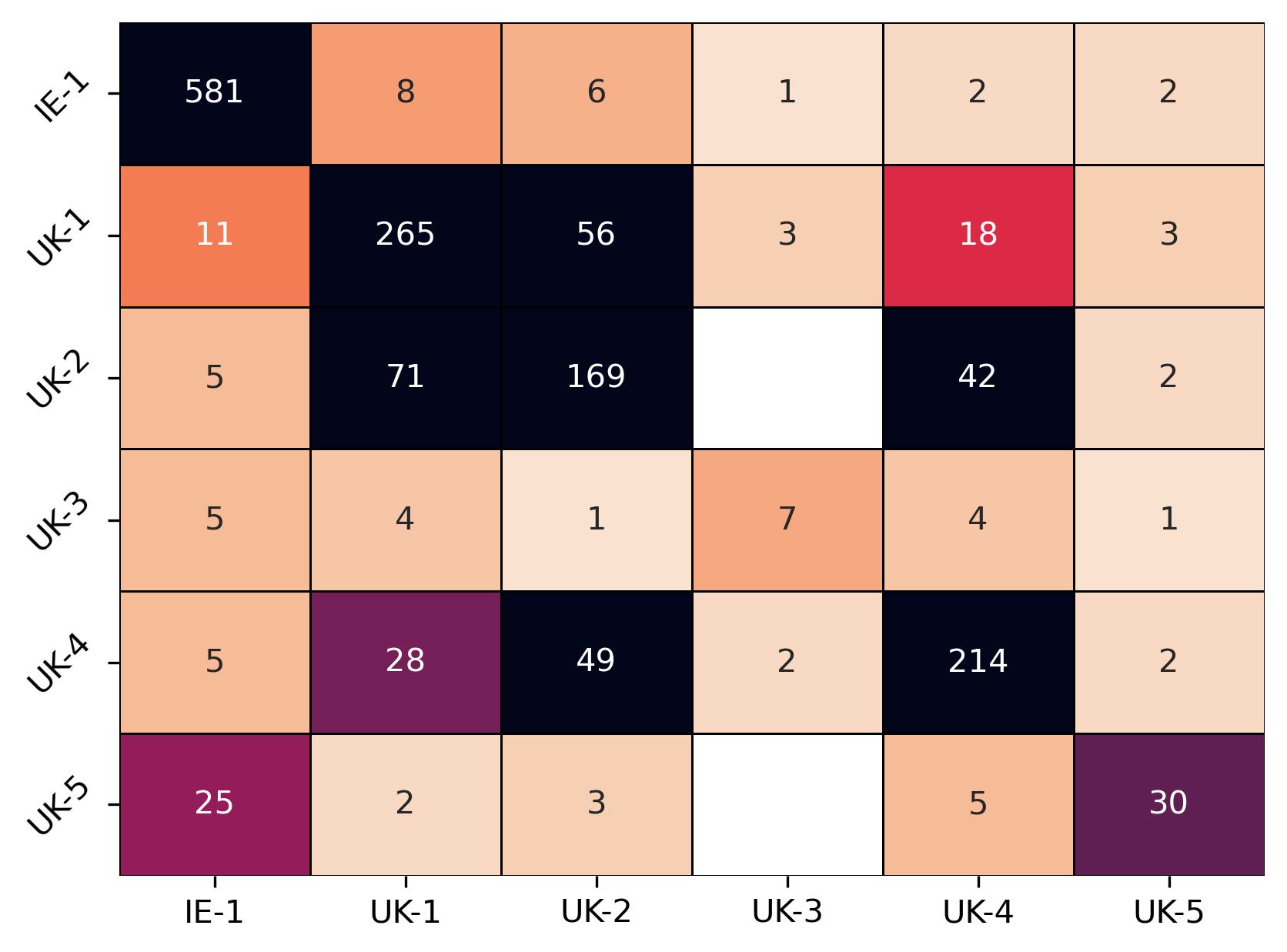}}
\caption{Distribution of Classification Errors for the Late-Stage Grammar with Local Dialects in Western Europe. Each row represents samples from a place (the ground-truth) and each column represents the predicted label. Thus, the diagonal represents true positives.}
\end{figure*}

\begin{figure*}
\centering
\fbox{\includegraphics[width = 500pt]{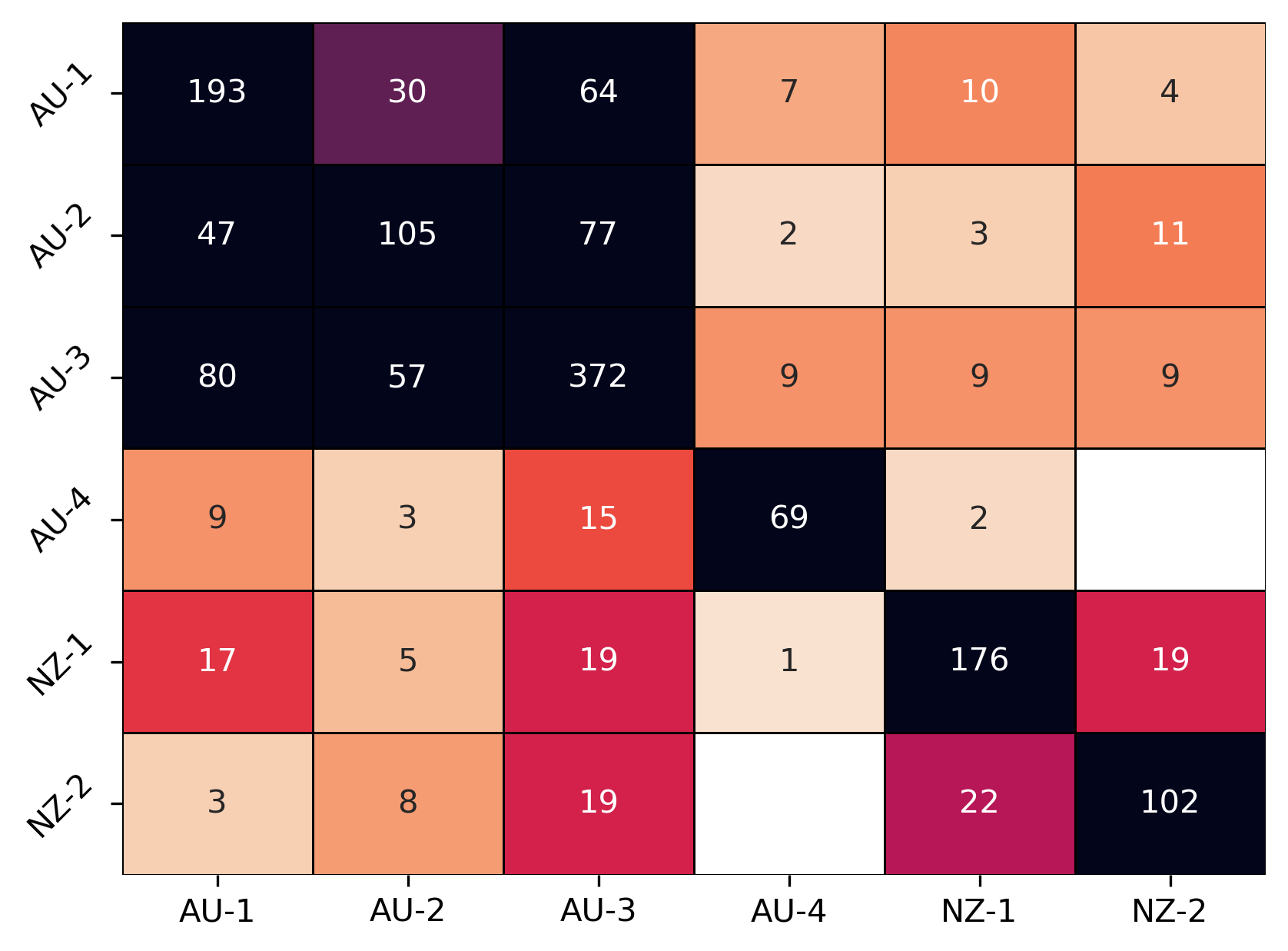}}
\caption{Distribution of Classification Errors for the Late-Stage Grammar with Local Dialects in Oceania. Each row represents samples from a place (the ground-truth) and each column represents the predicted label. Thus, the diagonal represents true positives.}
\end{figure*}

\begin{figure*}
\centering
\fbox{\includegraphics[width = 500pt]{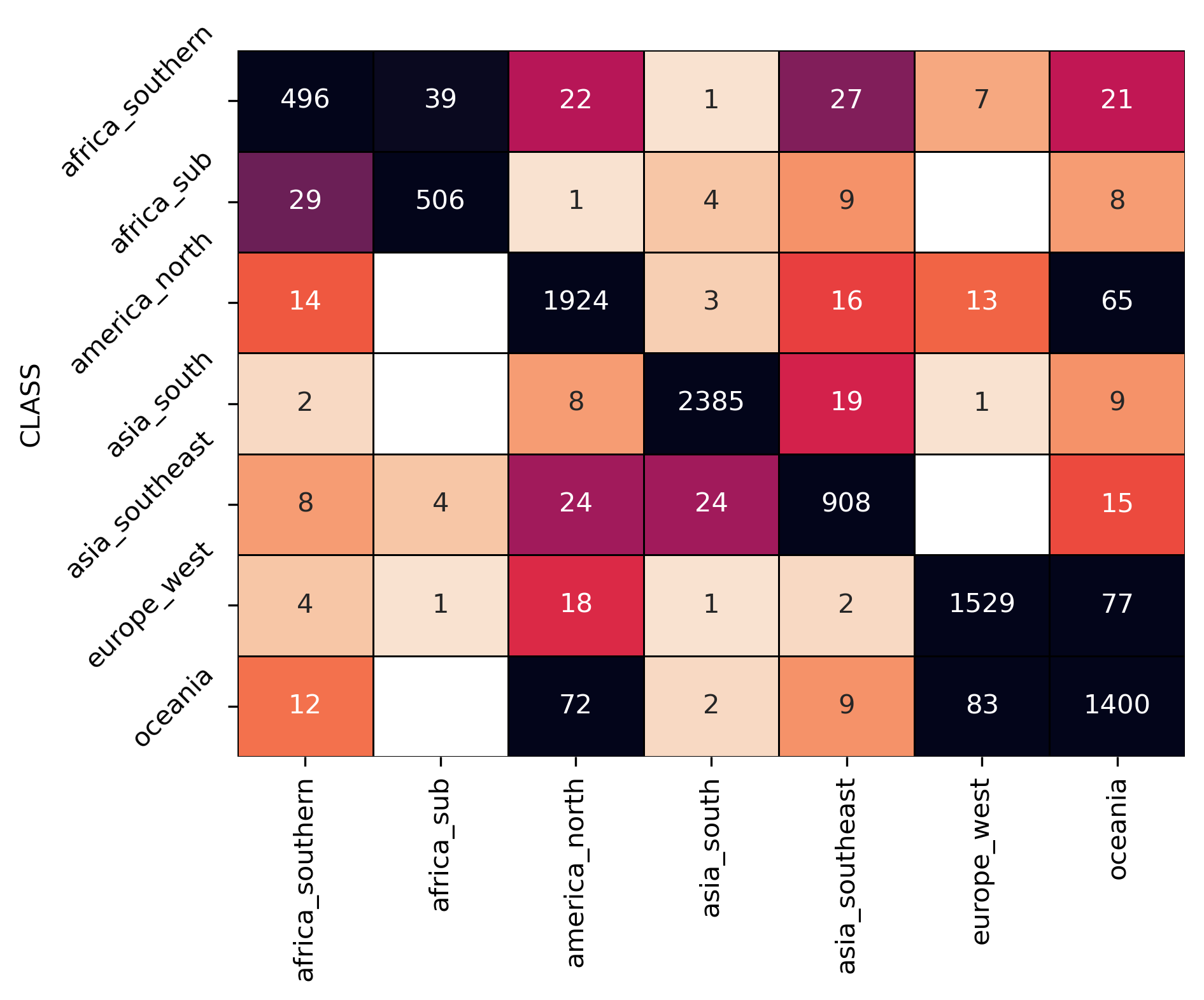}}
\caption{Distribution of Classification Errors for the Early-Stage Grammar with Regional Dialects. Each row represents samples from a place (the ground-truth) and each column represents the predicted label. Thus, the diagonal represents true positives.}
\end{figure*}

\begin{figure*}
\centering
\fbox{\includegraphics[width = 500pt]{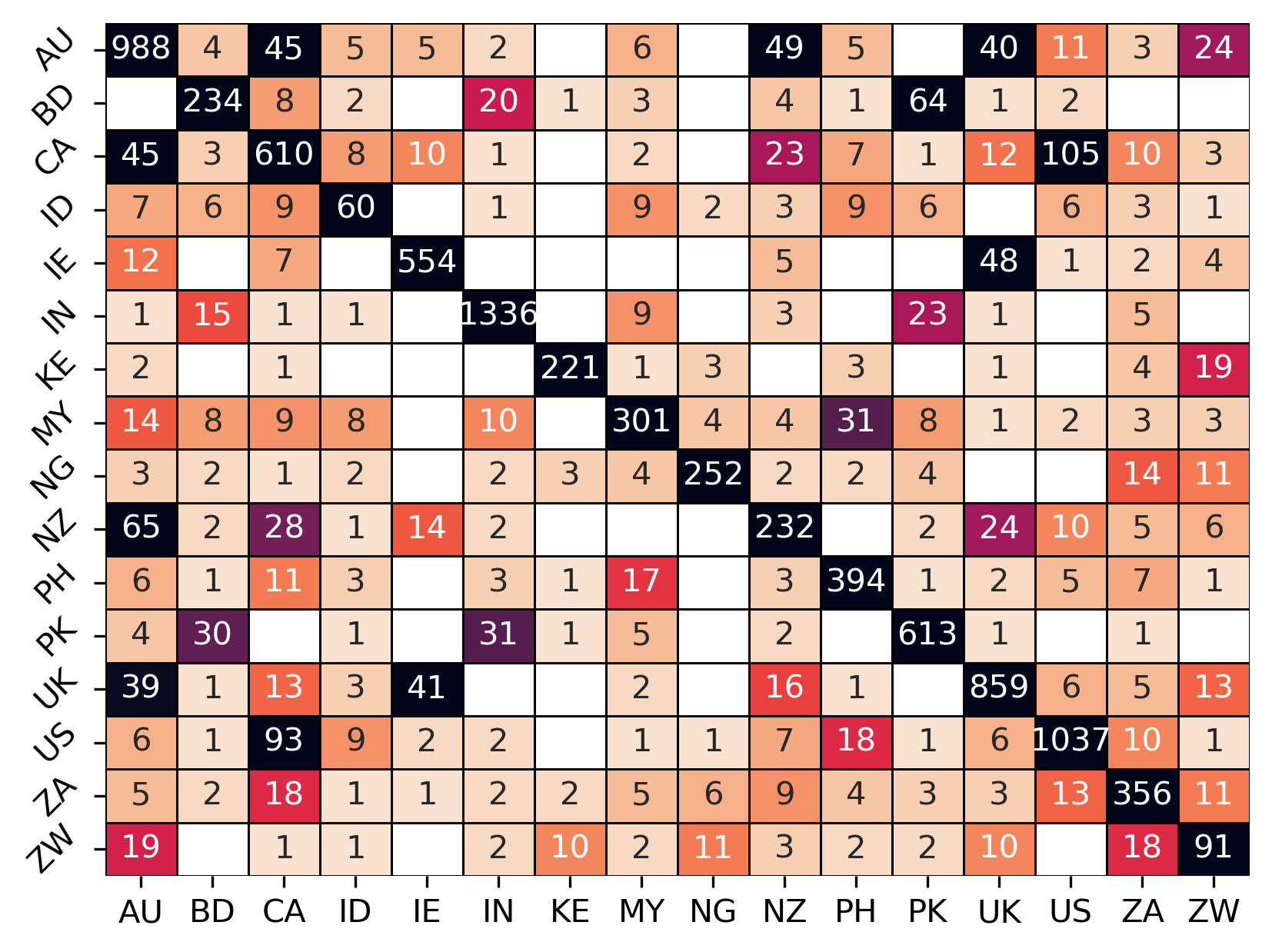}}
\caption{Distribution of Classification Errors for the Early-Stage Grammar with National Dialects. Each row represents samples from a place (the ground-truth) and each column represents the predicted label. Thus, the diagonal represents true positives.}
\end{figure*}

\begin{figure*}
\centering
\fbox{\includegraphics[width = 500pt]{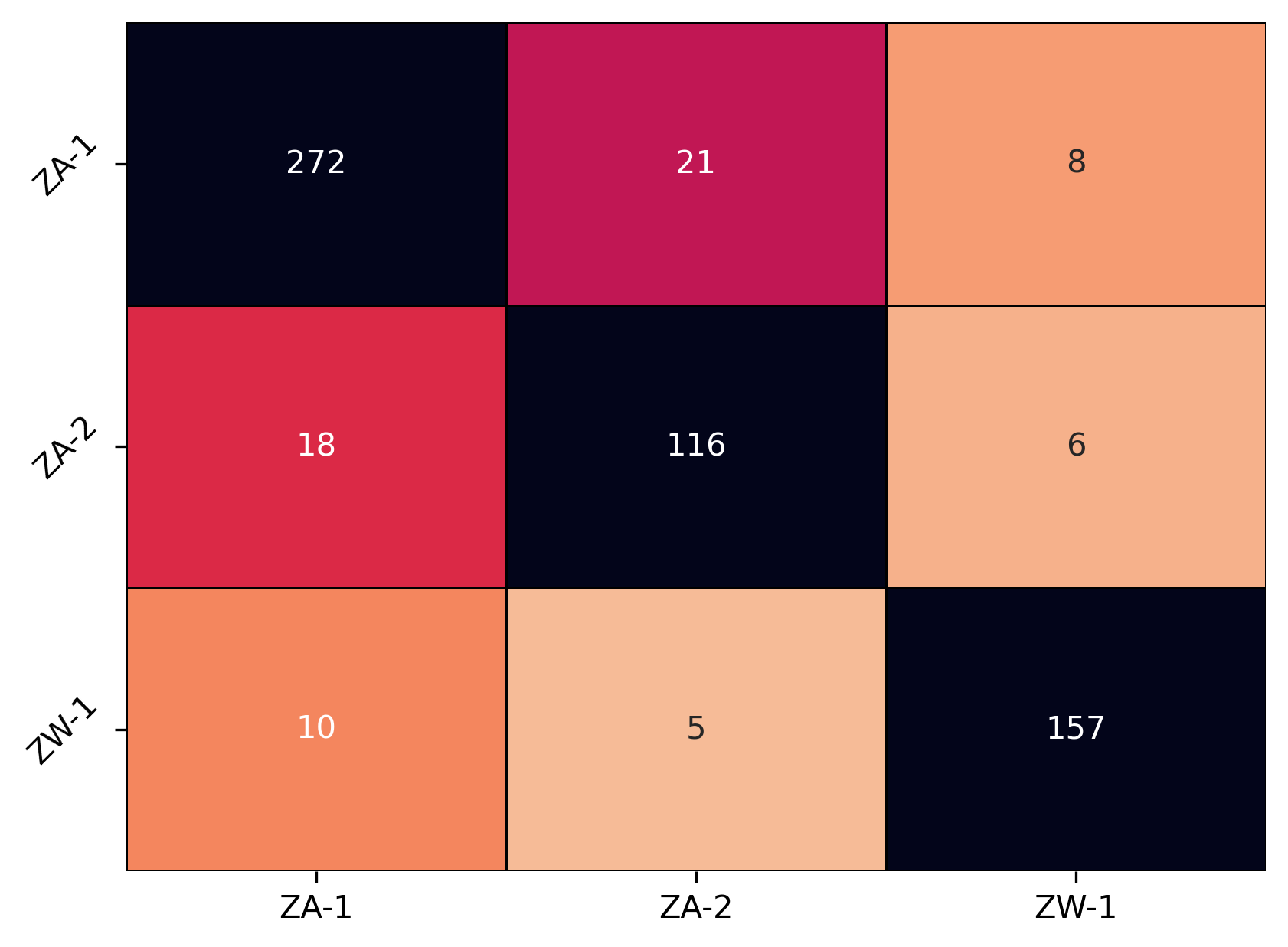}}
\caption{Distribution of Classification Errors for the Early-Stage Grammar with Local Dialects in Southern Africa. Each row represents samples from a place (the ground-truth) and each column represents the predicted label. Thus, the diagonal represents true positives.}
\end{figure*}

\begin{figure*}
\centering
\fbox{\includegraphics[width = 500pt]{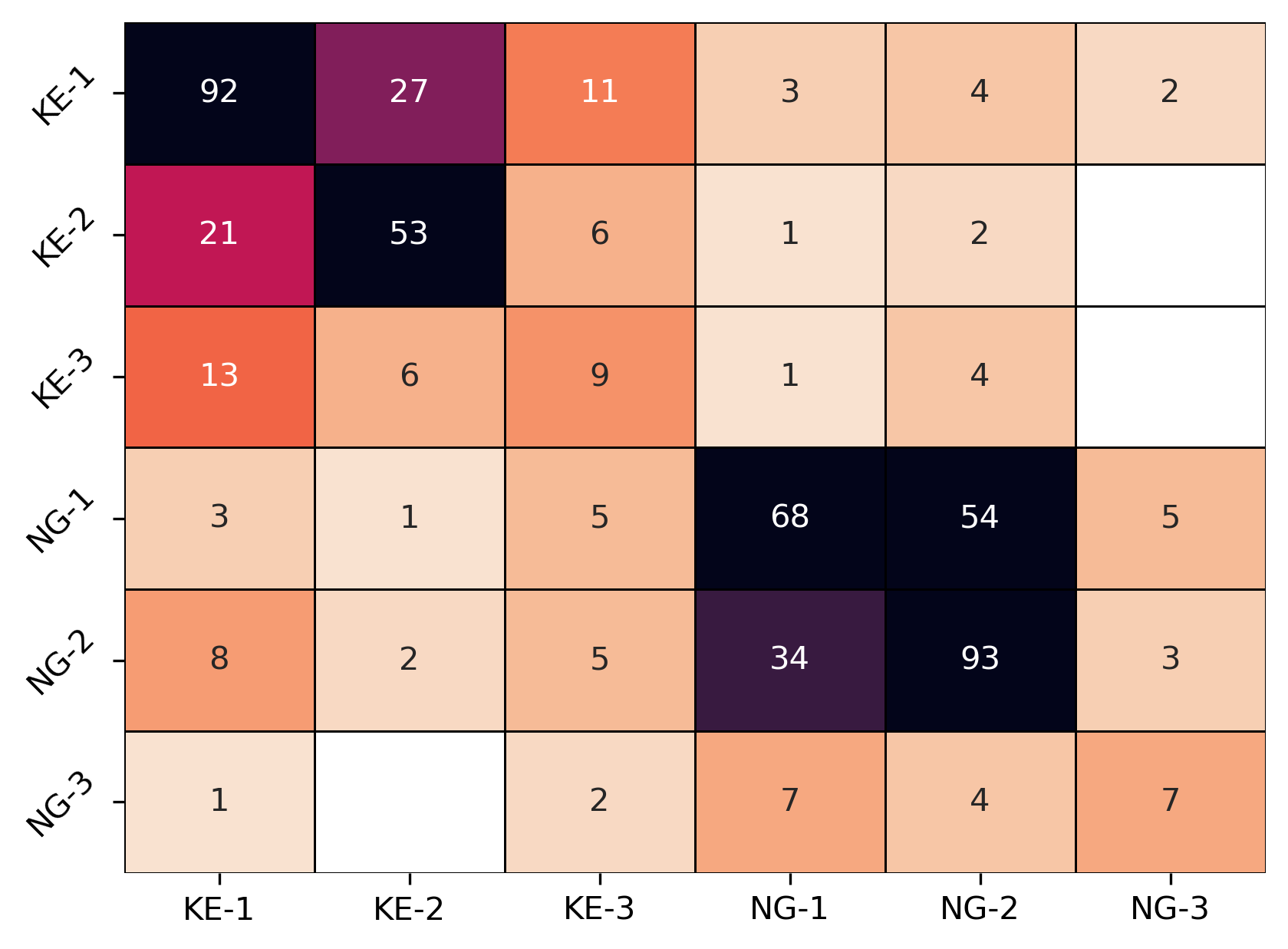}}
\caption{Distribution of Classification Errors for the Early-Stage Grammar with Local Dialects in Sub-Saharan Africa. Each row represents samples from a place (the ground-truth) and each column represents the predicted label. Thus, the diagonal represents true positives.}
\end{figure*}

\begin{figure*}
\centering
\fbox{\includegraphics[width = 500pt]{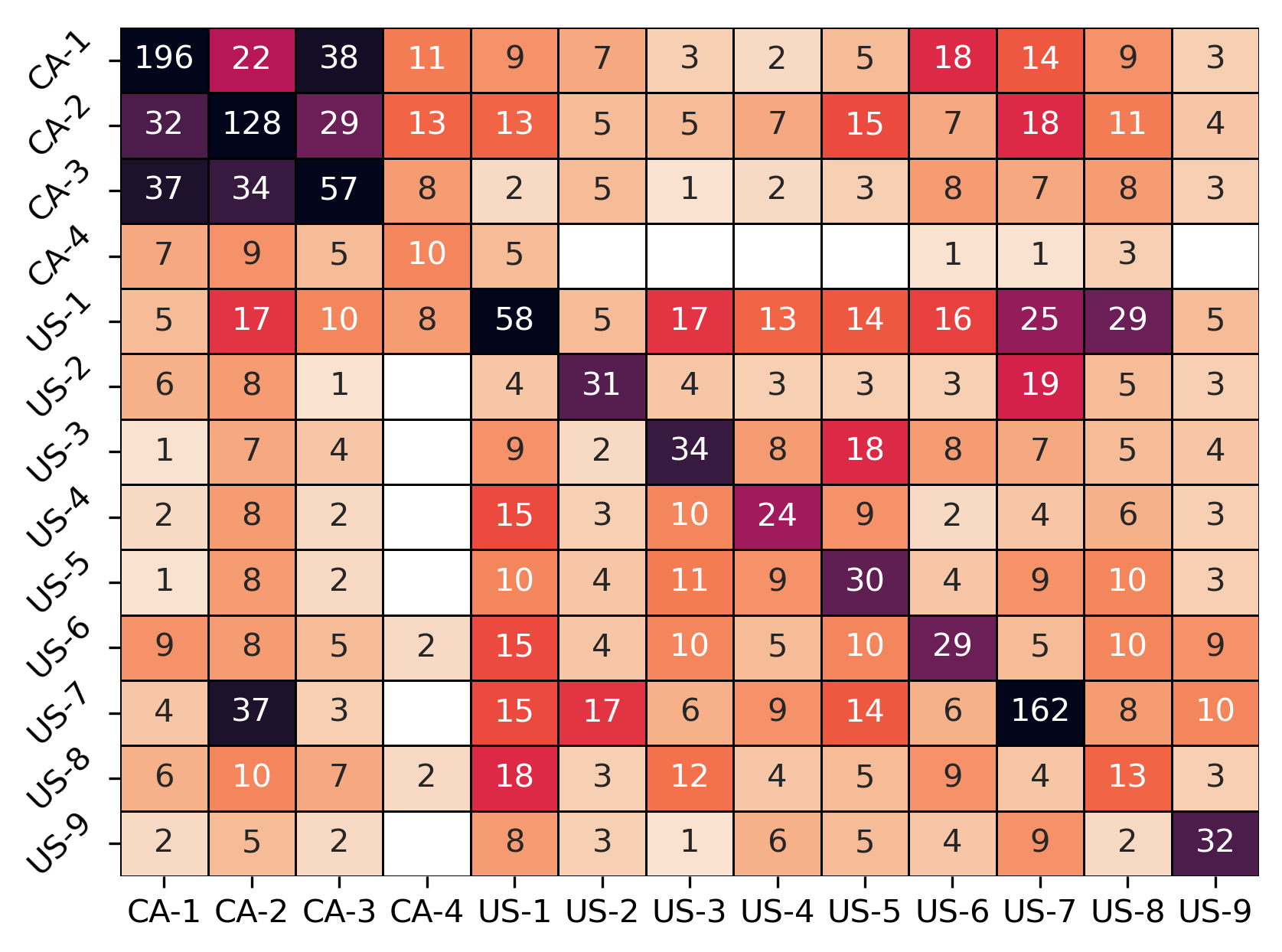}}
\caption{Distribution of Classification Errors for the Early-Stage Grammar with Local Dialects in North America. Each row represents samples from a place (the ground-truth) and each column represents the predicted label. Thus, the diagonal represents true positives.}
\end{figure*}

\begin{figure*}
\centering
\fbox{\includegraphics[width = 500pt]{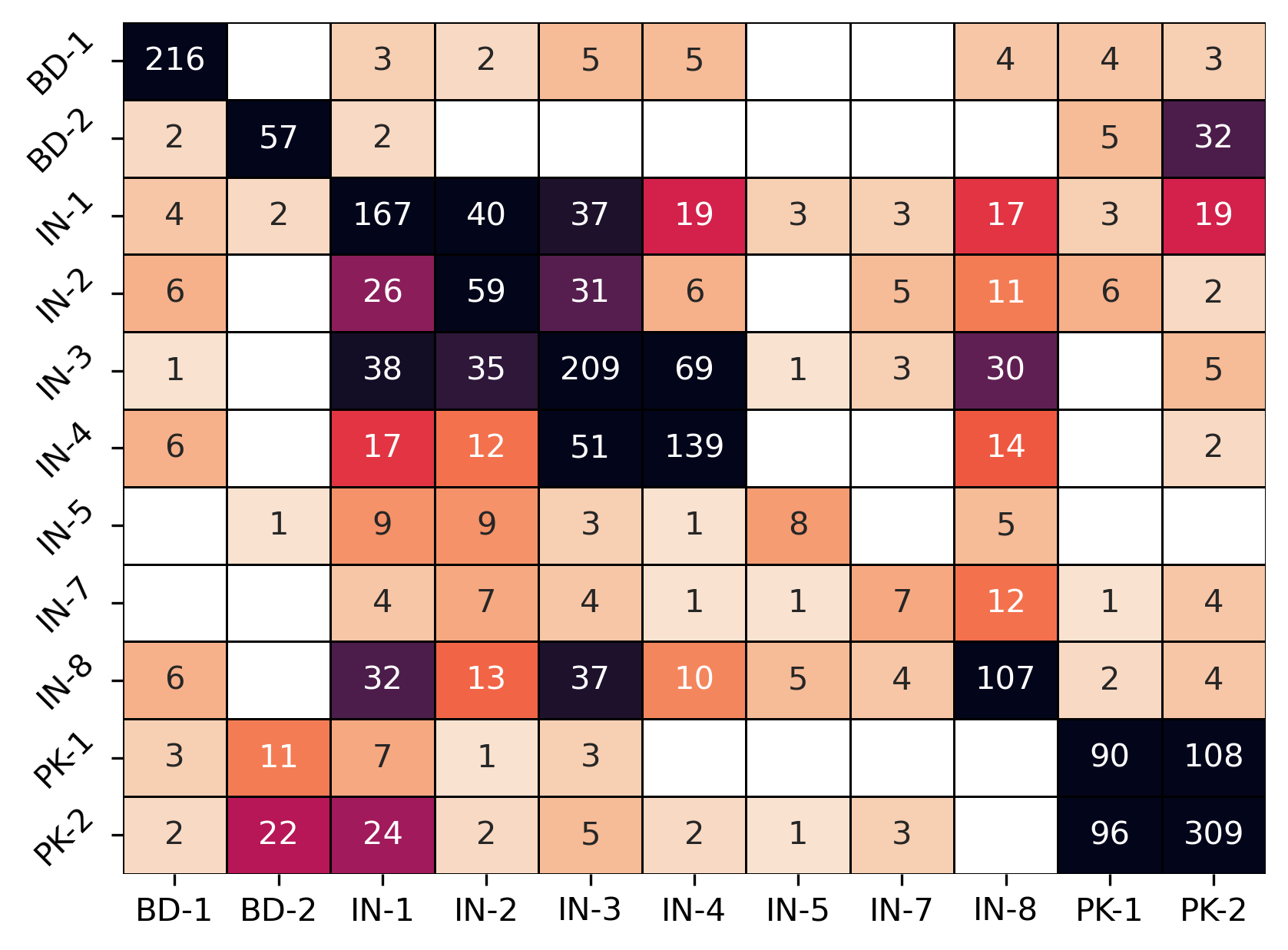}}
\caption{Distribution of Classification Errors for the Early-Stage Grammar with Local Dialects in South Asia. Each row represents samples from a place (the ground-truth) and each column represents the predicted label. Thus, the diagonal represents true positives.}
\end{figure*}

\begin{figure*}
\centering
\fbox{\includegraphics[width = 500pt]{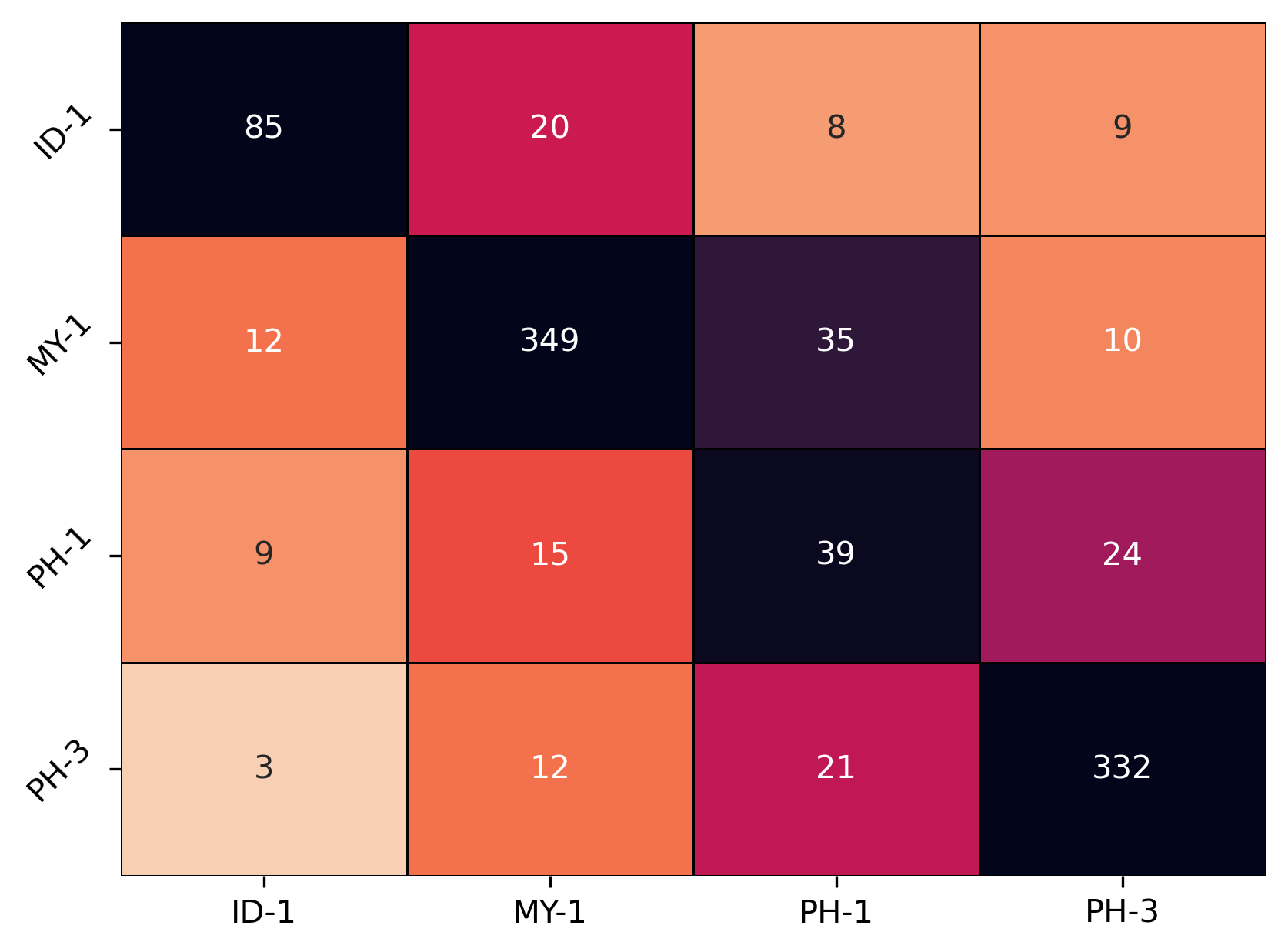}}
\caption{Distribution of Classification Errors for the Early-Stage Grammar with Local Dialects in Southeast Asia. Each row represents samples from a place (the ground-truth) and each column represents the predicted label. Thus, the diagonal represents true positives.}
\end{figure*}

\begin{figure*}
\centering
\fbox{\includegraphics[width = 500pt]{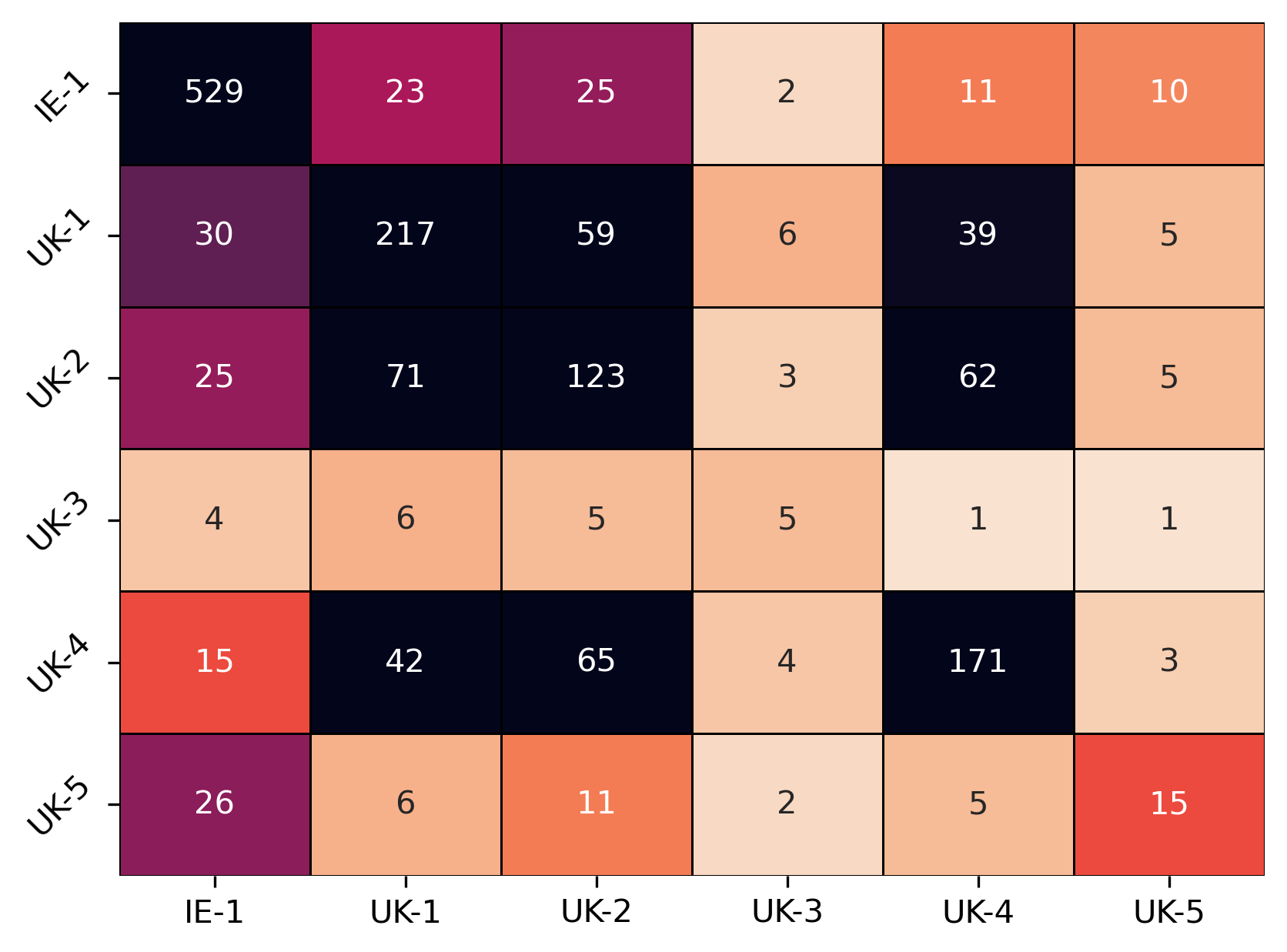}}
\caption{Distribution of Classification Errors for the Early-Stage Grammar with Local Dialects in Western Europe. Each row represents samples from a place (the ground-truth) and each column represents the predicted label. Thus, the diagonal represents true positives.}
\end{figure*}

\begin{figure*}
\centering
\fbox{\includegraphics[width = 500pt]{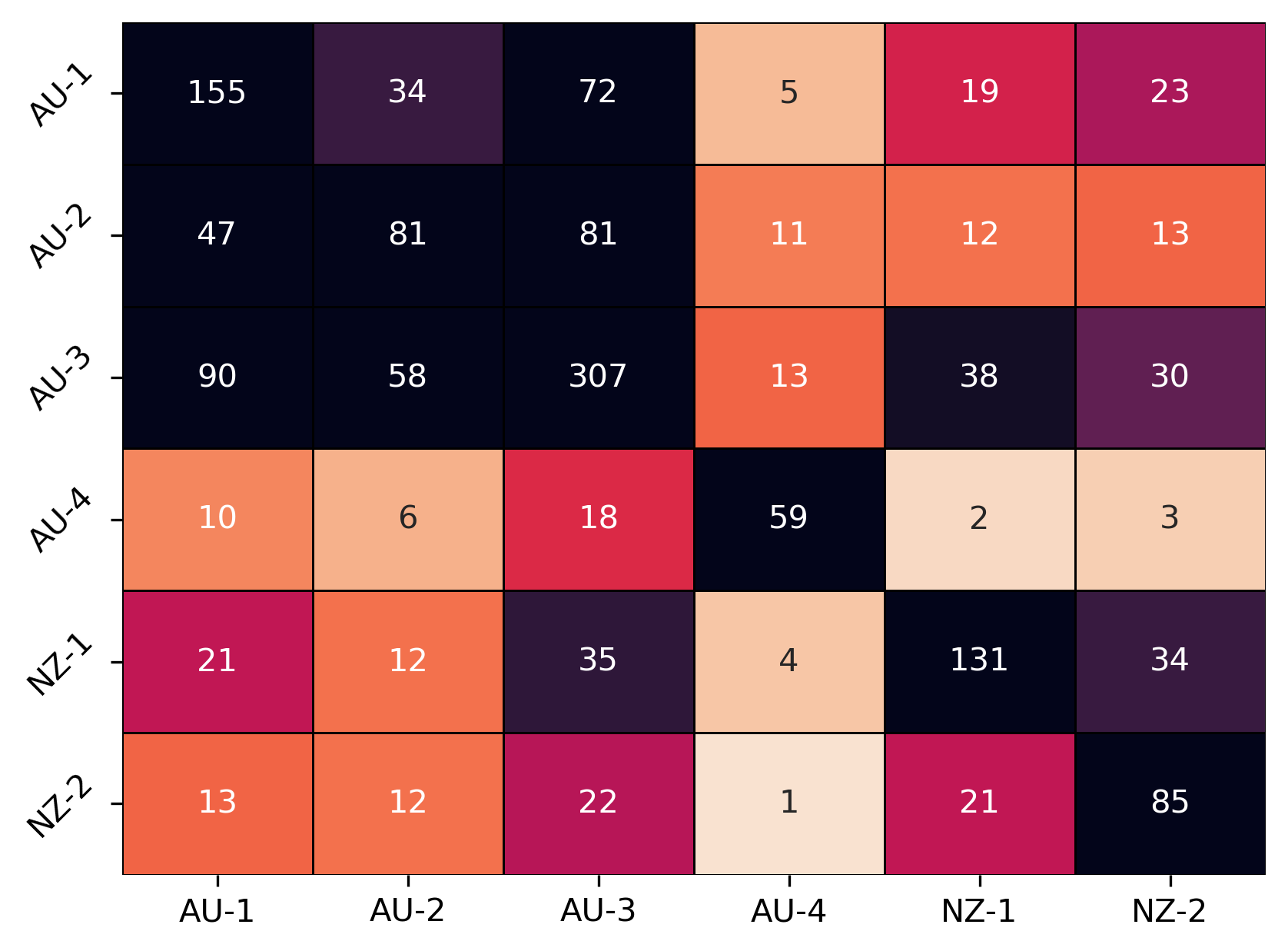}}
\caption{Distribution of Classification Errors for the Early-Stage Grammar with Local Dialects in Oceania. Each row represents samples from a place (the ground-truth) and each column represents the predicted label. Thus, the diagonal represents true positives.}
\end{figure*}

\begin{figure*}
\centering
\fbox{\includegraphics[width = 500pt]{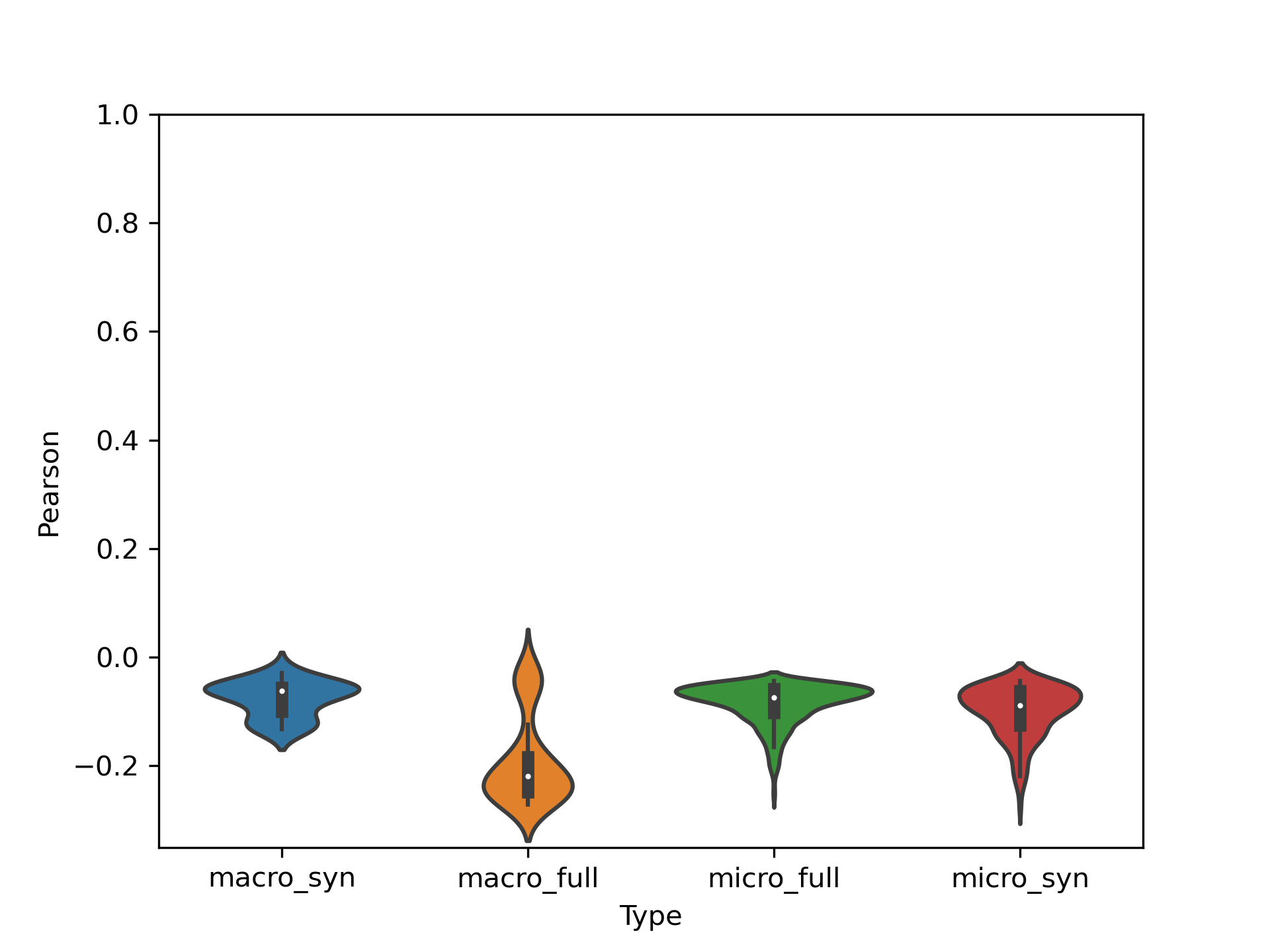}}
\caption{Correlation of Error Distribution Between Late-Stage Grammar and Nodes within the Grammar for Regional Dialects. High Correlation indicates that the same dialects are similar in each model type while low correlation indicates that the relationships between dialects for a given component of the grammar differ from the late-stage grammar.}
\end{figure*}

\begin{figure*}
\centering
\fbox{\includegraphics[width = 500pt]{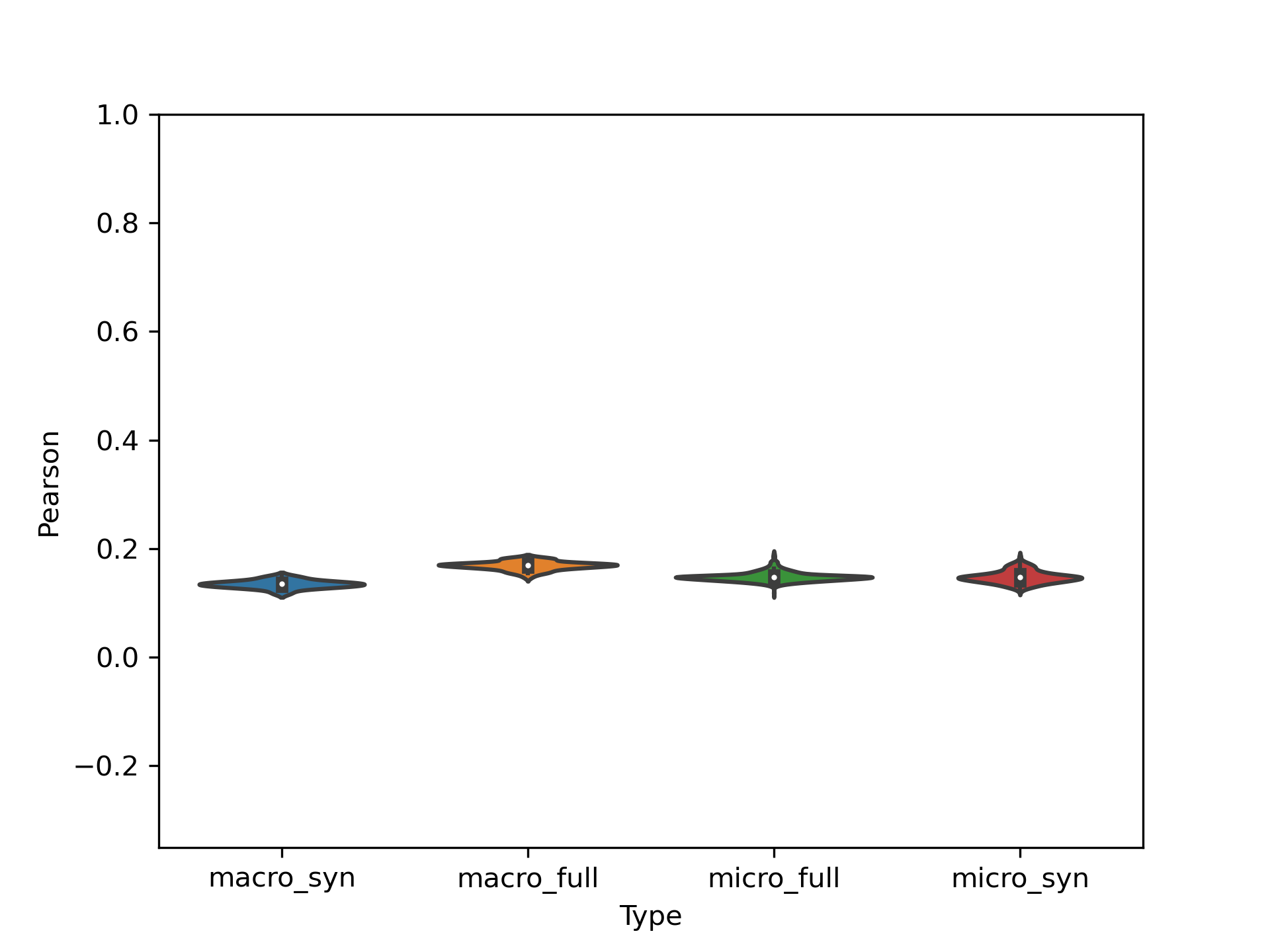}}
\caption{Correlation of Error Distribution Between Late-Stage Grammar and Nodes within the Grammar for National Dialects. High Correlation indicates that the same dialects are similar in each model type while low correlation indicates that the relationships between dialects for a given component of the grammar differ from the late-stage grammar.}
\end{figure*}

\begin{figure*}
\centering
\fbox{\includegraphics[width = 500pt]{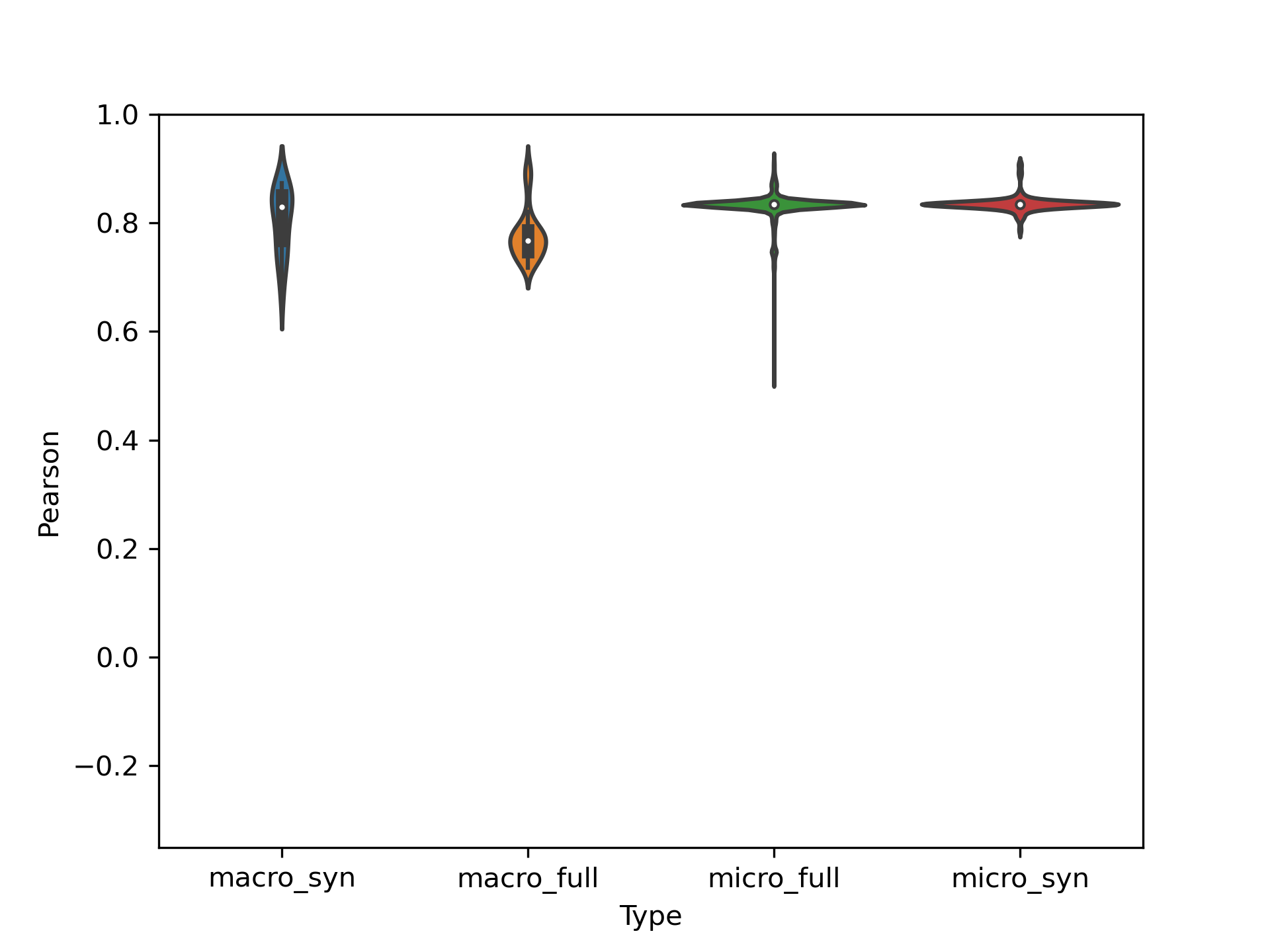}}
\caption{Correlation of Error Distribution Between Late-Stage Grammar and Nodes within the Grammar for Local Dialects in Southern Africa. High Correlation indicates that the same dialects are similar in each model type while low correlation indicates that the relationships between dialects for a given component of the grammar differ from the late-stage grammar.}
\end{figure*}

\begin{figure*}
\centering
\fbox{\includegraphics[width = 500pt]{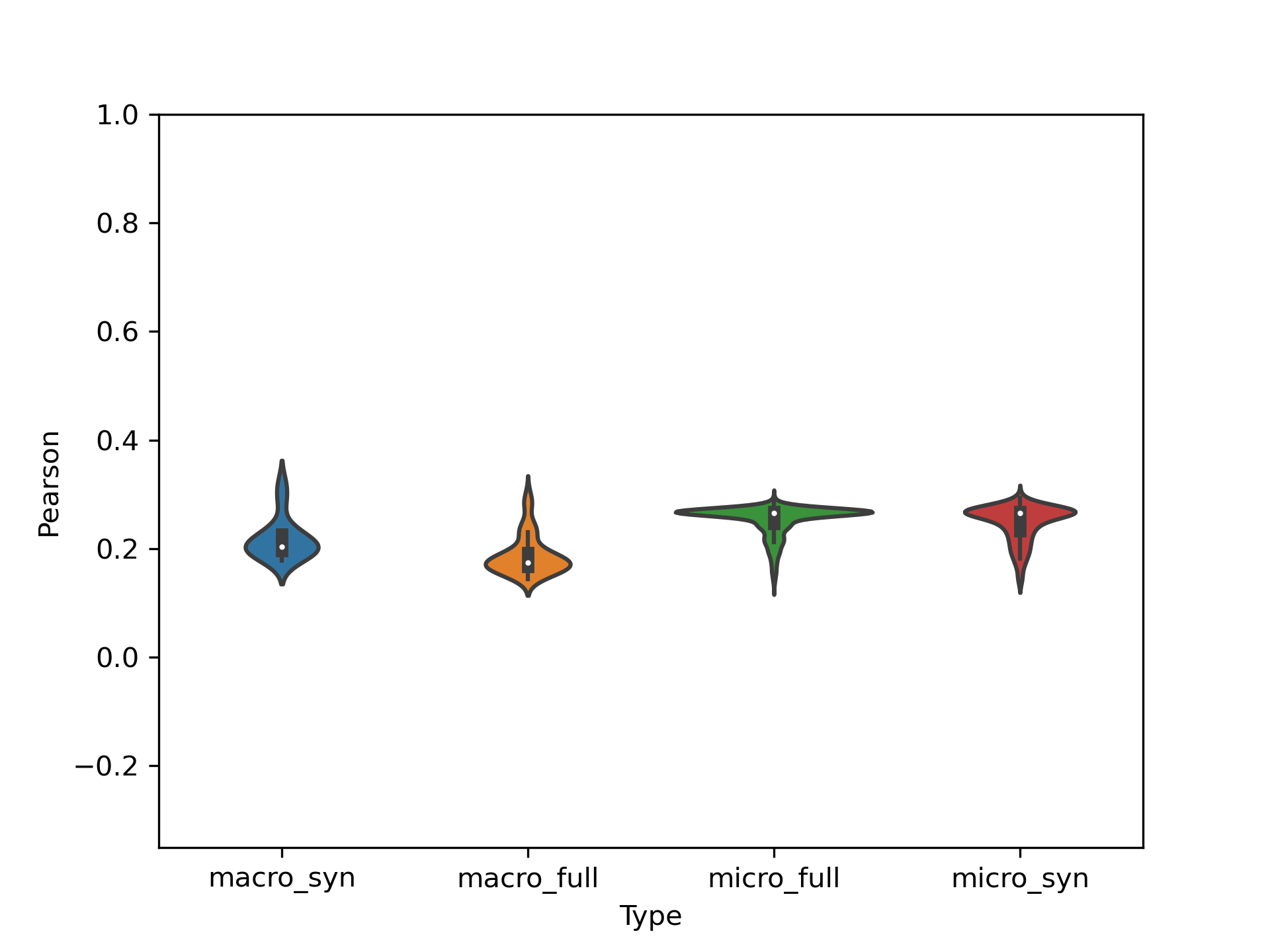}}
\caption{Correlation of Error Distribution Between Late-Stage Grammar and Nodes within the Grammar for Local Dialects in Sub-Saharan Africa. High Correlation indicates that the same dialects are similar in each model type while low correlation indicates that the relationships between dialects for a given component of the grammar differ from the late-stage grammar.}
\end{figure*}

\begin{figure*}
\centering
\fbox{\includegraphics[width = 500pt]{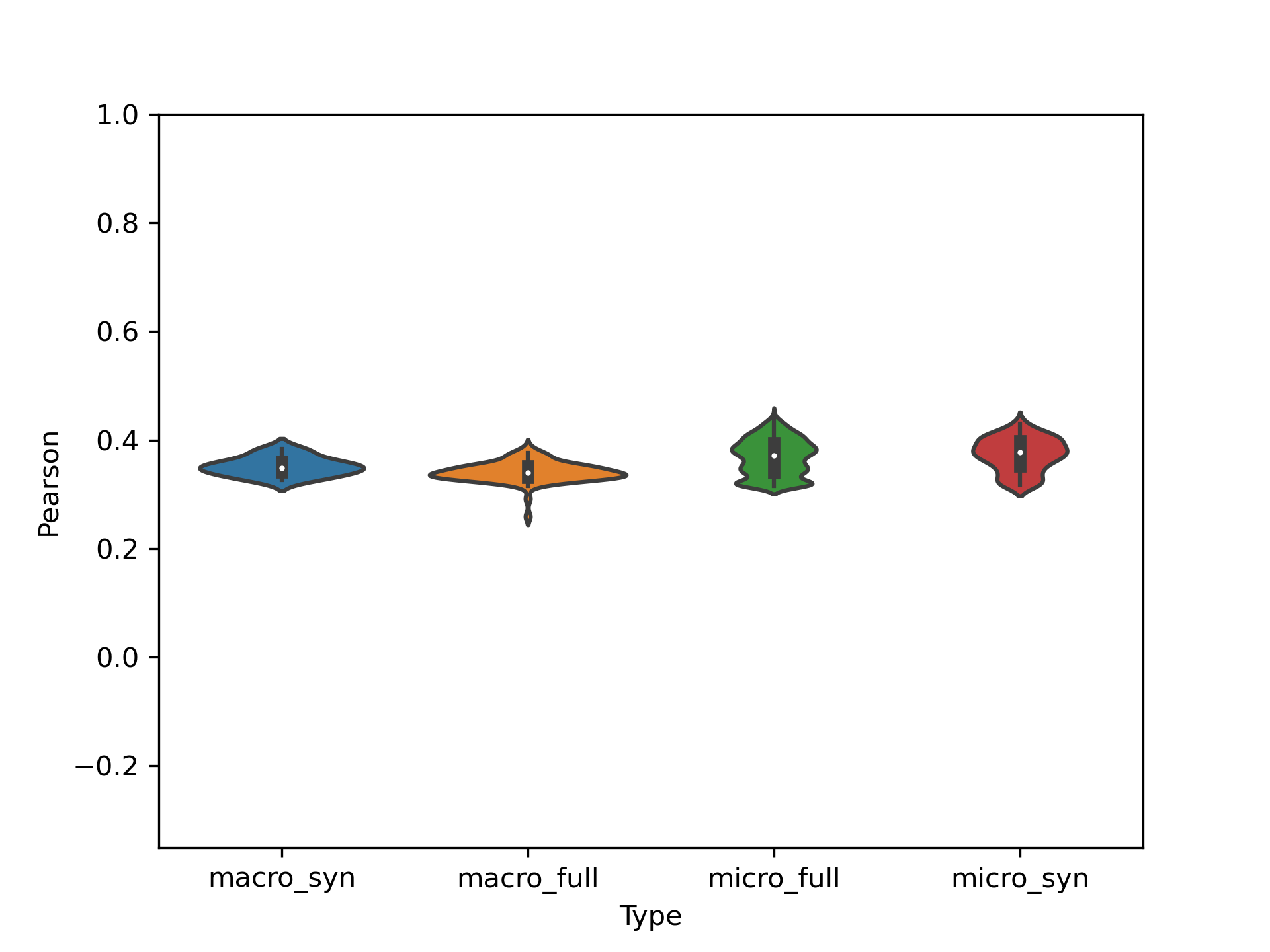}}
\caption{Correlation of Error Distribution Between Late-Stage Grammar and Nodes within the Grammar for Local Dialects in North America. High Correlation indicates that the same dialects are similar in each model type while low correlation indicates that the relationships between dialects for a given component of the grammar differ from the late-stage grammar.}
\end{figure*}

\begin{figure*}
\centering
\fbox{\includegraphics[width = 500pt]{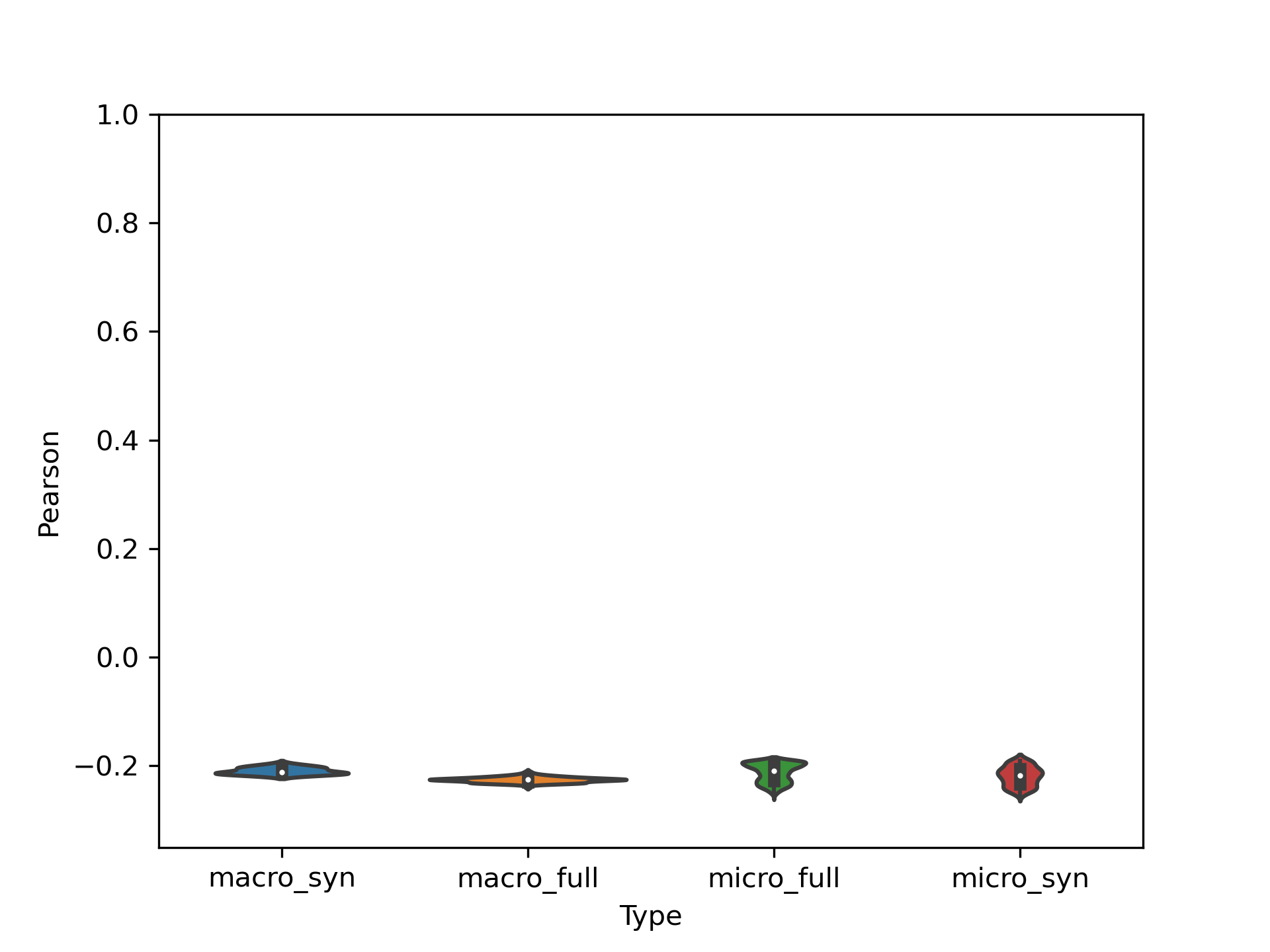}}
\caption{Correlation of Error Distribution Between Late-Stage Grammar and Nodes within the Grammar for Local Dialects in South Asia. High Correlation indicates that the same dialects are similar in each model type while low correlation indicates that the relationships between dialects for a given component of the grammar differ from the late-stage grammar.}
\end{figure*}

\begin{figure*}
\centering
\fbox{\includegraphics[width = 500pt]{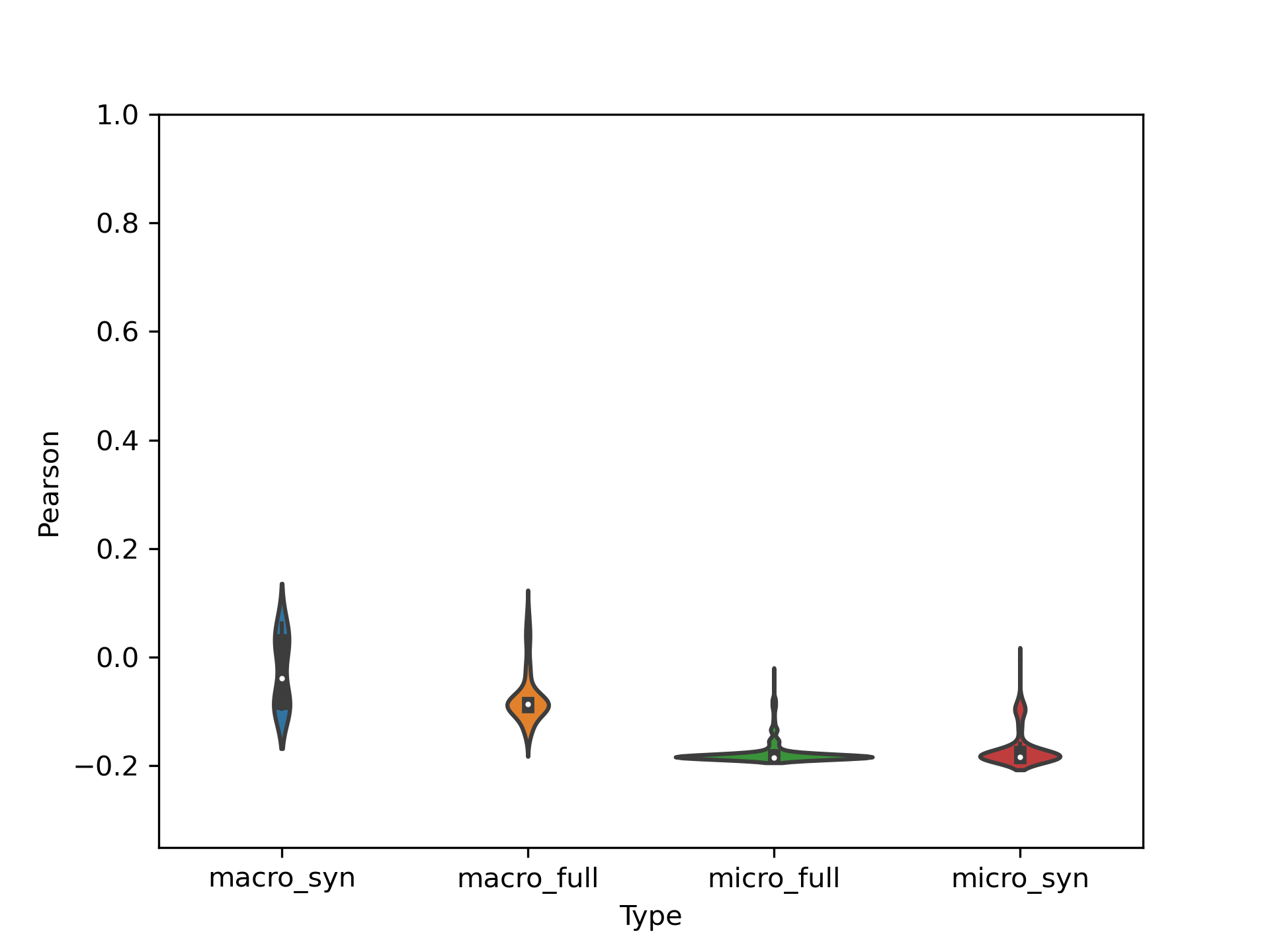}}
\caption{Correlation of Error Distribution Between Late-Stage Grammar and Nodes within the Grammar for Local Dialects in Western Europe. High Correlation indicates that the same dialects are similar in each model type while low correlation indicates that the relationships between dialects for a given component of the grammar differ from the late-stage grammar.}
\end{figure*}

\begin{figure*}
\centering
\fbox{\includegraphics[width = 500pt]{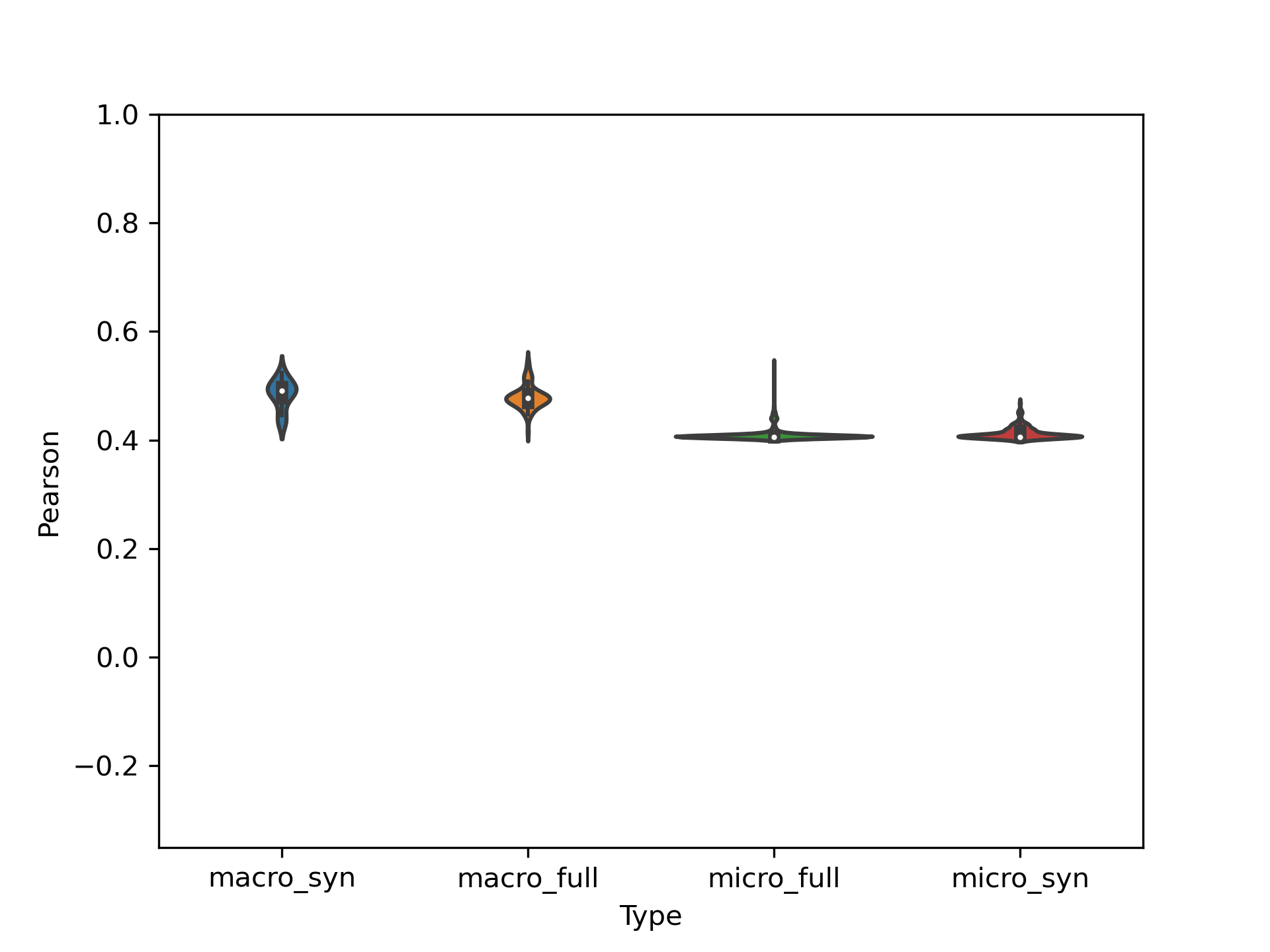}}
\caption{Correlation of Error Distribution Between Late-Stage Grammar and Nodes within the Grammar for Local Dialects in Oceania. High Correlation indicates that the same dialects are similar in each model type while low correlation indicates that the relationships between dialects for a given component of the grammar differ from the late-stage grammar.}
\end{figure*}

\clearpage

\begin{figure*}
\centering
\fbox{\includegraphics[width = 500pt]{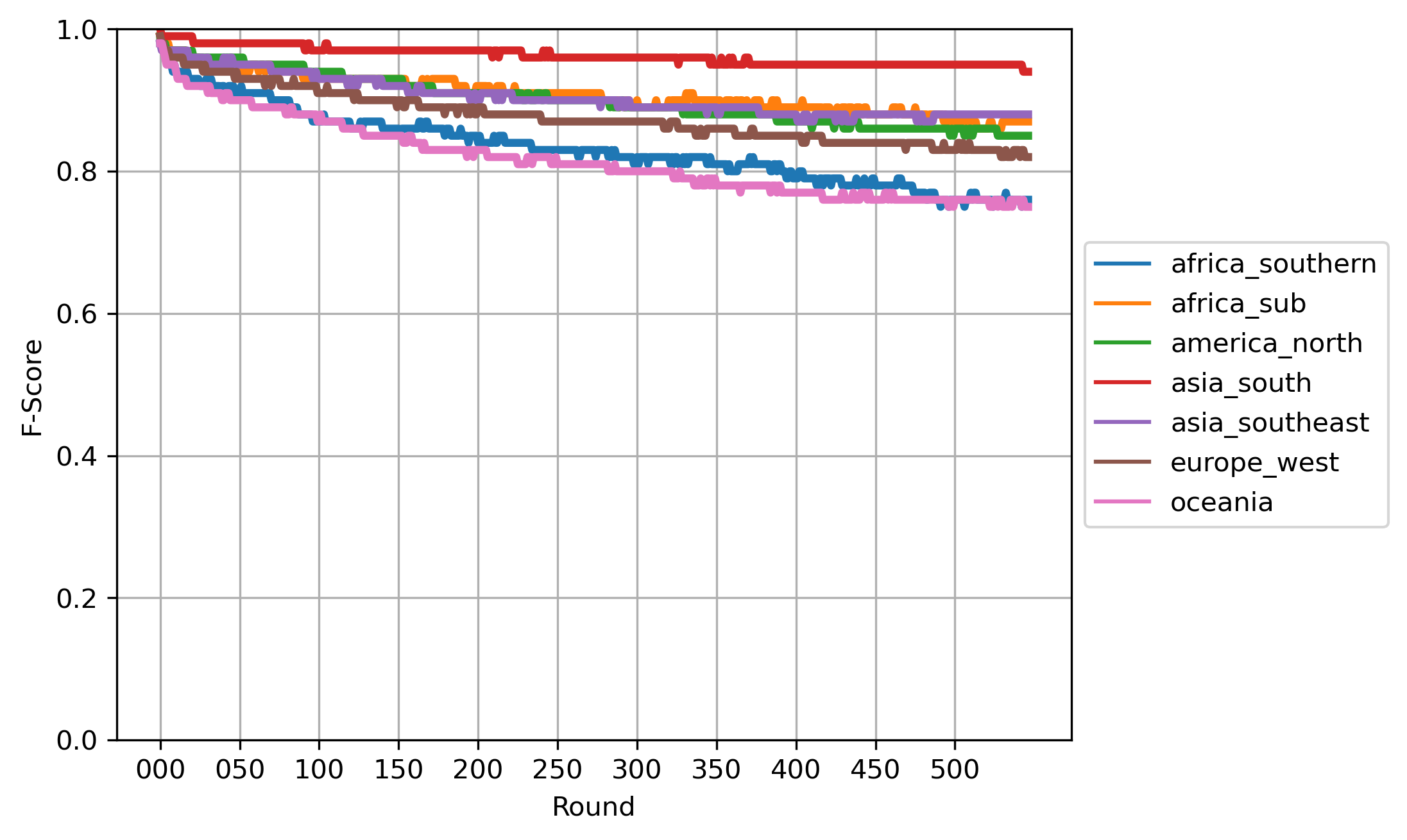}}
\caption{Unmasking of Regional Dialects with the Late-Stage Grammar. Each 50 rounds removes approximately 2.5\% of the grammar, so that round 500 includes only 75\% of the original grammar. The most predictive features are removed first.}
\end{figure*}

\begin{figure*}
\centering
\fbox{\includegraphics[width = 500pt]{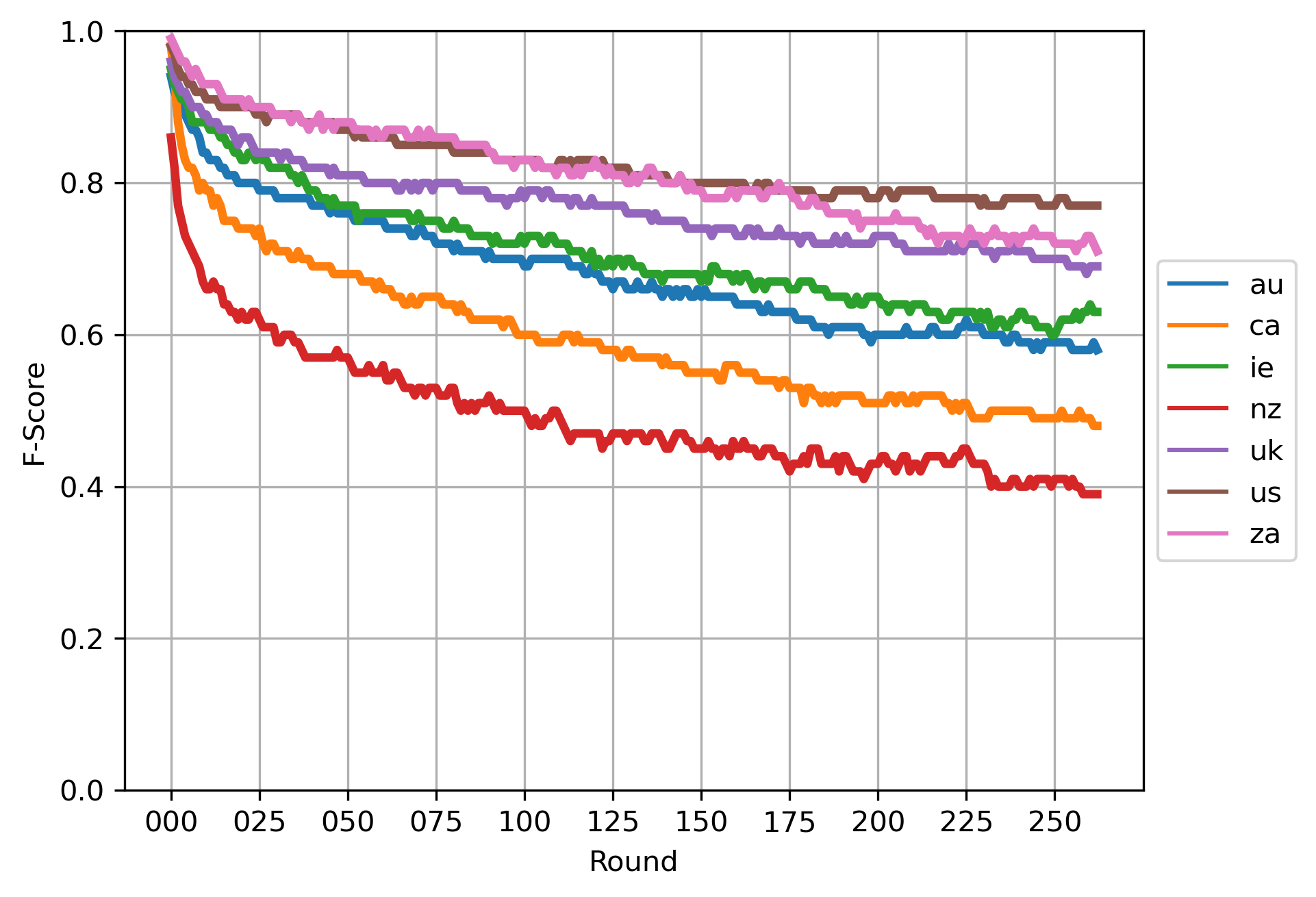}}
\caption{Unmasking of National Dialects (Inner-Circle) with the Late-Stage Grammar. Each 50 rounds removes approximately 2.5\% of the grammar, so that round 500 includes only 75\% of the original grammar. The most predictive features are removed first.}
\end{figure*}

\begin{figure*}
\centering
\fbox{\includegraphics[width = 500pt]{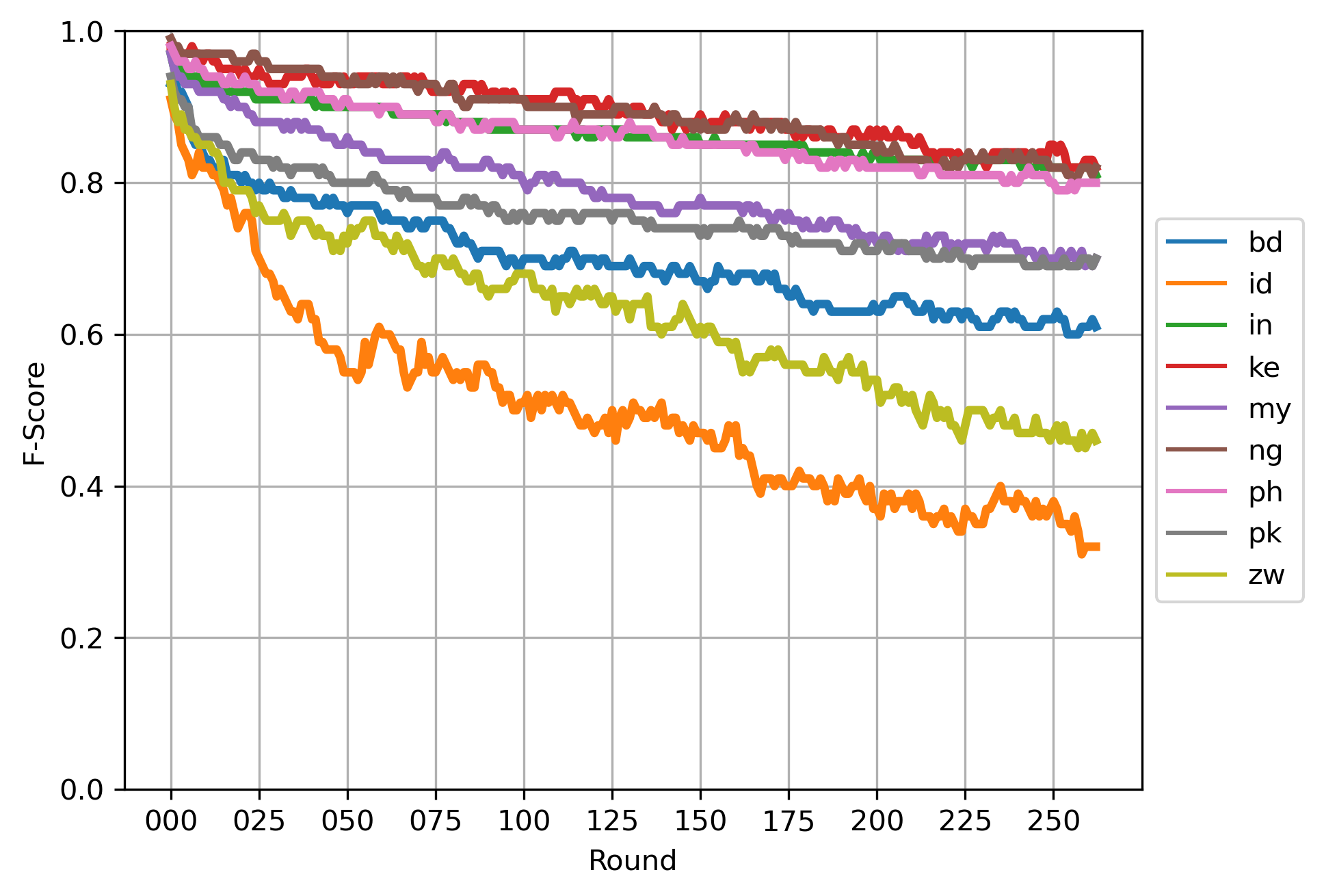}}
\caption{Unmasking of National Dialects (Outer-Circle) with the Late-Stage Grammar. Each 50 rounds removes approximately 2.5\% of the grammar, so that round 500 includes only 75\% of the original grammar. The most predictive features are removed first.}
\end{figure*}



